\documentclass[journal]{IEEEtran}
\usepackage{graphicx,color,psfrag}
\usepackage{subcaption}
\usepackage{booktabs,dcolumn}
\usepackage[cmex10]{amsmath}
\usepackage{amssymb} 
\ifCLASSOPTIONcompsoc
\usepackage[caption=false,font=normalsize,labelfont=sf,textfont=sf]{subfig}
\else
\usepackage[caption=false,font=footnotesize]{subfig}
\fi
\usepackage{url}

\begin{document}
\onecolumn
This manuscript has been submitted to IEEE transactions on robotics for possible publication. Once this manuscript is accepted for publication, the copyright will be transferred without notice. 
\newpage
\twocolumn
	%
	\title{Adaptive Control Strategy for\\ Constant Optical Flow Divergence Landing}
	%
	%
	%
	
	\author{H.~W.~Ho,
		G.~C.~H.~E.~de~Croon,
		E.~van~Kampen,
		Q.~P.~Chu,
		and~M.~Mulder
		\thanks{H.~W.~Ho is with the Micro Air Vehicle laboratory of the Faculty of Aerospace Engineering, Delft University of Technology, 2629HS Delft, The Netherlands and the Faculty of Aerospace Engineering, Universiti Sains Malaysia, 14300 Nibong Tebal, Malaysia (e-mail: H.W.Ho@TUDelft.nl).}
		\thanks{G.~C.~H.~E.~de~Croon is with the Micro Air Vehicle laboratory of the Faculty of Aerospace Engineering, Delft University of Technology, 2629HS Delft, The Netherlands (e-mail: G.C.H.E.deCroon@TUDelft.nl).}
		\thanks{E.~van~Kampen, Q.~P.~Chu, M.~Mulder are with Control and Simulation Section of the Faculty of Aerospace Engineering, Delft University of Technology, 2629HS Delft, The Netherlands (e-mail: E.vanKampen@TUDelft.nl; Q.P.Chu@TUDelft.nl; M.Mulder@TUDelft.nl).}
	}

	\maketitle
	
		
	\begin{abstract}
	Bio-inspired methods can provide efficient solutions to perform autonomous landing for Micro Air Vehicles (MAVs). Flying insects such as honeybees perform vertical landings by keeping flow divergence constant. This leads to an exponential decay of both height and vertical velocity, and allows for smooth and safe landings. However, the presence of noise and delay in obtaining flow divergence estimates will cause instability of the landing when the control gains are not adapted to the height. In this paper, we propose a strategy that deals with this fundamental problem of optical flow control. The key to the strategy lies in the use of a recent theory that allows the MAV to see distance by means of its control instability. At the start of a landing, the MAV detects the height by means of an oscillating movement and sets the control gains accordingly. Then, during descent, the gains are reduced exponentially, with mechanisms in place to reduce or increase the gains if the actual trajectory deviates too much from an ideal constant divergence landing. Real-world experiments demonstrate stable landings of the MAV in both indoor and windy outdoor environments.
	\end{abstract}
	
	
	%
	\IEEEpeerreviewmaketitle
	
\section{Introduction}
\label{sec:Introduction}



%
\IEEEPARstart{P}{erforming} a smooth landing is challenging for Micro Air Vehicles (MAVs) which have payload constraints and limited computing capability. Many earlier studies have used traditional methods of sensing and navigating involving active sensors, such as a laser range finder \cite{achtelik2009stereo, bachrach2009autonomous}, or a stereo camera \cite{goldberg2002stereo, park2012landing}. Although they give accurate and redundant measurements, they are costly and heavy for MAVs. In addition, methods using stereo camera are limited in their perception range. A monocular camera would be preferred for MAVs, also due to its light weight and low power consumption \cite{sanchez2014approach, li2015monocular}. 

Visual Simultaneous Localization and Mapping (SLAM) is the most commonly used method for navigating using a monocular camera. This method locates all the detected features in the camera field of view and determines the vehicle's location and 3D-structure of the landing surface at these points \cite{celik2008mvcslam,davison2007monoslam}. Although its computational efficiency and accuracy has been improved over the years \cite{blosch2010vision,weiss2011monocular,artieda2009visual}, visual SLAM still requires more computational resources than strictly necessary for landing. 

Besides SLAM, another method is inspired by tiny flying insects, which accomplish complex flight control tasks using limited neural and sensory resources \cite{franceschini2009neuromimetic}. For instance, honeybees mainly rely on their eyes to perform smooth landings \cite{baird2005visual,baird2006visual}. They possess extremely efficient and robust solutions to tackle these control problems. These bio-inspired solutions can provide design principles for flight control strategies in MAVs \cite{ruffier2005optic, garratt2007biologicallyphd}. 

In flying insects, optical flow is probably the mostly used source of visual information. When approaching a ground surface, the expansion of the flow (flow divergence) provides a perception of the observer's relative motion to the ground. Honeybees reduce their speeds to almost zero at touchdown by keeping flow divergence constant \cite{baird2013universal, ruffier2008aerial}. This strategy is typically praised, since it does not seem to rely on knowledge about the height and approaching speed. Optical flow only captures the ratio of height and velocity. Several studies have implemented a constant flow divergence strategy with a fixed-gain controller, e.g., on an MAV for landing \cite{herisse2008hovering, alkowatly2014bioinspired}. Additionally, time-to-contact (the reciprocal of flow divergence) based landing strategies have been performed on rotorcraft, such as the TauPilot \cite{kendoul2014four}, which implemented the guidance and control scheme as proposed by Tau theory \cite{lee1998guiding,lee2009general}.

However, there is a fundamental problem when actually controlling a constant flow divergence landing, i.e. the controller gain(s) depends on the height. No true solution has been presented for this problem. A few studies, focusing on decreasing time-to-contact landings, schedule the gains according to the time-to-contact \cite{kendoul2014four, de2015controlling}. However, the initial gain depends on the height and velocity. Deviating significantly from these assumed initial conditions leads to severely hampered performance or even a crash. In addition, such gain-scheduling is not possible for constant flow divergence (or constant time-to-contact) landings. 

A recent theory developed by one of the authors shows the relationship between the height and the controller gain \cite{de2016monocular}. It was shown analytically that for a specific fixed gain, optical flow control becomes unstable at a specific given height. Instead of regarding this as a problem, it was proposed that the MAV can detect oncoming self-induced oscillations and use these for estimating the height. One of the uses of this theory is to detect self-induced oscillations close to the landing surface for landing triggering.

The main contribution of this paper is that we leverage this theory to propose a strategy that solves the fundamental problem of gain selection for optical flow landing. This strategy allows for a smooth and high performance landing of the MAV, by adjusting the controller gains during landing. Two other, smaller, contributions of this paper are: 1) to develop a novel way to detect oscillations in real-time based on observation of the flow divergence, and 2) to characterize the flow divergence measured from a single camera mounted on a quadrotor MAV (this part of the work is partly based on a conference paper \cite{ho2016characterization}). 

The remainder of the article is structured as follows. In Section~\ref{sec:DivGuidanceControl} we provide some background on the constant flow divergence guidance strategy, and show results of computer simulations with a conventional control scheme assuming a perfect flow divergence estimate. Section~\ref{sec:Characterization} presents a characterization of the flow divergence estimates as obtained with a monocular camera. Section~\ref{sec:FixedGainControl} then shows the analysis of the conventional closed-loop control with the delay and noisy estimates in both computer simulation and flight test. In Section~\ref{sec:AdaptiveGainControl} the adaptive control scheme is introduced to deal with the instability problems encountered with the conventional control scheme. Section~\ref{sec:FlightTests} demonstrates the real-world experiments. Conclusions and recommendations are given in Section~\ref{sec:Conclusion}.


\section{Background}
\label{sec:DivGuidanceControl}
\subsection{Flow Divergence}
\begin{figure}[thpb]
	\centering
	\begin{psfrags}%
\psfragscanon%
%
\psfrag{x}[b][b][1]{\color[rgb]{0,0,0}\setlength{\tabcolsep}{0pt}\begin{tabular}{c}$y^b$\end{tabular}}%
\psfrag{y}[b][b][1]{\color[rgb]{0,0,0}\setlength{\tabcolsep}{0pt}\begin{tabular}{c}$x^b$\end{tabular}}%
\psfrag{z}[b][b][1]{\color[rgb]{0,0,0}\setlength{\tabcolsep}{0pt}\begin{tabular}{c}$z^b$\end{tabular}}%
\psfrag{d}[b][b][1]{\color[rgb]{0,0,0}\setlength{\tabcolsep}{0pt}\begin{tabular}{c}$x^w$\end{tabular}}%
\psfrag{e}[b][b][1]{\color[rgb]{0,0,0}\setlength{\tabcolsep}{0pt}\begin{tabular}{c}$y^w$\end{tabular}}%
\psfrag{f}[b][b][1]{\color[rgb]{0,0,0}\setlength{\tabcolsep}{0pt}\begin{tabular}{c}$z^w$\end{tabular}}%

\psfrag{o}[t][t][1]{\color[rgb]{0,0,0}\setlength{\tabcolsep}{0pt}\begin{tabular}{c}$o^b$\end{tabular}}%
\psfrag{p}[t][t][1]{\color[rgb]{0,0,0}\setlength{\tabcolsep}{0pt}\begin{tabular}{c}$o^w$\end{tabular}}%

\psfrag{b}[b][b][1]{\color[rgb]{0,0,0}\setlength{\tabcolsep}{0pt}\begin{tabular}{c}b: body\end{tabular}}%
\psfrag{w}[b][b][1]{\color[rgb]{0,0,0}\setlength{\tabcolsep}{0pt}\begin{tabular}{c}w: world\end{tabular}}%
\includegraphics[width=0.35\textwidth]{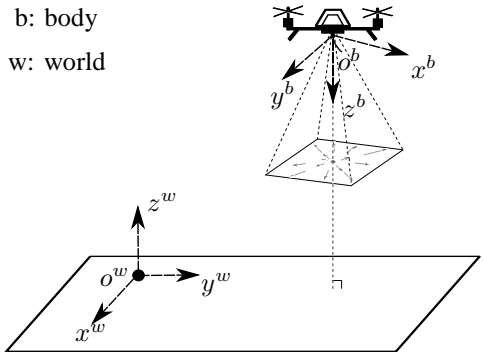}
\end{psfrags}%
%

	\caption{MAV body ($o^bx^by^bz^b$) and world ($o^wx^wy^wz^w$) reference frames.}
	\label{fig:uav_body_axis}
\end{figure}

The definition of axes used in this paper is illustrated in Fig.~\ref{fig:uav_body_axis}. In this figure, the body reference frame is denoted as $o^bx^by^bz^b$, where $o^b$ is located at the center of gravity of an MAV, $x^b$ points forward, $y^b$ is starboard, and $z^b$ points downward. World reference frame is a fixed frame on the ground and uses North-East-Up ($x^w$-$y^w$-$z^w$) system. 

\begin{figure}[thpb]
	\centering
	\begin{psfrags}%
\psfragscanon%
%
\psfrag{a}[c][c][1]{\color[rgb]{0,0,0}\setlength{\tabcolsep}{0pt}\begin{tabular}{c}$o^c$\end{tabular}}%
\psfrag{b}[c][c][1]{\color[rgb]{0,0,0}\setlength{\tabcolsep}{0pt}\begin{tabular}{c}$y^c$\end{tabular}}%
\psfrag{c}[c][c][1]{\color[rgb]{0,0,0}\setlength{\tabcolsep}{0pt}\begin{tabular}{c}$x^c$\end{tabular}}%
\psfrag{d}[c][c][0.8]{\color[rgb]{0,0,0}\setlength{\tabcolsep}{0pt}\begin{tabular}{c}$v$\end{tabular}}%
\psfrag{e}[c][c][0.8]{\color[rgb]{0,0,0}\setlength{\tabcolsep}{0pt}\begin{tabular}{c}$u$\end{tabular}}%
\psfrag{f}[c][c][1]{\color[rgb]{0,0,0}\setlength{\tabcolsep}{0pt}\begin{tabular}{c} \end{tabular}}%
\psfrag{g}[c][c][1]{\color[rgb]{0,0,0}\setlength{\tabcolsep}{0pt}\begin{tabular}{c}c: camera\end{tabular}}%
\psfrag{h}[c][c][0.8]{\color[rgb]{0,0,0}\setlength{\tabcolsep}{0pt}\begin{tabular}{c}$x$\end{tabular}}%
\psfrag{i}[c][c][0.8]{\color[rgb]{0,0,0}\setlength{\tabcolsep}{0pt}\begin{tabular}{c}$y$\end{tabular}}%

\includegraphics[width=0.35\textwidth]{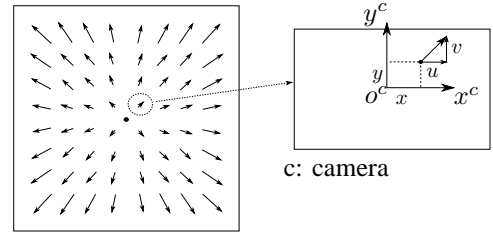}
\end{psfrags}%
%

	\caption{Divergence of optical flow (flow divergence) when an observer is approaching a surface.}
	\label{fig:divergence_flow}
\end{figure}

Since a camera is attached to the body of the MAV, the camera reference frame ($o^cx^cy^cz^c$) can be assumed to be aligned with the body reference frame, where the camera is facing downward or in the positive $z^c$ direction. When the MAV is approaching a flat ground surface, the camera observes a divergent pattern of optical flow, the so-called flow divergence shown in Fig.~\ref{fig:divergence_flow}. 
Flow divergence of a feature point in an image is defined as the partial derivatives of its velocities ($u$ and $v$) at its position ($x$ and $y$) in the camera image coordinates system \cite{mccarthy2008robust}:
\begin{equation}
	D(x,y) = \frac{\partial u(x,y)}{\partial x} + \frac{\partial v(x,y)}{\partial y}.
	\label{equation:divergence_definition_image}
\end{equation}
This is illustrated on the right side of Fig.~\ref{fig:divergence_flow} in which one of optical flow vectors is enlarged and shown in an image. By examining all $D(x,y)$ of the available features in an image, a `global' flow divergence, $D$ which is of particular interest in this paper can be obtained. For vertical landings, flow divergence can be defined as the ratio of the vertical velocity, $V_Z$ to its height from the ground, $Z$:
\begin{equation}
	D = \frac{V_Z}{Z}.
	\label{equation:Div}
\end{equation}
Flow divergence can be used to determine the time-to-contact, $\tau$ which is reciprocal to $D$. For vertical landing of an MAV, $Z>0$, $V_Z<0$, and thus $D<0$.

\subsection{Constant Flow Divergence Guidance Strategy}
The common guidance strategy using flow divergence for vertical landing is the constant flow divergence approach \cite{herisse2008hovering}. By keeping the flow divergence constant, $D=-k$, we can control the dynamics of the landing with a suitable $k$. To examine the influence of $k$ on this strategy during a landing maneuver, the equations of motion describing the height $Z$, vertical velocity $V_Z$, and vertical acceleration $A_Z$ of the MAV are:
\begin{equation}
Z = Z_0e^{-kt},~V_Z = -kZ_0e^{-kt},~A_Z = k^2Z_0e^{-kt},
\label{equation:ZConstantD}
\end{equation}
\noindent where $Z_0$ is the initial height above the landing surface.

\begin{figure}[thpb]
	\centering
%
%
\begin{psfrags}%
\psfragscanon%
\newcommand{\tsize}{0.7}
%
\psfrag{s11}[t][t][\tsize]{\color[rgb]{0,0,0}\setlength{\tabcolsep}{0pt}\begin{tabular}{c}$Z~(m)$\end{tabular}}%
\psfrag{s12}[b][b][\tsize]{\color[rgb]{0,0,0}\setlength{\tabcolsep}{0pt}\begin{tabular}{c}Time (s)\end{tabular}}%
\psfrag{s13}[t][t][\tsize]{\color[rgb]{0,0,0}\setlength{\tabcolsep}{0pt}\begin{tabular}{c}$V_Z~(m/s)$\end{tabular}}%
\psfrag{s14}[b][b][\tsize]{\color[rgb]{0,0,0}\setlength{\tabcolsep}{0pt}\begin{tabular}{c}Time (s)\end{tabular}}%
\psfrag{s15}[t][t][\tsize]{\color[rgb]{0,0,0}\setlength{\tabcolsep}{0pt}\begin{tabular}{c}$A_Z~(m/s^2)$\end{tabular}}%
\psfrag{s16}[b][b][\tsize]{\color[rgb]{0,0,0}\setlength{\tabcolsep}{0pt}\begin{tabular}{c}Time (s)\end{tabular}}%
\psfrag{s17}[t][t][\tsize]{\color[rgb]{0,0,0}\setlength{\tabcolsep}{0pt}\begin{tabular}{c}$D^*~(1/s)$\end{tabular}}%
\psfrag{s18}[b][b][\tsize]{\color[rgb]{0,0,0}\setlength{\tabcolsep}{0pt}\begin{tabular}{c}Time (s)\end{tabular}}%
\psfrag{s22}[][]{\color[rgb]{0,0,0}\setlength{\tabcolsep}{0pt}\begin{tabular}{c} \end{tabular}}%
\psfrag{s23}[][]{\color[rgb]{0,0,0}\setlength{\tabcolsep}{0pt}\begin{tabular}{c} \end{tabular}}%
\psfrag{s24}[l][l][\tsize]{\color[rgb]{0,0,0}k = 14}%
\psfrag{s25}[l][l][\tsize]{\color[rgb]{0,0,0}k = -2}%
\psfrag{s26}[l][l][\tsize]{\color[rgb]{0,0,0}k = 2}%
\psfrag{s27}[l][l][\tsize]{\color[rgb]{0,0,0}k = 6}%
\psfrag{s28}[l][l][\tsize]{\color[rgb]{0,0,0}k = 10}%
\psfrag{s29}[l][l][\tsize]{\color[rgb]{0,0,0}k = 14}%
%
\psfrag{x01}[t][t][\tsize]{0}%
\psfrag{x02}[t][t][\tsize]{}%
\psfrag{x03}[t][t][\tsize]{0.2}%
\psfrag{x04}[t][t][\tsize]{}%
\psfrag{x05}[t][t][\tsize]{0.4}%
\psfrag{x06}[t][t][\tsize]{}%
\psfrag{x07}[t][t][\tsize]{0.6}%
\psfrag{x08}[t][t][\tsize]{}%
\psfrag{x09}[t][t][\tsize]{0.8}%
\psfrag{x10}[t][t][\tsize]{}%
\psfrag{x11}[t][t][\tsize]{1}%
\psfrag{x12}[t][t][\tsize]{0}%
\psfrag{x13}[t][t][\tsize]{}%
\psfrag{x14}[t][t][\tsize]{0.2}%
\psfrag{x15}[t][t][\tsize]{}%
\psfrag{x16}[t][t][\tsize]{0.4}%
\psfrag{x17}[t][t][\tsize]{}%
\psfrag{x18}[t][t][\tsize]{0.6}%
\psfrag{x19}[t][t][\tsize]{}%
\psfrag{x20}[t][t][\tsize]{0.8}%
\psfrag{x21}[t][t][\tsize]{}%
\psfrag{x22}[t][t][\tsize]{1}%
\psfrag{x23}[t][t][\tsize]{0}%
\psfrag{x24}[t][t][\tsize]{}%
\psfrag{x25}[t][t][\tsize]{0.2}%
\psfrag{x26}[t][t][\tsize]{}%
\psfrag{x27}[t][t][\tsize]{0.4}%
\psfrag{x28}[t][t][\tsize]{}%
\psfrag{x29}[t][t][\tsize]{0.6}%
\psfrag{x30}[t][t][\tsize]{}%
\psfrag{x31}[t][t][\tsize]{0.8}%
\psfrag{x32}[t][t][\tsize]{}%
\psfrag{x33}[t][t][\tsize]{1}%
\psfrag{x34}[t][t][\tsize]{0}%
\psfrag{x35}[t][t][\tsize]{}%
\psfrag{x36}[t][t][\tsize]{0.2}%
\psfrag{x37}[t][t][\tsize]{}%
\psfrag{x38}[t][t][\tsize]{0.4}%
\psfrag{x39}[t][t][\tsize]{}%
\psfrag{x40}[t][t][\tsize]{0.6}%
\psfrag{x41}[t][t][\tsize]{}%
\psfrag{x42}[t][t][\tsize]{0.8}%
\psfrag{x43}[t][t][\tsize]{}%
\psfrag{x44}[t][t][\tsize]{1}%
%
\psfrag{v01}[r][r][\tsize]{}%
\psfrag{v02}[r][r][\tsize]{-14}%
\psfrag{v03}[r][r][\tsize]{}%
\psfrag{v04}[r][r][\tsize]{-10}%
\psfrag{v05}[r][r][\tsize]{}%
\psfrag{v06}[r][r][\tsize]{-6}%
\psfrag{v07}[r][r][\tsize]{}%
\psfrag{v08}[r][r][\tsize]{-2}%
\psfrag{v09}[r][r][\tsize]{}%
\psfrag{v10}[r][r][\tsize]{2}%
\psfrag{v11}[r][r][\tsize]{0}%
\psfrag{v12}[r][r][\tsize]{50}%
\psfrag{v13}[r][r][\tsize]{100}%
\psfrag{v14}[r][r][\tsize]{150}%
\psfrag{v15}[r][r][\tsize]{200}%
\psfrag{v16}[r][r][\tsize]{-15}%
\psfrag{v17}[r][r][\tsize]{-10}%
\psfrag{v18}[r][r][\tsize]{-5}%
\psfrag{v19}[r][r][\tsize]{0}%
\psfrag{v20}[r][r][\tsize]{5}%
\psfrag{v21}[r][r][\tsize]{0}%
\psfrag{v22}[r][r][\tsize]{0.5}%
\psfrag{v23}[r][r][\tsize]{1}%
\psfrag{v24}[r][r][\tsize]{1.5}%
%
\includegraphics[width=0.5\textwidth]{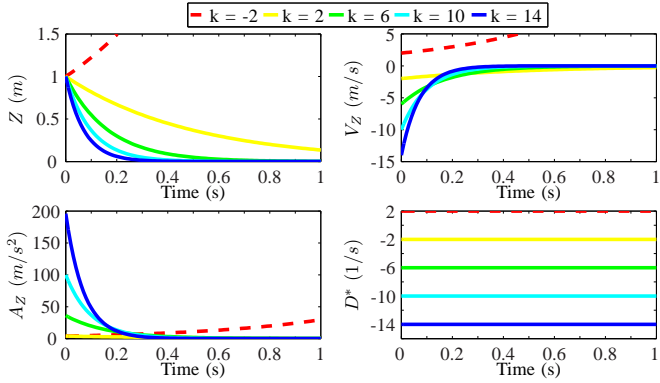}
\end{psfrags}%
%

	\caption{Constant flow divergence guidance with different desired flow divergence $D^*=-k$.}
	\label{fig:ConstantDivergenceGuidance}
\end{figure}

Fig.~\ref{fig:ConstantDivergenceGuidance} shows the effect of $k$ on the height, velocity, and acceleration time histories with the same initial height. It is clear that only flow divergence, $k>0$ will lead to convergence of the states to zero. With different positive values of $k$, we can manipulate the dynamics of the maneuver. For example, the larger the $k$ is, the faster the states converge to zero. The practical feasibility of these maneuvers, however, also depends on the vehicle limitations, such as the maximum vertical velocity that can be achieved.

%


\subsection{Conventional Control Scheme}
\label{subsec:ControlScheme}
To track the desired flow divergence $D^*$, a relatively straightforward proportional feedback controller can be used:
\begin{equation}
	\mu = K_p(D^{*}-D),
	\label{equation:PController}
\end{equation}
\noindent where $K_p$ is the gain of the proportional controller.

To simplify the analysis, we model the dynamics of an object, moving towards a surface in one dimensional space, using a double integrator. The state space model can then be written as:
\begin{equation}
	\dot{\mathbf{x}}(t) = \begin{bmatrix}
		0 & 1 \\
		0 & 0
	\end{bmatrix} \mathbf{x}(t) +
	\begin{bmatrix}
		0 \\
		1
	\end{bmatrix} \mu(t),
	\label{equation:ModelDynamics}
\end{equation}
\begin{equation}
	y(t)=[x_2(t)/x_1(t)]=D,
	\label{equation:ModelObservation}
\end{equation}
\noindent where $\mathbf{x}=[x_1,x_2]^T=[Z,V_Z]^T$ and $\mu$ is the control input.

\begin{figure}[thpb]
	\centering
%
%
\begin{psfrags}%
\psfragscanon%
\newcommand{\tsize}{0.65}
%
\psfrag{s05}[t][t][\tsize]{\color[rgb]{0,0,0}\setlength{\tabcolsep}{0pt}\begin{tabular}{c}Time~(s)\end{tabular}}%
\psfrag{s10}[][]{\color[rgb]{0,0,0}\setlength{\tabcolsep}{0pt}\begin{tabular}{c} \end{tabular}}%
\psfrag{s11}[][]{\color[rgb]{0,0,0}\setlength{\tabcolsep}{0pt}\begin{tabular}{c} \end{tabular}}%
\psfrag{s12}[l][l][\tsize]{\color[rgb]{0,0,0}$D^*(1/s)$}%
\psfrag{s13}[l][l][\tsize]{\color[rgb]{0,0,0}$\mu(m/s^2)$}%
\psfrag{s14}[l][l][\tsize]{\color[rgb]{0,0,0}$D(1/s)$}%
\psfrag{s15}[l][l][\tsize]{\color[rgb]{0,0,0}$Z(m)$}%
\psfrag{s16}[l][l][\tsize]{\color[rgb]{0,0,0}$V_Z(m/s)$}%
\psfrag{s17}[l][l][\tsize]{\color[rgb]{0,0,0}$D^*(1/s)$}%
%
\psfrag{x01}[t][t][\tsize]{0}%
\psfrag{x02}[t][t][\tsize]{2}%
\psfrag{x03}[t][t][\tsize]{4}%
\psfrag{x04}[t][t][\tsize]{6}%
\psfrag{x05}[t][t][\tsize]{8}%
\psfrag{x06}[t][t][\tsize]{10}%
\psfrag{x07}[t][t][\tsize]{12}%
\psfrag{x08}[t][t][\tsize]{14}%
\psfrag{x09}[t][t][\tsize]{16}%
%
\psfrag{v01}[r][r][\tsize]{-1}%
\psfrag{v02}[r][r][\tsize]{0}%
\psfrag{v03}[r][r][\tsize]{1}%
\psfrag{v04}[r][r][\tsize]{2}%
\psfrag{v05}[r][r][\tsize]{3}%
\psfrag{v06}[r][r][\tsize]{4}%
%
\includegraphics[width=0.4\textwidth]{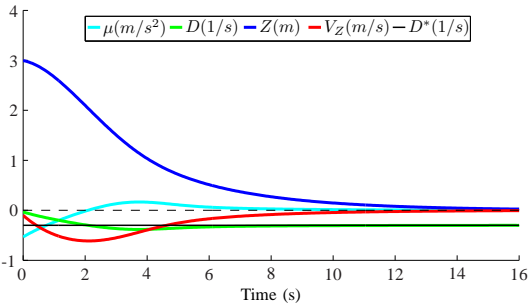}
\end{psfrags}%
%

	\caption{Fixed-gain closed-loop landing control using a constant flow divergence strategy.}
	\label{fig:Perfect_Const_Div_Control}
\end{figure}

Eqs.~(\ref{equation:ModelDynamics}) and (\ref{equation:ModelObservation}) show that the model dynamics are linear but its observation is nonlinear. To visualize the feasibility of the proportional controller to track a constant reference (e.g., $D^* = -0.3$), a time response of the system is plotted in Fig.~\ref{fig:Perfect_Const_Div_Control}. In this figure, both height and velocity are approaching zero in the end. During this maneuver, the vehicle accelerated in the first $2s$ and then decelerated to zero velocity to touch the ground. 
%
\section{Characterization of Flow Divergence}
\label{sec:Characterization}
In the previous section we showed that, with a simple proportional controller and `perfect' estimate of flow divergence, the vehicle can be landed smoothly with zero velocity touching the ground. However, in a real scenario, we cannot avoid having delays and noise in the sensor measurements. For this reason, we will need to characterize the inaccuracies induced in estimating the flow divergence, to investigate the effects of these sensor inaccuracies on the feasibility of using a simple controller in practice. In this paper, we use two different methods to estimate the flow divergence. One is based on flow field fit, the other uses a more direct method to compute the expansion and compression of the optical flow vectors. 

\subsection{Flow Field Divergence Estimator}
We first need to estimate the `raw' flow divergence from a camera sensor. In this study, optical flow vectors are computed based on a sparse corner tracking method using Features from Accelerated Segment Test (FAST) \cite{rosten2006machine, rosten2005fusing}, integrated with a Lucas-Kanade tracker \cite{bouquet2000pyramidal}. The first vision algorithm that estimates the flow divergence of the optic flow field \cite{de2013optic} is based on early findings by Longuet-Higgins and Prazdny \cite{longuet1980interpretation}. The algorithm assumes that (a) a pinhole camera model pointing downward is used, (b) the surface in sight is planar, and (c) the angular rates of the camera can be measured and used to de-rotate the optical flow. Under these assumptions, the optical flow equation can be expressed as follows:
\begin{equation}
	\mathbf{u} = \overbrace{-\omega_x}^{\textbf{$p_{u1}$}} + \overbrace{(\omega_x \zeta + \omega_z)}^{\textbf{$p_{u2}$}} \mathbf{x} + \overbrace{\omega_x \eta}^{\textbf{$p_{u3}$}} \mathbf{y} - \overbrace{\zeta \omega_z}^{\textbf{$p_{u4}$}} \mathbf{x}^2 - \overbrace{\eta \omega_z}^{\textbf{$p_{u5}$}} \mathbf{xy},
	\label{equation:optic_flow_u}
\end{equation}
\begin{equation}
	\mathbf{v} = \overbrace{-\omega_y}^{\textbf{$p_{v1}$}} + \overbrace{\omega_y \zeta}^{\textbf{$p_{v2}$}} \mathbf{x} + \overbrace{(\omega_y \eta + \omega_z)}^{\textbf{$p_{v3}$}} \mathbf{y} - \overbrace{\eta \omega_z}^{\textbf{$p_{v4}$}} \mathbf{y}^2 - \overbrace{\zeta \omega_z}^{\textbf{$p_{v5}$}} \mathbf{xy},
	\label{equation:optic_flow_v}
\end{equation}
\noindent where $\omega_x = V_X/Z$, $\omega_y = V_Y/Z$, and $\omega_z = V_Z/Z$ are the velocities in $x^b$, $y^b$, and $z^b$ direction scaled with respect to the height $Z$. $\zeta$ and $\eta$ are the gradient of the ground surface. 

By re-writing Eqs.~(\ref{equation:optic_flow_u}) and (\ref{equation:optic_flow_v}) into matrix form, as shown in Eq.~(\ref{equation:pu_pv}), the parameter vectors $\mathbf{p_u} = [p_{u1},p_{u2},p_{u3},p_{u4},p_{u5}]$ and $\mathbf{p_v} = [p_{v1},p_{v2},p_{v3},p_{v4},p_{v5}]$ can be estimated using a maximum likelihood linear least squares estimate within a robust random sample consensus (RANSAC) estimation procedure \cite{fischler1981random}:
\begin{equation}
u = \mathbf{p_u} [1, x, y, x^2, xy]^T,~v = \mathbf{p_v} [1, x, y, y^2, xy]^T.
\label{equation:pu_pv}
\end{equation}
The estimated parameters provide important information for bio-inspired navigation, such as ventral flow, surface slope \cite{de2013optic}, flow divergence, time-to-contact, etc. In this study, we are primarily interested in estimating the flow divergence:
\begin{equation}
	\widehat{D} = p_{u2} + p_{v3}.
	\label{equation:divergence_flow}
\end{equation}
Note that Eqs.~(\ref{equation:optic_flow_u}) and (\ref{equation:optic_flow_v}) can be simplified by neglecting the second-order terms, as in this study we focus on landing upon flat surfaces without any inclination ($\zeta=0$ and $\eta=0$). Therefore, a linear fit of the optical flow field can be obtained. 

\subsection{Size Divergence Estimator}
\begin{figure}[thpb]
	\centering
	\begin{psfrags}%
\psfragscanon%
\newcommand{\tsize}{0.7}
%
\psfrag{F}[b][b][\tsize]{\color[rgb]{0,0,0}\setlength{\tabcolsep}{0pt}\begin{tabular}{c}Focal point\end{tabular}}%
\psfrag{i}[b][b][\tsize]{\color[rgb]{0,0,0}\setlength{\tabcolsep}{0pt}\begin{tabular}{c}Image plane\end{tabular}}%
\psfrag{g}[b][b][\tsize]{\color[rgb]{0,0,0}\setlength{\tabcolsep}{0pt}\begin{tabular}{c}Ground\end{tabular}}%
\psfrag{P}[b][b][\tsize]{\color[rgb]{0,0,0}\setlength{\tabcolsep}{0pt}\begin{tabular}{c}Features\end{tabular}}%
\psfrag{f}[b][b][\tsize]{\color[rgb]{0,0,0}\setlength{\tabcolsep}{0pt}\begin{tabular}{c}$f$\end{tabular}}%
\psfrag{d}[c][c][\tsize]{\color[rgb]{0,0,0}\setlength{\tabcolsep}{0pt}\begin{tabular}{c}$l$\end{tabular}}%
\psfrag{s}[t][t][\tsize]{\color[rgb]{0,0,0}\setlength{\tabcolsep}{0pt}\begin{tabular}{c}$L$\end{tabular}}%
\psfrag{Z}[b][b][\tsize]{\color[rgb]{0,0,0}\setlength{\tabcolsep}{0pt}\begin{tabular}{c}$Z$\end{tabular}}%

\psfrag{z}[t][t][\tsize]{\color[rgb]{0,0,0}\setlength{\tabcolsep}{0pt}\begin{tabular}{c}$Z_{t-\Delta t}$\end{tabular}}%
\psfrag{h}[t][t][\tsize]{\color[rgb]{0,0,0}\setlength{\tabcolsep}{0pt}\begin{tabular}{c}$Z_t$\end{tabular}}%
\psfrag{x}[t][t][\tsize]{\color[rgb]{0,0,0}\setlength{\tabcolsep}{0pt}\begin{tabular}{c}$l_{t-\Delta t}$\end{tabular}}%
\psfrag{y}[t][t][\tsize]{\color[rgb]{0,0,0}\setlength{\tabcolsep}{0pt}\begin{tabular}{c}$l_t$\end{tabular}}%

\includegraphics[width=0.45\textwidth]{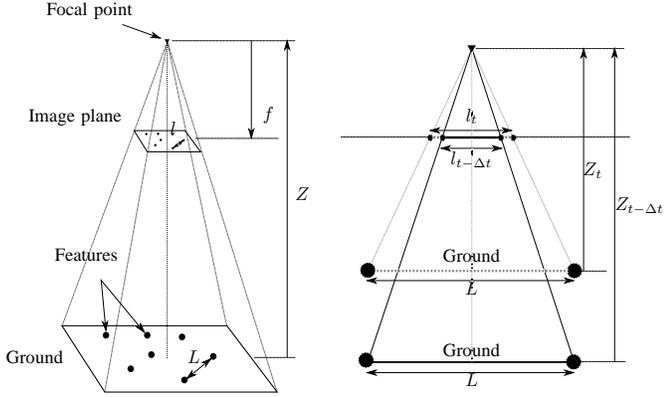}
\end{psfrags}%
%

	\caption{Pinhole camera model (left) and projected lines on image plane when approaching ground (right).}
	\label{fig:pinhole_lines}
\end{figure}

We propose a more straightforward way to estimate flow divergence by measuring the size of the lines connecting between features in consecutive image frames. The left side of Fig.~\ref{fig:pinhole_lines} is a pinhole camera model showing the actual and image size of the lines connecting two features indicated by $L$ and $l$, respectively. On the right side of this figure, the geometry illustrates the change of the size of the projected lines in the image plane, from $l_{t-\Delta t}$ to $l_t$ when the MAV is moving towards the ground, from $Z_{t-\Delta t}$ to $Z_t$. 
Using similar triangles, we can write the following relationships:
\begin{equation}
	\frac{L}{Z_{t-\Delta t}} = \frac{l_{t-\Delta t}}{f},~\frac{L}{Z_t} = \frac{l_t}{f},
	\label{equation:triangular_relationship1}
\end{equation}
\noindent where $f$ is the focal length of the camera while $\Delta t$ is the timestamp between two consecutive images. 
By substituting $L$, we can obtain:
\begin{equation}
	\frac{l_t}{Z_{t-\Delta t}} = \frac{l_{t-\Delta t}}{Z_t}.
	\label{equation:triangular_relationship2}
\end{equation}
From the geometry in Fig.~\ref{fig:pinhole_lines} and Eq.~(\ref{equation:triangular_relationship2}), it is reasonable that when the MAV moves closer to the ground, i.e., $Z_t < Z_{t-\Delta t}$, the size of the line in the image plane becomes larger, i.e., $l_t > l_{t-\Delta t}$. 
By recalling Eq.~(\ref{equation:Div}), flow divergence can also be expressed as follows:
\begin{equation}
		D_t = \frac{1}{\Delta t}[1-\frac{Z_{t-\Delta t}}{Z_t}].
		\label{equation:divergence_derivation}
\end{equation}
By substituting Eq.~(\ref{equation:triangular_relationship2}) into Eq.~(\ref{equation:divergence_derivation}), we can obtain the size divergence of one feature line:
\begin{equation}
		D_{s_t} = \frac{1}{\Delta t}[\frac{l_{t-\Delta t}-l_t}{l_{t-\Delta t}}].
		\label{equation:size_divergence_derivation}
\end{equation}
To obtain a more reliable estimate of size divergence, we can use the average of all detected feature lines, $\widehat{D}_s$ in our computation:
\begin{equation}
	\widehat{D}_s=\frac{1}{N}\sum_{i=1}^{N} D_{s_{t_i}}.
	\label{equation:size_divergence_mean}
\end{equation}
When the MAV moves towards the ground surface, the lines connecting features extend leading to $\widehat{D}_s<0$, and vice versa.



\subsection{Testing Platform and Data Logging}
\label{subsec:TestingPlatform}
\begin{figure}[thpb]
	\centering
%
%
\begin{psfrags}%
\psfragscanon%
\newcommand{\tsize}{0.7}
%
\psfrag{a}[c][c][\tsize]{\color[rgb]{0,0,0}\setlength{\tabcolsep}{0pt}\begin{tabular}{c}MAV\end{tabular}}%
\psfrag{b}[c][c][\tsize]{\color[rgb]{0,0,0}\setlength{\tabcolsep}{0pt}\begin{tabular}{c}Hardware\end{tabular}}%
\psfrag{c}[c][c][\tsize]{\color[rgb]{0,0,0}\setlength{\tabcolsep}{0pt}\begin{tabular}{c}Software~(Paparazzi~Autopilot)\end{tabular}}%
\psfrag{d}[c][c][\tsize]{\color[rgb]{0,0,0}\setlength{\tabcolsep}{0pt}\begin{tabular}{c}Vision~Module\end{tabular}}%
\psfrag{e}[c][c][\tsize]{\color[rgb]{0,0,0}\setlength{\tabcolsep}{0pt}\begin{tabular}{c}Camera\end{tabular}}%
\psfrag{f}[c][c][\tsize]{\color[rgb]{0,0,0}\setlength{\tabcolsep}{0pt}\begin{tabular}{c}IMU\end{tabular}}%
\psfrag{g}[c][c][\tsize]{\color[rgb]{0,0,0}\setlength{\tabcolsep}{0pt}\begin{tabular}{c}Actuator\end{tabular}}%
\psfrag{h}[c][c][\tsize]{\color[rgb]{0,0,0}\setlength{\tabcolsep}{0pt}\begin{tabular}{c}Image\\Capturing\end{tabular}}%
\psfrag{i}[c][c][\tsize]{\color[rgb]{0,0,0}\setlength{\tabcolsep}{0pt}\begin{tabular}{c}Optical~Flow\\Computation\end{tabular}}%
\psfrag{j}[c][c][\tsize]{\color[rgb]{0,0,0}\setlength{\tabcolsep}{0pt}\begin{tabular}{c}Divergence\\Estimator\end{tabular}}%
\psfrag{k}[c][c][\tsize]{\color[rgb]{0,0,0}\setlength{\tabcolsep}{0pt}\begin{tabular}{c}Vertical\\Guidance\end{tabular}}%
\psfrag{l}[c][c][\tsize]{\color[rgb]{0,0,0}\setlength{\tabcolsep}{0pt}\begin{tabular}{c}Control~Loop\\(Rotorcraft)\end{tabular}}%
\psfrag{m}[c][c][\tsize]{\color[rgb]{0,0,0}\setlength{\tabcolsep}{0pt}\begin{tabular}{c}$p,q,r$\end{tabular}}%
\psfrag{n}[c][c][\tsize]{\color[rgb]{0,0,0}\setlength{\tabcolsep}{0pt}\begin{tabular}{c}$x,y$\end{tabular}}%
\psfrag{o}[c][c][\tsize]{\color[rgb]{0,0,0}\setlength{\tabcolsep}{0pt}\begin{tabular}{c}$u,v$\end{tabular}}%
\psfrag{p}[c][c][\tsize]{\color[rgb]{0,0,0}\setlength{\tabcolsep}{0pt}\begin{tabular}{c}$\widehat{D},\widehat{D}_s$\end{tabular}}%
\psfrag{q}[c][c][\tsize]{\color[rgb]{0,0,0}\setlength{\tabcolsep}{0pt}\begin{tabular}{c}$\mu$\end{tabular}}%
%
\includegraphics[width=0.5\textwidth]{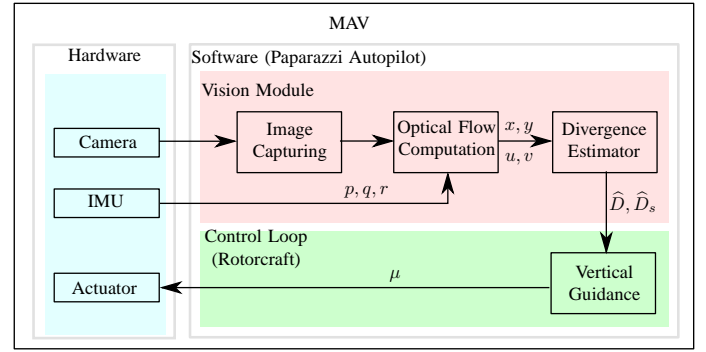}
\end{psfrags}%
%

	\caption{Control architecture of Paparazzi Autopilot and the integration of the computer vision module.}
	\label{fig:HardwareSoftwareVertical}
\end{figure}

A Parrot AR.Drone 2.0 is used as our platform for performing flight tests. The downward-facing camera on this MAV is of particular interest here. We implemented aforementioned algorithms to estimate the flow divergence in Paparazzi Autopilot, an open-source autopilot software\footnote{Paparazzi Autopilot: http://wiki.paparazziuav.org/wiki/Main\_Page}. Fig.~\ref{fig:HardwareSoftwareVertical} shows the overview of the control architecture of Paparazzi and the integration of the computer vision module. All the computer vision and control algorithms are run on-board the MAV. 

In Fig.~\ref{fig:HardwareSoftwareVertical}, images are captured from the downward-looking camera in the vision module. These images are processed using the computer vision algorithms presented in the subsections above. The angular rates ($p$, $q$, $r$) from the IMU are used in the optical flow computations to reduce the effects of MAV rotation on the optical flow vectors. One of the flow divergence estimates, $\widehat{D}$ or $\widehat{D}_s$, is used in the vertical guidance loop to perform automatic landings. 

In our experiments, we start logging the data while the MAV is hovering at a height around $1.5m$. By only controlling the climb rate, we measure the variation of estimated flow divergences $\widehat{D}$ and $\widehat{D}_s$, the height above the ground $Z$, and the vertical velocity $V_Z$. Note that in order to guarantee good measurements of $Z$ and $V_Z$, which will serve as our ground truth for flow divergence $D$, we use an external motion tracking system (OpticTrack), to provide these measurements.

\begin{figure}[thpb]
	\centering
%
%
\begin{psfrags}%
\psfragscanon%
\newcommand{\tsize}{0.7}
\newcommand{\tsizeb}{0.6}
%
\psfrag{s11}[t][t][\tsize]{\color[rgb]{0,0,0}\setlength{\tabcolsep}{0pt}\begin{tabular}{c}$\widehat{D},~\widehat{D}_s~(1/s)$\end{tabular}}%
\psfrag{s15}[][]{\color[rgb]{0,0,0}\setlength{\tabcolsep}{0pt}\begin{tabular}{c} \end{tabular}}%
\psfrag{s16}[][]{\color[rgb]{0,0,0}\setlength{\tabcolsep}{0pt}\begin{tabular}{c} \end{tabular}}%
\psfrag{s17}[l][l][\tsizeb]{\color[rgb]{0,0,0}$\widehat{D}_s$}%
\psfrag{s18}[l][l][\tsizeb]{\color[rgb]{0,0,0}$\widehat{D}$}%
\psfrag{s19}[l][l][\tsizeb]{\color[rgb]{0,0,0}$\widehat{D}_s$}%
\psfrag{s20}[t][t][\tsize]{\color[rgb]{0,0,0}\setlength{\tabcolsep}{0pt}\begin{tabular}{c}$Z~(m)$\end{tabular}}%
\psfrag{s21}[t][t][\tsize]{\color[rgb]{0,0,0}\setlength{\tabcolsep}{0pt}\begin{tabular}{c}$V_Z~(m/s)$\end{tabular}}%
\psfrag{s22}[t][t][\tsize]{\color[rgb]{0,0,0}\setlength{\tabcolsep}{0pt}\begin{tabular}{c}Time~(s)\end{tabular}}%
%
\psfrag{x01}[t][t][\tsize]{0}%
\psfrag{x02}[t][t][\tsize]{1000}%
\psfrag{x03}[t][t][\tsize]{2000}%
\psfrag{x04}[t][t][\tsize]{3000}%
\psfrag{x05}[t][t][\tsize]{4000}%
\psfrag{x06}[t][t][\tsize]{5000}%
\psfrag{x07}[t][t][\tsize]{6000}%
\psfrag{x08}[t][t][\tsize]{0}%
\psfrag{x09}[t][t][\tsize]{1000}%
\psfrag{x10}[t][t][\tsize]{2000}%
\psfrag{x11}[t][t][\tsize]{3000}%
\psfrag{x12}[t][t][\tsize]{4000}%
\psfrag{x13}[t][t][\tsize]{5000}%
\psfrag{x14}[t][t][\tsize]{6000}%
\psfrag{x15}[t][t][\tsize]{0}%
\psfrag{x16}[t][t][\tsize]{1000}%
\psfrag{x17}[t][t][\tsize]{2000}%
\psfrag{x18}[t][t][\tsize]{3000}%
\psfrag{x19}[t][t][\tsize]{4000}%
\psfrag{x20}[t][t][\tsize]{5000}%
\psfrag{x21}[t][t][\tsize]{6000}%
%
\psfrag{v01}[r][r][\tsize]{-1}%
\psfrag{v02}[r][r][\tsize]{}%
\psfrag{v03}[r][r][\tsize]{0}%
\psfrag{v04}[r][r][\tsize]{}%
\psfrag{v05}[r][r][\tsize]{1}%
\psfrag{v06}[r][r][\tsize]{}%
\psfrag{v07}[r][r][\tsize]{1}%
\psfrag{v08}[r][r][\tsize]{}%
\psfrag{v09}[r][r][\tsize]{2}%
\psfrag{v10}[r][r][\tsize]{}%
\psfrag{v11}[r][r][\tsize]{3}%
\psfrag{v12}[r][r][\tsize]{-5}%
\psfrag{v13}[r][r][\tsize]{0}%
\psfrag{v14}[r][r][\tsize]{5}%
%
\includegraphics[width=0.5\textwidth]{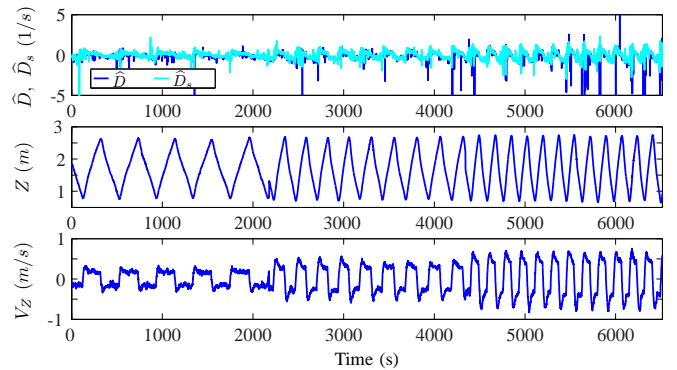}
\end{psfrags}%
%

	\caption{Log of flow divergence estimates, height, and vertical velocity during a vertical maneuver of an AR.Drone 2.0.}
	\label{fig:LOG_CLIMB_DESCEND}
\end{figure}

Fig.~\ref{fig:LOG_CLIMB_DESCEND} shows the measurements log for estimating the delay model and noise model of the flow divergence estimates. We deliberately varied the vehicle climb rate to obtain a wide range of $\widehat{D}$, $\widehat{D}_s$, $Z$ and $V_Z$ measurements.

\subsection{Delay Estimation and Noise Model}
After obtaining flow divergence estimates, this subsection describes how to characterize their properties. There are two steps proposed: 1) to estimate the delays in the estimates, and 2) to model the noise using the delay-corrected data. 

\subsubsection{Delay Estimation}
\begin{figure}[thpb]
	\centering
%
%
\begin{psfrags}%
\psfragscanon%
\newcommand{\tsize}{0.7}
%
\psfrag{s09}[b][b][\tsize]{\color[rgb]{0,0,0}\setlength{\tabcolsep}{0pt}\begin{tabular}{c}$Time~(s)$\end{tabular}}%
\psfrag{s10}[t][t][\tsize]{\color[rgb]{0,0,0}\setlength{\tabcolsep}{0pt}\begin{tabular}{c}$\widehat{D}_{cr},\widehat{D}_{s_{cr}},D~(1/s)$\end{tabular}}%
\psfrag{s14}[][]{\color[rgb]{0,0,0}\setlength{\tabcolsep}{0pt}\begin{tabular}{c} \end{tabular}}%
\psfrag{s15}[][]{\color[rgb]{0,0,0}\setlength{\tabcolsep}{0pt}\begin{tabular}{c} \end{tabular}}%
\psfrag{s16}[l][l][\tsize]{\color[rgb]{0,0,0}$D~$}%
\psfrag{s17}[l][l][\tsize]{\color[rgb]{0,0,0}$\widehat{D}_{cr}$}%
\psfrag{s18}[l][l][\tsize]{\color[rgb]{0,0,0}$\widehat{D}_{s_{cr}}$}%
\psfrag{s19}[l][l][\tsize]{\color[rgb]{0,0,0}$D~$}%
\psfrag{s20}[b][b][\tsize]{\color[rgb]{0,0,0}\setlength{\tabcolsep}{0pt}\begin{tabular}{c}$Time~(s)$\end{tabular}}%
\psfrag{s21}[t][t][\tsize]{\color[rgb]{0,0,0}\setlength{\tabcolsep}{0pt}\begin{tabular}{c}$\widehat{D},\widehat{D}_s,D~(1/s)$\end{tabular}}%
\psfrag{s22}[b][b][\tsize]{\color[rgb]{0,0,0}\setlength{\tabcolsep}{0pt}\begin{tabular}{c}$Time~(s)$\end{tabular}}%
\psfrag{s23}[t][t][\tsize]{\color[rgb]{0,0,0}\setlength{\tabcolsep}{0pt}\begin{tabular}{c}$\widehat{D}_{cr},\widehat{D}_{s_{cr}},D~(1/s)$\end{tabular}}%
%
\psfrag{x01}[t][t][\tsize]{2200}%
\psfrag{x02}[t][t][\tsize]{}%
\psfrag{x03}[t][t][\tsize]{2240}%
\psfrag{x04}[t][t][\tsize]{}%
\psfrag{x05}[t][t][\tsize]{2280}%
\psfrag{x06}[t][t][\tsize]{}%
\psfrag{x07}[t][t][\tsize]{2320}%
\psfrag{x08}[t][t][\tsize]{}%
\psfrag{x09}[t][t][\tsize]{2360}%
\psfrag{x10}[t][t][\tsize]{}%
\psfrag{x11}[t][t][\tsize]{2400}%
\psfrag{x12}[t][t][\tsize]{2200}%
\psfrag{x13}[t][t][\tsize]{}%
\psfrag{x14}[t][t][\tsize]{2240}%
\psfrag{x15}[t][t][\tsize]{}%
\psfrag{x16}[t][t][\tsize]{2280}%
\psfrag{x17}[t][t][\tsize]{}%
\psfrag{x18}[t][t][\tsize]{2320}%
\psfrag{x19}[t][t][\tsize]{}%
\psfrag{x20}[t][t][\tsize]{2360}%
\psfrag{x21}[t][t][\tsize]{}%
\psfrag{x22}[t][t][\tsize]{2400}%
\psfrag{x23}[t][t][\tsize]{1000}%
\psfrag{x24}[t][t][\tsize]{2000}%
\psfrag{x25}[t][t][\tsize]{3000}%
\psfrag{x26}[t][t][\tsize]{4000}%
\psfrag{x27}[t][t][\tsize]{5000}%
\psfrag{x28}[t][t][\tsize]{6000}%
%
\psfrag{v01}[r][r][\tsize]{-2}%
\psfrag{v02}[r][r][\tsize]{}%
\psfrag{v03}[r][r][\tsize]{-1}%
\psfrag{v04}[r][r][\tsize]{}%
\psfrag{v05}[r][r][\tsize]{0}%
\psfrag{v06}[r][r][\tsize]{}%
\psfrag{v07}[r][r][\tsize]{1}%
\psfrag{v08}[r][r][\tsize]{}%
\psfrag{v09}[r][r][\tsize]{2}%
\psfrag{v10}[r][r][\tsize]{-2}%
\psfrag{v11}[r][r][\tsize]{}%
\psfrag{v12}[r][r][\tsize]{-1}%
\psfrag{v13}[r][r][\tsize]{}%
\psfrag{v14}[r][r][\tsize]{0}%
\psfrag{v15}[r][r][\tsize]{}%
\psfrag{v16}[r][r][\tsize]{1}%
\psfrag{v17}[r][r][\tsize]{}%
\psfrag{v18}[r][r][\tsize]{2}%
\psfrag{v19}[r][r][\tsize]{-2}%
\psfrag{v20}[r][r][\tsize]{}%
\psfrag{v21}[r][r][\tsize]{-1}%
\psfrag{v22}[r][r][\tsize]{}%
\psfrag{v23}[r][r][\tsize]{0}%
\psfrag{v24}[r][r][\tsize]{}%
\psfrag{v25}[r][r][\tsize]{1}%
\psfrag{v26}[r][r][\tsize]{}%
\psfrag{v27}[r][r][\tsize]{2}%
%
\includegraphics[width=0.5\textwidth]{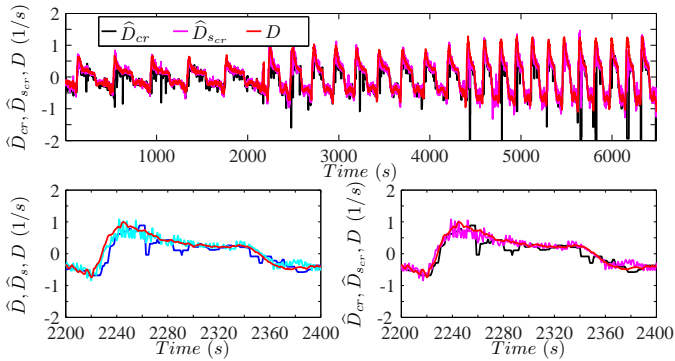}
\end{psfrags}%
%

	\caption{The estimated and delay-corrected flow divergences, together with their ground truth.}
	\label{fig:DIV_Correct_CLIMB_DESCEND}
\end{figure}

We estimated the delay $Lag_i$ of every sample $i=1,2,...,N$ by comparing the datasets of flow divergence estimates $\widehat{D}$ and $\widehat{D}_s$ with the ground truth $D$. First, $W$ windowed samples, $f(i,i+W)$ and $g(i+m,i+W+m)$ of ground truth and flow divergence estimates, respectively, were selected, where $m$ is the moving index. Second, the sums of square error between both samples sets were computed. Then, by repeating the aforementioned steps with different sample sets of flow divergence estimates (from $m=0$ to $m=M$), we looked for the sample set with minimum error. The number of estimates lagging behind the ground truth can then be estimated as:
\begin{equation}
	Lag_i = \arg~\!\min_m~\sum_{j=0}^{W} (f(i,j)-g(i,j,m))^2.
	\label{equation:lag}
\end{equation}
Based on observation of the estimates compared with the ground truth, the delays are expected to be consistent. Multiplying the average of $Lag$ with the sampling time $\Delta t$, we can obtain the time delay of each data set. The estimates have, on average, a lag of $2$ and $1$, which are equivalent to $0.1s$ and $0.05s$ for $\widehat{D}$ and $\widehat{D}_s$, respectively ($\Delta t \approx 0.05s$). These lags are used to correct the flow divergence estimates. Fig.~\ref{fig:DIV_Correct_CLIMB_DESCEND} illustrates that the corrected flow divergence $\widehat{D}_{cr}$, and the corrected size divergence $\widehat{D}_{s_{cr}}$, indeed better match the ground truth $D$. In this figure, the bottom left plots show the enlarged view of flow divergence estimates with the ground truth from $2200s$ to $2400s$ while their corresponding plots which are corrected for the delay are presented on the bottom right. 

\noindent Note that to avoid the estimate of $Lag$ to be driven by noise and outliers, the flow divergence estimates were pre-filtered using a moving median filter. The delay caused by this filter (i.e., $(N_{win}-1)/2$, where $N_{win}$ is its window size) was subtracted from the estimated $Lag$ to obtain the actual lag.

\subsubsection{Noise Model}
\begin{figure}[thpb]
	\centering
%
%
\begin{psfrags}%
\psfragscanon%
\newcommand{\tsize}{0.7}
%
\psfrag{s09}[t][t][\tsize]{\color[rgb]{0,0,0}\setlength{\tabcolsep}{0pt}\begin{tabular}{c}$D~(1/s)$\end{tabular}}%
\psfrag{s10}[t][t][\tsize]{\color[rgb]{0,0,0}\setlength{\tabcolsep}{0pt}\begin{tabular}{c}$\widehat{D}~(1/s)$\end{tabular}}%
\psfrag{s14}[][]{\color[rgb]{0,0,0}\setlength{\tabcolsep}{0pt}\begin{tabular}{c} \end{tabular}}%
\psfrag{s15}[][]{\color[rgb]{0,0,0}\setlength{\tabcolsep}{0pt}\begin{tabular}{c} \end{tabular}}%
\psfrag{s16}[l][l][\tsize]{\color[rgb]{0,0,0}Fit Model}%
\psfrag{s17}[l][l][\tsize]{\color[rgb]{0,0,0}Raw}%
\psfrag{s18}[l][l][\tsize]{\color[rgb]{0,0,0}Delay Corrected}%
\psfrag{s19}[l][l][\tsize]{\color[rgb]{0,0,0}Ideal}%
\psfrag{s20}[l][l][\tsize]{\color[rgb]{0,0,0}Fit Model}%
\psfrag{s21}[t][t][\tsize]{\color[rgb]{0,0,0}\setlength{\tabcolsep}{0pt}\begin{tabular}{c}$D~(1/s)$\end{tabular}}%
\psfrag{s22}[t][t][\tsize]{\color[rgb]{0,0,0}\setlength{\tabcolsep}{0pt}\begin{tabular}{c}$\widehat{D}_s~(1/s)$\end{tabular}}%
\psfrag{s26}[][]{\color[rgb]{0,0,0}\setlength{\tabcolsep}{0pt}\begin{tabular}{c} \end{tabular}}%
\psfrag{s27}[][]{\color[rgb]{0,0,0}\setlength{\tabcolsep}{0pt}\begin{tabular}{c} \end{tabular}}%
\psfrag{s28}[l][l][\tsize]{\color[rgb]{0,0,0}Fit Model}%
\psfrag{s29}[l][l][\tsize]{\color[rgb]{0,0,0}Raw}%
\psfrag{s30}[l][l][\tsize]{\color[rgb]{0,0,0}Delay Corrected}%
\psfrag{s31}[l][l][\tsize]{\color[rgb]{0,0,0}Ideal}%
\psfrag{s32}[l][l][\tsize]{\color[rgb]{0,0,0}Fit Model}%
%
\psfrag{x01}[t][t][\tsize]{-1.5}%
\psfrag{x02}[t][t][\tsize]{-1}%
\psfrag{x03}[t][t][\tsize]{-0.5}%
\psfrag{x04}[t][t][\tsize]{0}%
\psfrag{x05}[t][t][\tsize]{0.5}%
\psfrag{x06}[t][t][\tsize]{1}%
\psfrag{x07}[t][t][\tsize]{1.5}%
\psfrag{x08}[t][t][\tsize]{-1.5}%
\psfrag{x09}[t][t][\tsize]{-1}%
\psfrag{x10}[t][t][\tsize]{-0.5}%
\psfrag{x11}[t][t][\tsize]{0}%
\psfrag{x12}[t][t][\tsize]{0.5}%
\psfrag{x13}[t][t][\tsize]{1}%
\psfrag{x14}[t][t][\tsize]{1.5}%
%
\psfrag{v01}[r][r][\tsize]{-5}%
\psfrag{v02}[r][r][\tsize]{-4}%
\psfrag{v03}[r][r][\tsize]{-3}%
\psfrag{v04}[r][r][\tsize]{-2}%
\psfrag{v05}[r][r][\tsize]{-1}%
\psfrag{v06}[r][r][\tsize]{0}%
\psfrag{v07}[r][r][\tsize]{1}%
\psfrag{v08}[r][r][\tsize]{2}%
\psfrag{v09}[r][r][\tsize]{3}%
\psfrag{v10}[r][r][\tsize]{4}%
\psfrag{v11}[r][r][\tsize]{5}%
\psfrag{v12}[r][r][\tsize]{-5}%
\psfrag{v13}[r][r][\tsize]{-4}%
\psfrag{v14}[r][r][\tsize]{-3}%
\psfrag{v15}[r][r][\tsize]{-2}%
\psfrag{v16}[r][r][\tsize]{-1}%
\psfrag{v17}[r][r][\tsize]{0}%
\psfrag{v18}[r][r][\tsize]{1}%
\psfrag{v19}[r][r][\tsize]{2}%
\psfrag{v20}[r][r][\tsize]{3}%
\psfrag{v21}[r][r][\tsize]{4}%
\psfrag{v22}[r][r][\tsize]{5}%
%
\includegraphics[width=0.5\textwidth]{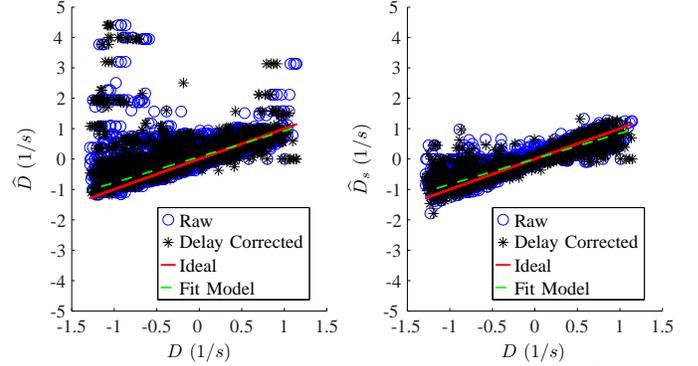}
\end{psfrags}%
%

	\caption{Deviation of the estimated flow divergences $\widehat{D}$ (left) and $\widehat{D}_s$ (right) from their ground truth $D$.}
	\label{fig:DIV_DIV_Ground_CLIMB_DESCEND}
\end{figure}

After correcting for the delay, we can proceed to model the noise of the flow divergence estimates.  Fig.~\ref{fig:DIV_DIV_Ground_CLIMB_DESCEND} plots the flow divergence estimates $\widehat{D}$ and $\widehat{D}_s$ against the corresponding ground truth $D$. This figure illustrates the deviation of the estimated flow divergence from its ideal condition. There are two groups of estimated flow divergence plotted in the figure, i.e., the flow divergence without (circles) and with delay correction (asterisks), respectively. The root mean square errors (RMSEs) of $\widehat{D}_{cr}$ and $\widehat{D}_{s_{cr}}$ with delay correction ($\approx 0.6059~1/s$ and $\approx 0.1469~1/s$) are smaller than their corresponding RMSEs without delay correction ($\approx 0.6206~1/s$ and $\approx 0.1526~1/s$). Overall, both are noisy signals and slightly deviate from their ideal condition, especially when the flow divergence becomes more negative and more positive. This could happen when the ground features can hardly be tracked due to, e.g., an aggressive maneuver, or when the vehicle is either very close to the ground or far away from the ground. 

To take into account this condition in the noise model, we fitted a linear function, as shown in Eq.~(\ref{equation:LinearFit}) to the delay-corrected estimates using a bisquare weights regression. This method is preferable as it minimizes the influence of outliers on the fit by giving less weight to the data far away from the fitted line. The fitted models are drawn in as a \textit{dashed line} in Fig.~\ref{fig:DIV_DIV_Ground_CLIMB_DESCEND}:
\begin{equation}
	f_1(D)=a\cdot D+b,
	\label{equation:LinearFit}
\end{equation}
\noindent where $a$ and $b$ are the fit coefficients. The fitted models for $\widehat{D}$ and $\widehat{D}_s$ are given as $\widehat{D}_{fit}$ and $\widehat{D}_{s_{fit}}$.

\begin{figure}[thpb]
	\centering
%
%
\begin{psfrags}%
\psfragscanon%
\newcommand{\tsize}{0.7}
%
\psfrag{s09}[t][t][\tsize]{\color[rgb]{0,0,0}\setlength{\tabcolsep}{0pt}\begin{tabular}{c}$D~(1/s)$\end{tabular}}%
\psfrag{s10}[b][b][\tsize]{\color[rgb]{0,0,0}\setlength{\tabcolsep}{0pt}\begin{tabular}{c}$|\widehat{D}-\widehat{D}_{Fit}|~(1/s)$\end{tabular}}%
\psfrag{s14}[][]{\color[rgb]{0,0,0}\setlength{\tabcolsep}{0pt}\begin{tabular}{c} \end{tabular}}%
\psfrag{s15}[][]{\color[rgb]{0,0,0}\setlength{\tabcolsep}{0pt}\begin{tabular}{c} \end{tabular}}%
\psfrag{s16}[l][l][\tsize]{\color[rgb]{0,0,0}fitted line}%
\psfrag{s17}[l][l][\tsize]{\color[rgb]{0,0,0}fitted line}%
\psfrag{s18}[t][t][\tsize]{\color[rgb]{0,0,0}\setlength{\tabcolsep}{0pt}\begin{tabular}{c}$D~(1/s)$\end{tabular}}%
\psfrag{s19}[b][b][\tsize]{\color[rgb]{0,0,0}\setlength{\tabcolsep}{0pt}\begin{tabular}{c}$|\widehat{D}_s-\widehat{D}_{s_{Fit}}|~(1/s)$\end{tabular}}%
\psfrag{s23}[][]{\color[rgb]{0,0,0}\setlength{\tabcolsep}{0pt}\begin{tabular}{c} \end{tabular}}%
\psfrag{s24}[][]{\color[rgb]{0,0,0}\setlength{\tabcolsep}{0pt}\begin{tabular}{c} \end{tabular}}%
\psfrag{s25}[l][l][\tsize]{\color[rgb]{0,0,0}fitted line}%
\psfrag{s26}[l][l][\tsize]{\color[rgb]{0,0,0}fitted line}%
%
\psfrag{x01}[t][t][\tsize]{-1.5}%
\psfrag{x02}[t][t][\tsize]{-1}%
\psfrag{x03}[t][t][\tsize]{-0.5}%
\psfrag{x04}[t][t][\tsize]{0}%
\psfrag{x05}[t][t][\tsize]{0.5}%
\psfrag{x06}[t][t][\tsize]{1}%
\psfrag{x07}[t][t][\tsize]{1.5}%
\psfrag{x08}[t][t][\tsize]{-1.5}%
\psfrag{x09}[t][t][\tsize]{-1}%
\psfrag{x10}[t][t][\tsize]{-0.5}%
\psfrag{x11}[t][t][\tsize]{0}%
\psfrag{x12}[t][t][\tsize]{0.5}%
\psfrag{x13}[t][t][\tsize]{1}%
\psfrag{x14}[t][t][\tsize]{1.5}%
%
\psfrag{v01}[r][r][\tsize]{0}%
\psfrag{v02}[r][r][\tsize]{}%
\psfrag{v03}[r][r][\tsize]{0.2}%
\psfrag{v04}[r][r][\tsize]{}%
\psfrag{v05}[r][r][\tsize]{0.4}%
\psfrag{v06}[r][r][\tsize]{}%
\psfrag{v07}[r][r][\tsize]{0.6}%
\psfrag{v08}[r][r][\tsize]{}%
\psfrag{v09}[r][r][\tsize]{0.8}%
\psfrag{v10}[r][r][\tsize]{}%
\psfrag{v11}[r][r][\tsize]{1}%
\psfrag{v12}[r][r][\tsize]{0}%
\psfrag{v13}[r][r][\tsize]{}%
\psfrag{v14}[r][r][\tsize]{0.2}%
\psfrag{v15}[r][r][\tsize]{}%
\psfrag{v16}[r][r][\tsize]{0.4}%
\psfrag{v17}[r][r][\tsize]{}%
\psfrag{v18}[r][r][\tsize]{0.6}%
\psfrag{v19}[r][r][\tsize]{}%
\psfrag{v20}[r][r][\tsize]{0.8}%
\psfrag{v21}[r][r][\tsize]{}%
\psfrag{v22}[r][r][\tsize]{1}%
%
\includegraphics[width=0.5\textwidth]{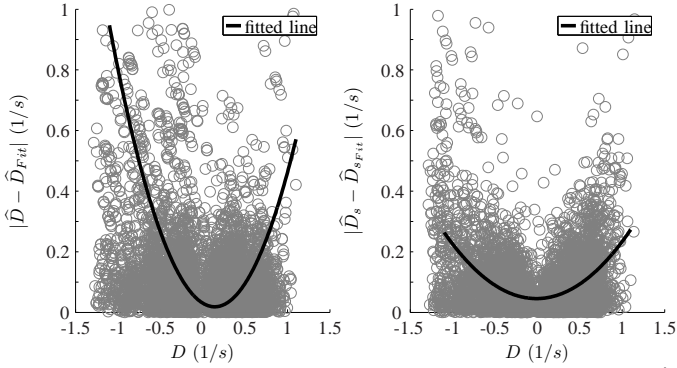}
\end{psfrags}%
%

	\caption{Absolute errors of the flow divergence estimates $\widehat{D}$ (left) and $\widehat{D}_s$ (right) versus their ground truth $D$.}
	\label{fig:DIV_ERR_CLIMB_DESCEND}
\end{figure}

The next step is to estimate the variances of the estimates with respect to the fitted models. Fig.~\ref{fig:DIV_ERR_CLIMB_DESCEND} shows the absolute errors of the estimates versus ground truth flow divergence. The \textit{circles} represent the absolute errors of the estimate with respect to the fitted model. Since we can observe (also from Fig.~\ref{fig:DIV_DIV_Ground_CLIMB_DESCEND}) that the errors are higher for larger values of $D$, a quadratic fit function is more suitable:
\begin{equation}
	f_2(D)=c\cdot D^2+d\cdot D+e,
	\label{equation:Quadratic}
\end{equation}
\noindent where $c$, $d$, and $e$ are the fit coefficients. The \textit{solid line} in Fig.~\ref{fig:DIV_ERR_CLIMB_DESCEND} is the fitted line of these absolute errors which represents the variance. Table~\ref{tab:FittingCoefficient} lists the fit coefficients of the noise models ($a$, $b$, $c$, $d$, and $e$) for both $\widehat{D}$ and $\widehat{D}_s$ that are used in the computer simulations and the controller design. 

\begin{table}
	\centering
	\caption{\label{tab:FittingCoefficient} Fit coefficients of noise models for $\widehat{D}$ and $\widehat{D}_s$.}
	\begin{tabular}{@{}l*{2}{D{.}{.}{5}}@{}}
		\hline\hline
		\\
		\multicolumn{1}{c}{Coefficients}& \multicolumn{1}{c}{$\widehat{D}$} & \multicolumn{1}{c}{$\widehat{D}_s$} \\
		\hline
		\multicolumn{1}{c}{a} & 0.8519     & 0.8393  \\
		\multicolumn{1}{c}{b} & -0.0655    & -0.0060 \\
		\multicolumn{1}{c}{c} & 0.5766     & 0.1841  \\
		\multicolumn{1}{c}{d} & 0.1918     & -0.0043 \\
		\multicolumn{1}{c}{e} & 0.0412     & 0.0455  \\
		\hline\hline
	\end{tabular}
\end{table}

\begin{figure}[thpb]
	\centering
%
%
\begin{psfrags}%
\psfragscanon%
\newcommand{\tsize}{0.7}
%
\psfrag{s01}[t][t][\tsize]{\color[rgb]{0.15,0.15,0.15}\setlength{\tabcolsep}{0pt}\begin{tabular}{c}$err_{\widehat{D}}~(1/s)$\end{tabular}}%
\psfrag{s02}[b][b][\tsize]{\color[rgb]{0.15,0.15,0.15}\setlength{\tabcolsep}{0pt}\begin{tabular}{c}$PDF$\end{tabular}}%
\psfrag{s03}[t][t][\tsize]{\color[rgb]{0.15,0.15,0.15}\setlength{\tabcolsep}{0pt}\begin{tabular}{c}$err_{\widehat{D}_s}~(1/s)$\end{tabular}}%
\psfrag{s04}[b][b][\tsize]{\color[rgb]{0.15,0.15,0.15}\setlength{\tabcolsep}{0pt}\begin{tabular}{c}$PDF$\end{tabular}}%
%
\color[rgb]{0.15,0.15,0.15}%
%
\psfrag{x01}[t][t][\tsize]{-2}%
\psfrag{x02}[t][t][\tsize]{-1}%
\psfrag{x03}[t][t][\tsize]{0}%
\psfrag{x04}[t][t][\tsize]{1}%
\psfrag{x05}[t][t][\tsize]{2}%
\psfrag{x06}[t][t][\tsize]{-2}%
\psfrag{x07}[t][t][\tsize]{0}%
\psfrag{x08}[t][t][\tsize]{2}%
\psfrag{x09}[t][t][\tsize]{4}%
\psfrag{x10}[t][t][\tsize]{6}%
%
\psfrag{v01}[r][r][\tsize]{0}%
\psfrag{v02}[r][r][\tsize]{0.5}%
\psfrag{v03}[r][r][\tsize]{1}%
\psfrag{v04}[r][r][\tsize]{1.5}%
\psfrag{v05}[r][r][\tsize]{0}%
\psfrag{v06}[r][r][\tsize]{0.2}%
\psfrag{v07}[r][r][\tsize]{0.4}%
\psfrag{v08}[r][r][\tsize]{0.6}%
\psfrag{v09}[r][r][\tsize]{0.8}%
\psfrag{v10}[r][r][\tsize]{1}%
\psfrag{v11}[r][r][\tsize]{1.2}%
\psfrag{v12}[r][r][\tsize]{1.4}%
%
\includegraphics[width=0.5\textwidth]{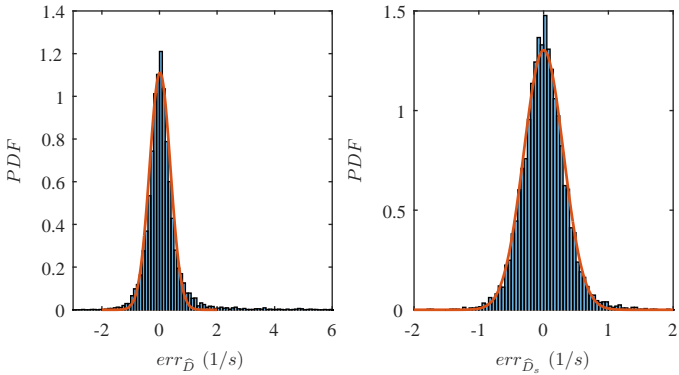}
\end{psfrags}%
%

	\caption{Probability density functions of the estimate errors of $\widehat{D}$ (left) and $\widehat{D}_s$ (right) with respect to the fitted model.}
	\label{fig:PDF_CLIMB_DESCEND}
\end{figure}

Fig.~\ref{fig:PDF_CLIMB_DESCEND} presents the probability density functions of the models errors, $err_{\widehat{D}}$ and $err_{\widehat{D}_s}$. These model errors are computed by subtracting the data generated based on Eqs.~(\ref{equation:LinearFit}) and (\ref{equation:Quadratic}) from the corresponding flow divergence estimates. In this figure, we fitted a Gaussian model (\textit{solid lines}) to each distribution, we obtain $err_{\widehat{D}}=N(0.0173,0.1292)$ and $err_{\widehat{D}_s}=N(6.1979\times10^{-4},0.0937)$. This shows that both estimated noise models are quite accurate. 

\begin{figure}[thpb]
	\centering
%
%
\begin{psfrags}%
\psfragscanon%
\newcommand{\tsize}{0.7}
%
\psfrag{s01}[t][t][\tsize]{\color[rgb]{0.15,0.15,0.15}\setlength{\tabcolsep}{0pt}\begin{tabular}{c}$D~(1/s)$\end{tabular}}%
\psfrag{s02}[b][b][\tsize]{\color[rgb]{0.15,0.15,0.15}\setlength{\tabcolsep}{0pt}\begin{tabular}{c}$\widehat{D}~(1/s)$\end{tabular}}%
\psfrag{s06}[][]{\color[rgb]{0,0,0}\setlength{\tabcolsep}{0pt}\begin{tabular}{c} \end{tabular}}%
\psfrag{s07}[][]{\color[rgb]{0,0,0}\setlength{\tabcolsep}{0pt}\begin{tabular}{c} \end{tabular}}%
\psfrag{s12}[l][l][\tsize]{\color[rgb]{0,0,0}Fit Model}%
\psfrag{s13}[l][l][\tsize]{\color[rgb]{0,0,0}Delay Corrected}%
\psfrag{s14}[l][l][\tsize]{\color[rgb]{0,0,0}Noise Model}%
\psfrag{s15}[l][l][\tsize]{\color[rgb]{0,0,0}Fit Model}%
\psfrag{s16}[t][t][\tsize]{\color[rgb]{0.15,0.15,0.15}\setlength{\tabcolsep}{0pt}\begin{tabular}{c}$D~(1/s)$\end{tabular}}%
\psfrag{s17}[b][b][\tsize]{\color[rgb]{0.15,0.15,0.15}\setlength{\tabcolsep}{0pt}\begin{tabular}{c}$\widehat{D}_s~(1/s)$\end{tabular}}%
\psfrag{s21}[][]{\color[rgb]{0,0,0}\setlength{\tabcolsep}{0pt}\begin{tabular}{c} \end{tabular}}%
\psfrag{s22}[][]{\color[rgb]{0,0,0}\setlength{\tabcolsep}{0pt}\begin{tabular}{c} \end{tabular}}%
\psfrag{s27}[l][l][\tsize]{\color[rgb]{0,0,0}Fit Model}%
\psfrag{s28}[l][l][\tsize]{\color[rgb]{0,0,0}Delay Corrected}%
\psfrag{s29}[l][l][\tsize]{\color[rgb]{0,0,0}Noise Model}%
\psfrag{s30}[l][l][\tsize]{\color[rgb]{0,0,0}Fit Model}%
%
\psfrag{x01}[t][t][\tsize]{-1.5}%
\psfrag{x02}[t][t][\tsize]{-1}%
\psfrag{x03}[t][t][\tsize]{-0.5}%
\psfrag{x04}[t][t][\tsize]{0}%
\psfrag{x05}[t][t][\tsize]{0.5}%
\psfrag{x06}[t][t][\tsize]{1}%
\psfrag{x07}[t][t][\tsize]{1.5}%
\psfrag{x08}[t][t][\tsize]{-1.5}%
\psfrag{x09}[t][t][\tsize]{-1}%
\psfrag{x10}[t][t][\tsize]{-0.5}%
\psfrag{x11}[t][t][\tsize]{0}%
\psfrag{x12}[t][t][\tsize]{0.5}%
\psfrag{x13}[t][t][\tsize]{1}%
\psfrag{x14}[t][t][\tsize]{1.5}%
%
\psfrag{v01}[r][r][\tsize]{-5}%
\psfrag{v02}[r][r][\tsize]{-4}%
\psfrag{v03}[r][r][\tsize]{-3}%
\psfrag{v04}[r][r][\tsize]{-2}%
\psfrag{v05}[r][r][\tsize]{-1}%
\psfrag{v06}[r][r][\tsize]{0}%
\psfrag{v07}[r][r][\tsize]{1}%
\psfrag{v08}[r][r][\tsize]{2}%
\psfrag{v09}[r][r][\tsize]{3}%
\psfrag{v10}[r][r][\tsize]{4}%
\psfrag{v11}[r][r][\tsize]{5}%
\psfrag{v12}[r][r][\tsize]{-5}%
\psfrag{v13}[r][r][\tsize]{-4}%
\psfrag{v14}[r][r][\tsize]{-3}%
\psfrag{v15}[r][r][\tsize]{-2}%
\psfrag{v16}[r][r][\tsize]{-1}%
\psfrag{v17}[r][r][\tsize]{0}%
\psfrag{v18}[r][r][\tsize]{1}%
\psfrag{v19}[r][r][\tsize]{2}%
\psfrag{v20}[r][r][\tsize]{3}%
\psfrag{v21}[r][r][\tsize]{4}%
\psfrag{v22}[r][r][\tsize]{5}%
%
\includegraphics[width=0.5\textwidth]{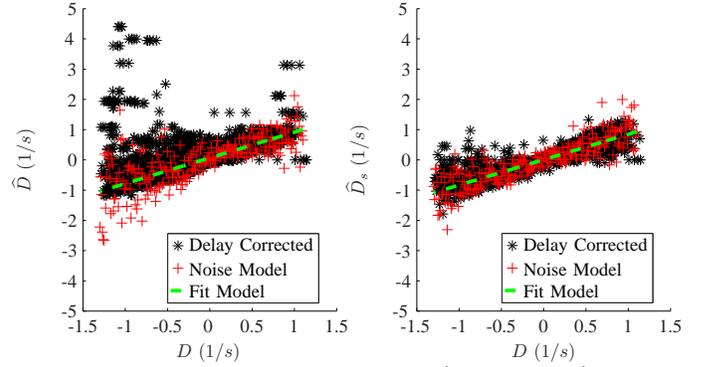}
\end{psfrags}%
%

	\caption{The flow divergence estimates $\widehat{D}$ (left) and $\widehat{D}_s$ (right) and the data generated using the noise models.}
	\label{fig:DIV_Noise_Model_CLIMB_DESCEND}
\end{figure}

To validate the noise model, we plotted generated data with estimates of $\widehat{D}$ and $\widehat{D}_s$ as shown in Fig.~\ref{fig:DIV_Noise_Model_CLIMB_DESCEND}. In this figure, the generated data sets show that the errors are higher for larger values of $D$, which are similar to the observation in Fig.~\ref{fig:DIV_ERR_CLIMB_DESCEND}. However, some inevitable outliers in the estimates remain, as there are neglected in both noise models. 

Note that since $\widehat{D}_s$ has less delay and noise, we use it for the following simulations and flight tests. Additionally, the term `flow divergence' and the symbol $\widehat{D}$ will be used, instead of size divergence and its symbol $\widehat{D}_s$, in order to have a more general expression and to maintain consistency.

\section{Influence of delay and noise on Fixed-Gain Constant Flow Divergence Landing}
\label{sec:FixedGainControl}
In the previous section, we estimated the delay and noise in the vision measurement obtained from an on-board camera. This section investigates the effects of the delay and noise on the constant flow divergence landing using the conventional control scheme as described in Section~\ref{sec:DivGuidanceControl}. This is performed in both computer simulations and real-world experiments. 

\subsection{Simulation Results}
The same model and controller described in Section~\ref{sec:DivGuidanceControl} are used in this simulation. For a fair comparison, the settings including $D^*$, $K_p$, and the initial conditions of $Z$ and $V_Z$ are set to be the same. The control analysis of the system is performed in sequence by adding: (1) a delay, (2) a noise model, and (3) both delay and noise model into the observation model in Eq.~(\ref{equation:ModelObservation}). Their results are presented in Fig.~\ref{fig:Delay_Noise_Const_Div_Control_all}.

\subsubsection{Adding Delay}
In Section~\ref{sec:Characterization}, we estimated a delay of $2$ samples, i.e., a time delay of $0.1s$, in the flow divergence measured from the vision system. This delay is added to the observation model in the simulation. Fig.~\ref{fig:Delay_Const_Div_Control2} plots the time response of the states using the flow divergence based control. The result shows that with this delay both $Z$ and $V_Z$ converge to almost zero quicker than the response of the perfect observation, but the MAV becomes unstable when it is very close to the ground. 

\subsubsection{Adding Noise}
Next, only the noise model is added to the observation model. Fig.~\ref{fig:Noise_Const_Div_Control2} shows the effects of noise to the time response of the states. Similar to adding delay, the MAV becomes unstable when both height and velocity are very small. In practice, for both cases the oscillations can be avoided by throttling down or completely switching off the engine when the MAV is very close to the landing surface; also, in a real-world scenario a sufficiently low gain can be selected so that the MAV's landing gear touches the ground before oscillations occur.



\subsubsection{Adding Delay and Noise}
\begin{figure}
	\centering
	\begin{subfigure}[thpb]{0.4\textwidth}
%
%
\begin{psfrags}%
\psfragscanon%
\newcommand\tsize{0.6}
\newcommand{\tsizeb}{0.7}
%
\psfrag{s05}[t][t][\tsize]{\color[rgb]{0,0,0}\setlength{\tabcolsep}{0pt}\begin{tabular}{c}Time~(s)\end{tabular}}%
\psfrag{s10}[][]{\color[rgb]{0,0,0}\setlength{\tabcolsep}{0pt}\begin{tabular}{c} \end{tabular}}%
\psfrag{s11}[][]{\color[rgb]{0,0,0}\setlength{\tabcolsep}{0pt}\begin{tabular}{c} \end{tabular}}%
\psfrag{s12}[l][l][\tsize]{\color[rgb]{0,0,0}$D^*(1/s)$}%
\psfrag{s13}[l][l][\tsize]{\color[rgb]{0,0,0}$\mu(m/s^2)$}%
\psfrag{s14}[l][l][\tsize]{\color[rgb]{0,0,0}$\widehat{D}(1/s)$}%
\psfrag{s15}[l][l][\tsize]{\color[rgb]{0,0,0}$Z(m)$}%
\psfrag{s16}[l][l][\tsize]{\color[rgb]{0,0,0}$V_Z(m/s)$}%
\psfrag{s17}[l][l][\tsize]{\color[rgb]{0,0,0}$D^*(1/s)$}%
%
\psfrag{x01}[t][t][\tsizeb]{0}%
\psfrag{x02}[t][t][\tsizeb]{2}%
\psfrag{x03}[t][t][\tsizeb]{4}%
\psfrag{x04}[t][t][\tsizeb]{6}%
\psfrag{x05}[t][t][\tsizeb]{8}%
\psfrag{x06}[t][t][\tsizeb]{10}%
\psfrag{x07}[t][t][\tsizeb]{12}%
\psfrag{x08}[t][t][\tsizeb]{14}%
\psfrag{x09}[t][t][\tsizeb]{16}%
%
\psfrag{v01}[r][r][\tsizeb]{-2}%
\psfrag{v02}[r][r][\tsizeb]{-1}%
\psfrag{v03}[r][r][\tsizeb]{0}%
\psfrag{v04}[r][r][\tsizeb]{1}%
\psfrag{v05}[r][r][\tsizeb]{2}%
\psfrag{v06}[r][r][\tsizeb]{3}%
\psfrag{v07}[r][r][\tsizeb]{4}%
%
\includegraphics[width=\textwidth]{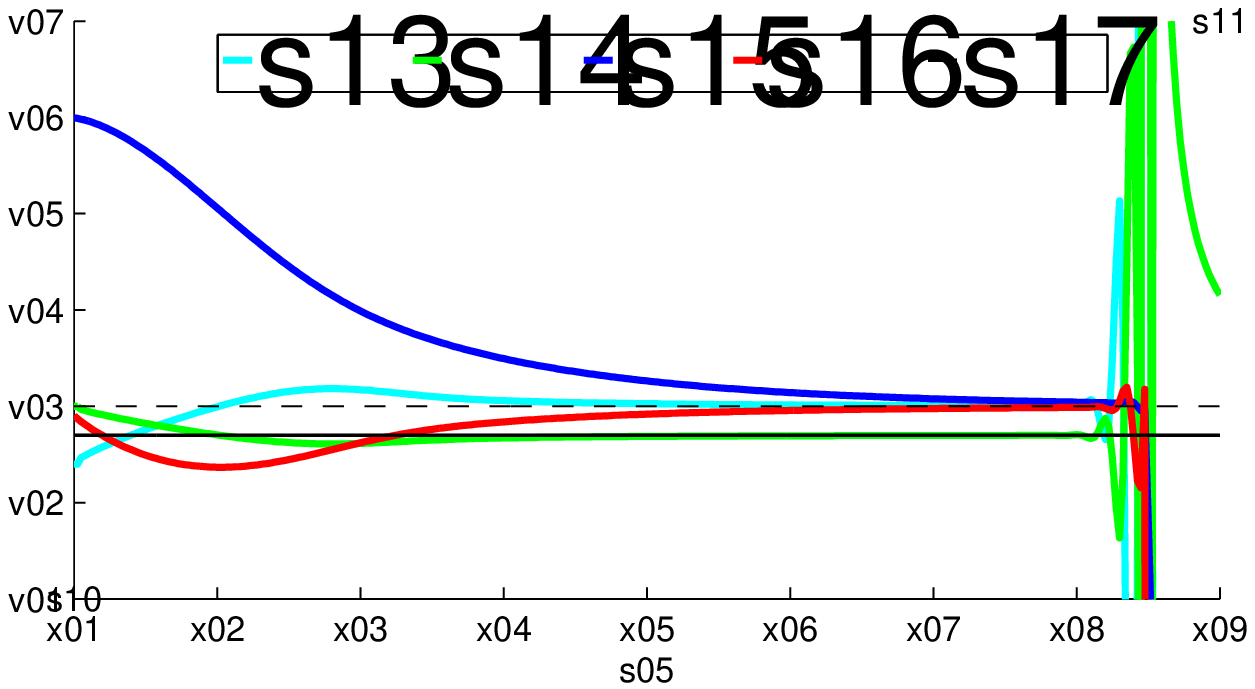}
\end{psfrags}%
%

		\caption{Adding delay.}
		\label{fig:Delay_Const_Div_Control2}
	\end{subfigure}
	\\ 
	\begin{subfigure}[thpb]{0.4\textwidth}
%
%
\begin{psfrags}%
\psfragscanon%
\newcommand{\tsize}{0.6}
\newcommand{\tsizeb}{0.7}
%
\psfrag{s05}[t][t][\tsize]{\color[rgb]{0,0,0}\setlength{\tabcolsep}{0pt}\begin{tabular}{c}Time~(s)\end{tabular}}%
\psfrag{s10}[][]{\color[rgb]{0,0,0}\setlength{\tabcolsep}{0pt}\begin{tabular}{c} \end{tabular}}%
\psfrag{s11}[][]{\color[rgb]{0,0,0}\setlength{\tabcolsep}{0pt}\begin{tabular}{c} \end{tabular}}%
\psfrag{s12}[l][l][\tsize]{\color[rgb]{0,0,0}$D^*(1/s)$}%
\psfrag{s13}[l][l][\tsize]{\color[rgb]{0,0,0}$\mu(m/s^2)$}%
\psfrag{s14}[l][l][\tsize]{\color[rgb]{0,0,0}$\widehat{D}(1/s)$}%
\psfrag{s15}[l][l][\tsize]{\color[rgb]{0,0,0}$Z(m)$}%
\psfrag{s16}[l][l][\tsize]{\color[rgb]{0,0,0}$V_Z(m/s)$}%
\psfrag{s17}[l][l][\tsize]{\color[rgb]{0,0,0}$D^*(1/s)$}%
%
\psfrag{x01}[t][t][\tsizeb]{0}%
\psfrag{x02}[t][t][\tsizeb]{2}%
\psfrag{x03}[t][t][\tsizeb]{4}%
\psfrag{x04}[t][t][\tsizeb]{6}%
\psfrag{x05}[t][t][\tsizeb]{8}%
\psfrag{x06}[t][t][\tsizeb]{10}%
\psfrag{x07}[t][t][\tsizeb]{12}%
\psfrag{x08}[t][t][\tsizeb]{14}%
\psfrag{x09}[t][t][\tsizeb]{16}%
%
\psfrag{v01}[r][r][\tsizeb]{-2}%
\psfrag{v02}[r][r][\tsizeb]{-1}%
\psfrag{v03}[r][r][\tsizeb]{0}%
\psfrag{v04}[r][r][\tsizeb]{1}%
\psfrag{v05}[r][r][\tsizeb]{2}%
\psfrag{v06}[r][r][\tsizeb]{3}%
\psfrag{v07}[r][r][\tsizeb]{4}%
%
\includegraphics[width=\textwidth]{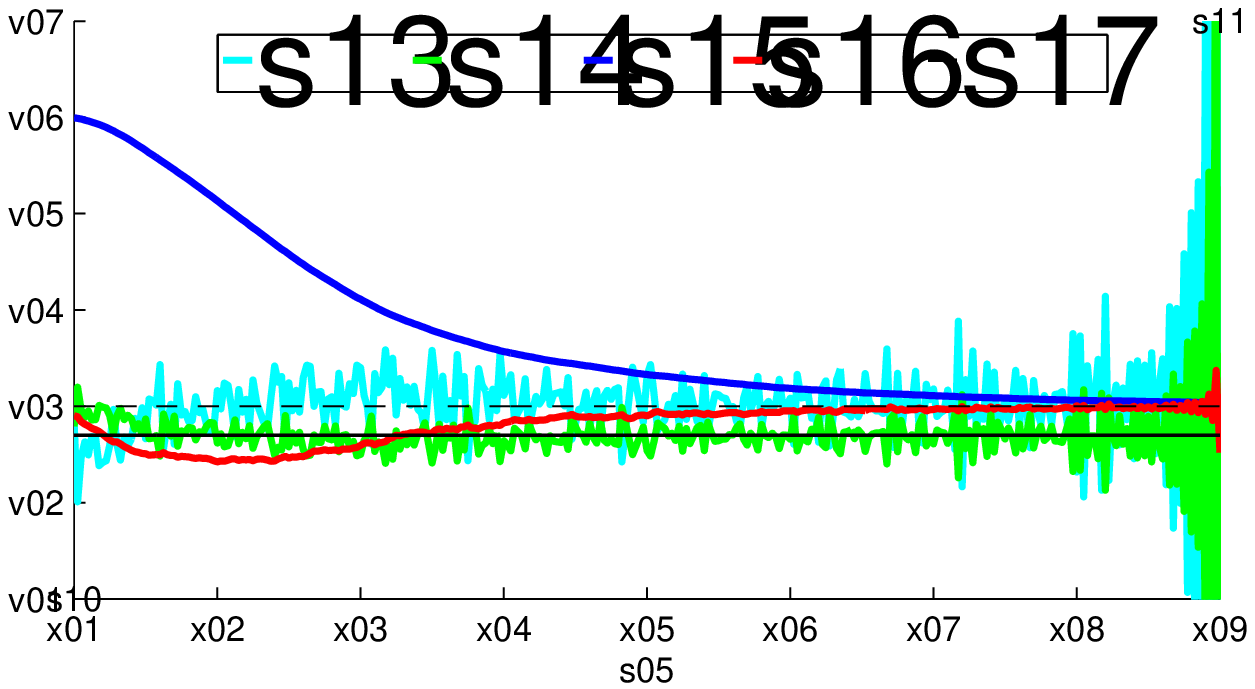}
\end{psfrags}%
%

		\caption{Adding noise model.}
		\label{fig:Noise_Const_Div_Control2}
	\end{subfigure}
	\\ 
	\begin{subfigure}[thpb]{0.4\textwidth}
%
%
\begin{psfrags}%
\psfragscanon%
\newcommand{\tsize}{0.6}
\newcommand{\tsizeb}{0.7}
%
\psfrag{s05}[t][t][\tsize]{\color[rgb]{0,0,0}\setlength{\tabcolsep}{0pt}\begin{tabular}{c}Time~(s)\end{tabular}}%
\psfrag{s10}[][]{\color[rgb]{0,0,0}\setlength{\tabcolsep}{0pt}\begin{tabular}{c} \end{tabular}}%
\psfrag{s11}[][]{\color[rgb]{0,0,0}\setlength{\tabcolsep}{0pt}\begin{tabular}{c} \end{tabular}}%
\psfrag{s12}[l][l][\tsize]{\color[rgb]{0,0,0}$D^*(1/s)$}%
\psfrag{s13}[l][l][\tsize]{\color[rgb]{0,0,0}$\mu(m/s^2)$}%
\psfrag{s14}[l][l][\tsize]{\color[rgb]{0,0,0}$\widehat{D}(1/s)$}%
\psfrag{s15}[l][l][\tsize]{\color[rgb]{0,0,0}$Z(m)$}%
\psfrag{s16}[l][l][\tsize]{\color[rgb]{0,0,0}$V_Z(m/s)$}%
\psfrag{s17}[l][l][\tsize]{\color[rgb]{0,0,0}$D^*(1/s)$}%
%
\psfrag{x01}[t][t][\tsizeb]{0}%
\psfrag{x02}[t][t][\tsizeb]{2}%
\psfrag{x03}[t][t][\tsizeb]{4}%
\psfrag{x04}[t][t][\tsizeb]{6}%
\psfrag{x05}[t][t][\tsizeb]{8}%
\psfrag{x06}[t][t][\tsizeb]{10}%
\psfrag{x07}[t][t][\tsizeb]{12}%
\psfrag{x08}[t][t][\tsizeb]{14}%
\psfrag{x09}[t][t][\tsizeb]{16}%
%
\psfrag{v01}[r][r][\tsizeb]{-2}%
\psfrag{v02}[r][r][\tsizeb]{-1}%
\psfrag{v03}[r][r][\tsizeb]{0}%
\psfrag{v04}[r][r][\tsizeb]{1}%
\psfrag{v05}[r][r][\tsizeb]{2}%
\psfrag{v06}[r][r][\tsizeb]{3}%
\psfrag{v07}[r][r][\tsizeb]{4}%
%
\includegraphics[width=\textwidth]{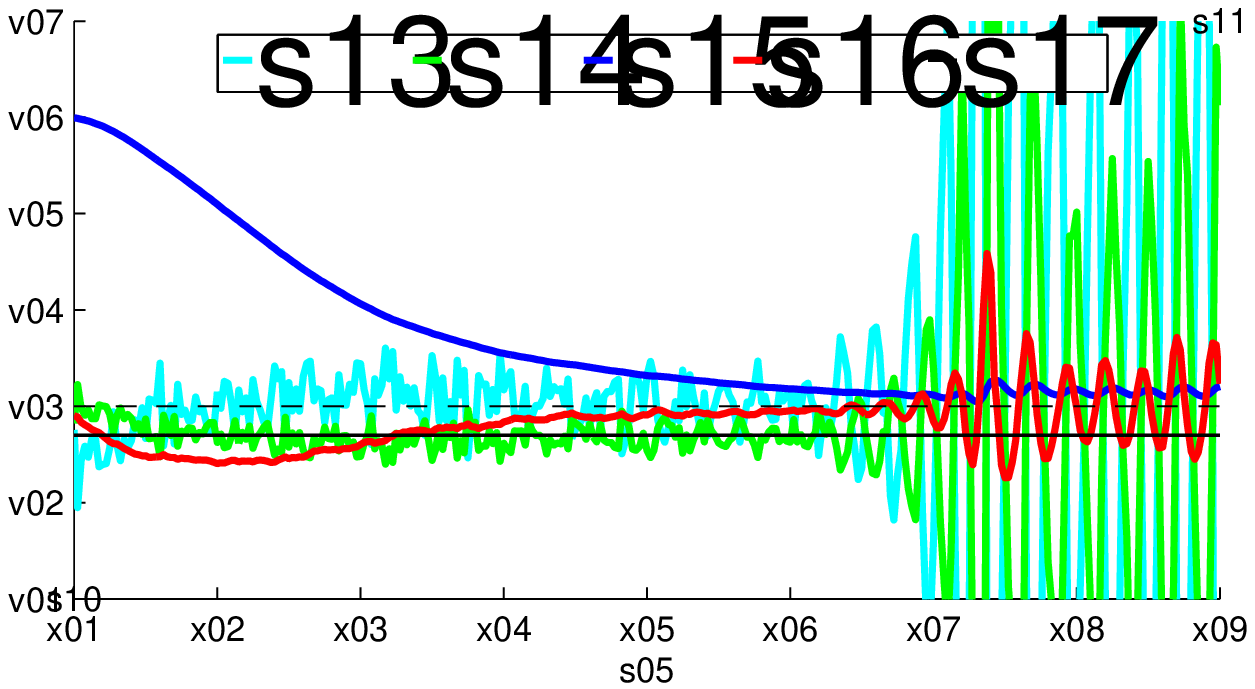}
\end{psfrags}%
%

		\caption{Adding delay and noise model.}
		\label{fig:Delay_Noise_Const_Div_Control2}
	\end{subfigure}
	\caption{Adding delay and noise to constant flow divergence landing using fixed-gain controller leads to self-induced instability at a low height.}
	\label{fig:Delay_Noise_Const_Div_Control_all}
\end{figure}

In reality, we have both delay and noise in the estimate of flow divergence from the vision system. Therefore, we also examine the effects of both delay and noise on the control scheme performance. Fig.~\ref{fig:Delay_Noise_Const_Div_Control2} shows that large oscillations occur sooner and are amplified further when the MAV moves close to the landing surface.

\subsection{Flight Test Results}
In this subsection, we present the flight test results for vertical landing controls using the conventional control scheme (e.g., a fixed gain). For the experiments, a proportional and integral (PI) controller is used to reject external disturbances and thus minimize the steady-state errors:
\begin{equation}
\mu=K_p \left[ (D^{*}-D) + \frac{1}{\kappa} \int (D^{*}-D) dt\right]
\label{equation:PIController}
\end{equation}
\noindent $\kappa = \frac{K_p}{K_i}$ is the integral time constant and $K_i$ is the integral gain.

\begin{figure}[thpb]
	\centering
%
%
\begin{psfrags}%
\psfragscanon%
\newcommand{\tsize}{0.7}
%
\psfrag{s09}[b][b][\tsize]{\color[rgb]{0,0,0}\setlength{\tabcolsep}{0pt}\begin{tabular}{c}Time~(s)\end{tabular}}%
\psfrag{s10}[t][t][\tsize]{\color[rgb]{0,0,0}\setlength{\tabcolsep}{0pt}\begin{tabular}{c}$Z~(m)$\end{tabular}}%
\psfrag{s11}[b][b][\tsize]{\color[rgb]{0,0,0}\setlength{\tabcolsep}{0pt}\begin{tabular}{c}Time~(s)\end{tabular}}%
\psfrag{s12}[t][t][\tsize]{\color[rgb]{0,0,0}\setlength{\tabcolsep}{0pt}\begin{tabular}{c}$\widehat{D}~(1/s)$\end{tabular}}%
\psfrag{s13}[b][b][\tsize]{\color[rgb]{0,0,0}\setlength{\tabcolsep}{0pt}\begin{tabular}{c}Time~(s)\end{tabular}}%
\psfrag{s14}[t][t][\tsize]{\color[rgb]{0,0,0}\setlength{\tabcolsep}{0pt}\begin{tabular}{c}$V_Z~(m/s)$\end{tabular}}%
\psfrag{s15}[b][b][\tsize]{\color[rgb]{0,0,0}\setlength{\tabcolsep}{0pt}\begin{tabular}{c}Time~(s)\end{tabular}}%
\psfrag{s16}[t][t][\tsize]{\color[rgb]{0,0,0}\setlength{\tabcolsep}{0pt}\begin{tabular}{c}$\mu~(m/s^2)$\end{tabular}}%
%
\psfrag{x01}[t][t][\tsize]{0}%
\psfrag{x02}[t][t][\tsize]{2}%
\psfrag{x03}[t][t][\tsize]{4}%
\psfrag{x04}[t][t][\tsize]{6}%
\psfrag{x05}[t][t][\tsize]{8}%
\psfrag{x06}[t][t][\tsize]{10}%
\psfrag{x07}[t][t][\tsize]{12}%
\psfrag{x08}[t][t][\tsize]{14}%
\psfrag{x09}[t][t][\tsize]{16}%
\psfrag{x10}[t][t][\tsize]{18}%
\psfrag{x11}[t][t][\tsize]{20}%
\psfrag{x12}[t][t][\tsize]{0}%
\psfrag{x13}[t][t][\tsize]{2}%
\psfrag{x14}[t][t][\tsize]{4}%
\psfrag{x15}[t][t][\tsize]{6}%
\psfrag{x16}[t][t][\tsize]{8}%
\psfrag{x17}[t][t][\tsize]{10}%
\psfrag{x18}[t][t][\tsize]{12}%
\psfrag{x19}[t][t][\tsize]{14}%
\psfrag{x20}[t][t][\tsize]{16}%
\psfrag{x21}[t][t][\tsize]{18}%
\psfrag{x22}[t][t][\tsize]{20}%
\psfrag{x23}[t][t][\tsize]{0}%
\psfrag{x24}[t][t][\tsize]{2}%
\psfrag{x25}[t][t][\tsize]{4}%
\psfrag{x26}[t][t][\tsize]{6}%
\psfrag{x27}[t][t][\tsize]{8}%
\psfrag{x28}[t][t][\tsize]{10}%
\psfrag{x29}[t][t][\tsize]{12}%
\psfrag{x30}[t][t][\tsize]{14}%
\psfrag{x31}[t][t][\tsize]{16}%
\psfrag{x32}[t][t][\tsize]{18}%
\psfrag{x33}[t][t][\tsize]{20}%
\psfrag{x34}[t][t][\tsize]{0}%
\psfrag{x35}[t][t][\tsize]{2}%
\psfrag{x36}[t][t][\tsize]{4}%
\psfrag{x37}[t][t][\tsize]{6}%
\psfrag{x38}[t][t][\tsize]{8}%
\psfrag{x39}[t][t][\tsize]{10}%
\psfrag{x40}[t][t][\tsize]{12}%
\psfrag{x41}[t][t][\tsize]{14}%
\psfrag{x42}[t][t][\tsize]{16}%
\psfrag{x43}[t][t][\tsize]{18}%
\psfrag{x44}[t][t][\tsize]{20}%
%
\psfrag{v01}[r][r][\tsize]{-0.4}%
\psfrag{v02}[r][r][\tsize]{}%
\psfrag{v03}[r][r][\tsize]{-0.2}%
\psfrag{v04}[r][r][\tsize]{}%
\psfrag{v05}[r][r][\tsize]{0}%
\psfrag{v06}[r][r][\tsize]{}%
\psfrag{v07}[r][r][\tsize]{0.2}%
\psfrag{v08}[r][r][\tsize]{}%
\psfrag{v09}[r][r][\tsize]{-0.8}%
\psfrag{v10}[r][r][\tsize]{}%
\psfrag{v11}[r][r][\tsize]{-0.4}%
\psfrag{v12}[r][r][\tsize]{}%
\psfrag{v13}[r][r][\tsize]{0}%
\psfrag{v14}[r][r][\tsize]{}%
\psfrag{v15}[r][r][\tsize]{0.4}%
\psfrag{v16}[r][r][\tsize]{}%
\psfrag{v17}[r][r][\tsize]{}%
\psfrag{v18}[r][r][\tsize]{-0.4}%
\psfrag{v19}[r][r][\tsize]{}%
\psfrag{v20}[r][r][\tsize]{0}%
\psfrag{v21}[r][r][\tsize]{}%
\psfrag{v22}[r][r][\tsize]{0.4}%
\psfrag{v23}[r][r][\tsize]{0}%
\psfrag{v24}[r][r][\tsize]{}%
\psfrag{v25}[r][r][\tsize]{1}%
\psfrag{v26}[r][r][\tsize]{}%
\psfrag{v27}[r][r][\tsize]{2}%
\psfrag{v28}[r][r][\tsize]{}%
\psfrag{v29}[r][r][\tsize]{3}%
\psfrag{v30}[r][r][\tsize]{}%
\psfrag{v31}[r][r][\tsize]{4}%
%
\includegraphics[width=0.5\textwidth]{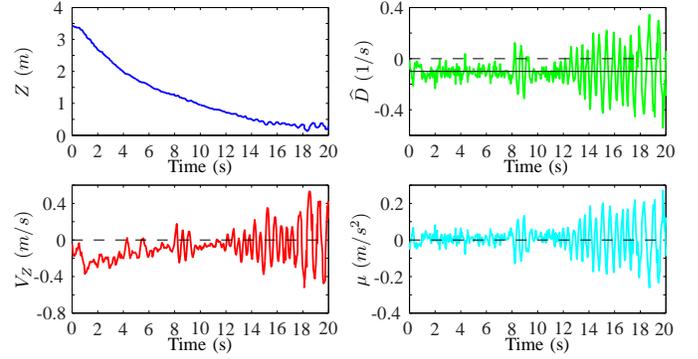}
\end{psfrags}%
%

	\caption{Constant flow divergence landing of the MAV using a fixed-gain controller.}
	\label{fig:fixed_gain_size_divergence_PI}
\end{figure}

The same MAV platform as described in Section~\ref{subsec:TestingPlatform} is used for the experiments with all vision and control algorithms running on-board. Fig.~\ref{fig:fixed_gain_size_divergence_PI} shows the experiment results of constant flow divergence landing using the basic, fixed-gain, controller (e.g., $K_p=0.6, K_i=0.1$). Clearly, large oscillations occur for flow divergence, height, and velocity due to the delay and noise of the flow divergence estimate, even though the height and velocity exponentially decay to zero. We used a slightly smaller desired flow divergence than the one used in the simulations, i.e., $D^*=0.1s^{-1}$ in order to obtain a better view of the oscillations before touching the ground surface. 

\section{Adaptive Gain Strategy for Constant Divergence Landing}
\label{sec:AdaptiveGainControl}
In the previous section, we observed that the presence of time delay and measurement noise leads to instability of constant flow divergence landings when the basic, fixed-gain, controller is used. The oscillations can be further amplified when the vehicle is getting closer to the ground or can even happen at an earlier stage of the landing when a large gain is used. This section introduces a novel way to reject oscillations in constant flow divergence landings. 

\subsection{Adaptive Control Strategy}
\label{subsec:AdaptiveControlStrategy}
\begin{figure}[thpb]
	\centering
	\begin{psfrags}%
\psfragscanon%
\newcommand{\tsize}{0.7}
%
\psfrag{a}[c][c][\tsize]{\color[rgb]{0,0,0}\setlength{\tabcolsep}{0pt}\begin{tabular}{c}$Z$\end{tabular}}%
\psfrag{b}[c][c][\tsize]{\color[rgb]{0,0,0}\setlength{\tabcolsep}{0pt}\begin{tabular}{c}$Time$\end{tabular}}%
\psfrag{c}[c][c][\tsize]{\color[rgb]{0,0,0}\setlength{\tabcolsep}{0pt}\begin{tabular}{c}$I$\end{tabular}}%
\psfrag{d}[c][c][\tsize]{\color[rgb]{0,0,0}\setlength{\tabcolsep}{0pt}\begin{tabular}{c}$II$\end{tabular}}%

\includegraphics[width=0.4\textwidth]{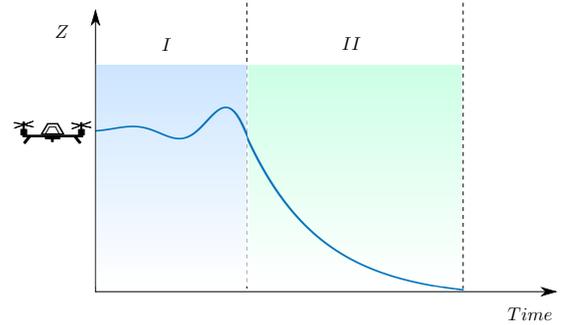}
\end{psfrags}%
%

	\caption{Two-phase adaptive control strategy for constant flow divergence landing. In phase $I$, the MAV increases the gain until it starts to oscillate. This allows the MAV to select a suitable gain for phase $II$, in which it lands while reducing the gain exponentially. }
	\label{fig:adaptive_mechanism_intro}
\end{figure}

Fig.~\ref{fig:adaptive_mechanism_intro} shows the proposed adaptive control strategy for constant divergence landings. There are two phases in this strategy: ($I$) Determination of near-optimal initial controller gains, and ($II$) Landing with an adaptive gain. 

\subsubsection{Determination of Near-Optimal Initial Gains}
\label{subsubsec:InitialGain}
The initial controller gains $K_0$ are the important parameters that we need to determine for the PI controller before starting the landing. If these initial values are too small, tracking of $D^*$ will not be accurate. In contrast, if they are too large, self-induced oscillations can happen at the beginning of the landing. To deal with both cases, the MAV hovers by tracking $D^*=0$ using a PI controller while small initial values of $K_0$, which are gradually increased until an oscillation is detected. Then, the stable gain just \emph{before} the oscillation is used as the initial $K_0$ for phase $II$. Both gains from P and I controllers undergo this process.

\subsubsection{Landing}
\label{subsubsec:OscillationRejection}
Once the stable initial gains are obtained, the constant flow divergence landing is activated by tracking $D^*=-k$. We know that the value of the gain at which instability starts to occur depends linearly on the height \cite{de2016monocular}, and that during a constant divergence landing the height decays exponentially. Therefore, we have the gains decay exponentially to mitigate and, if possible, eliminate the oscillations when moving close to the ground. In this strategy, phase $I$ ensures that proper initial gains are chosen to have a good performance of the tracking, while phase $II$ prevents self-induced oscillations when descending. In the following subsections, the stability analysis of the adaptive controller and the real-time oscillation detection method used in this strategy are described in details. 

\subsection{Stability Analysis of the Adaptive Controller}
\label{subsec:StabilityAnalysis}
In this subsection, we show that the linearized system is not subject to self-induced oscillations when the adaptive controller is used for constant flow divergence landings. In fact, we know from Eq.~(\ref{equation:ZConstantD}) that $Z = Z_0e^{D^*t}$ when $D=D^*$.  As mentioned, to cope with the instability problem, we introduce:
\begin{equation}
	K_p=K_{p_0} e^{D^* t},\quad K_i=K_{i_0} e^{D^* t}
	\label{equation:adaptivegain}
\end{equation}
\noindent where $K_{p_0}$ and $K_{i_0}$ are the initial gains of the PI controller which relates to the initial height ($K_0 = f(Z_0)$) and can be obtained using the method presented in Subsection~\ref{subsubsec:InitialGain}. By recalling Eqs.~(\ref{equation:ModelDynamics}) and (\ref{equation:ModelObservation}), $\dot{\mathbf{x}} = [x_2, \mu]^T$ and $y(t)=[x_2/x_1]$, where $\mathbf{x}=[x_1,x_2]^T=[Z,V_Z]^T$. 

\begin{figure}[thpb]
	\centering
%
%
\begin{psfrags}%
\psfragscanon%
\newcommand{\tsize}{0.8}
%
\psfrag{s01}[t][t][\tsize]{\color[rgb]{0.15,0.15,0.15}\setlength{\tabcolsep}{0pt}\begin{tabular}{c}$V_Z~(m/s)$\end{tabular}}%
\psfrag{s02}[b][b][\tsize]{\color[rgb]{0.15,0.15,0.15}\setlength{\tabcolsep}{0pt}\begin{tabular}{c}$Z~(m)$\end{tabular}}%
%
\color[rgb]{0.15,0.15,0.15}%
%
\psfrag{x01}[t][t][\tsize]{-5}%
\psfrag{x02}[t][t][\tsize]{0}%
\psfrag{x03}[t][t][\tsize]{5}%
%
\psfrag{v01}[r][r][\tsize]{0}%
\psfrag{v02}[r][r][\tsize]{2}%
\psfrag{v03}[r][r][\tsize]{4}%
\psfrag{v04}[r][r][\tsize]{6}%
\psfrag{v05}[r][r][\tsize]{8}%
\psfrag{v06}[r][r][\tsize]{10}%
\psfrag{v07}[r][r][\tsize]{12}%
%
\includegraphics[trim = 8mm 0mm 0mm 0mm, clip, width=0.35\textwidth]{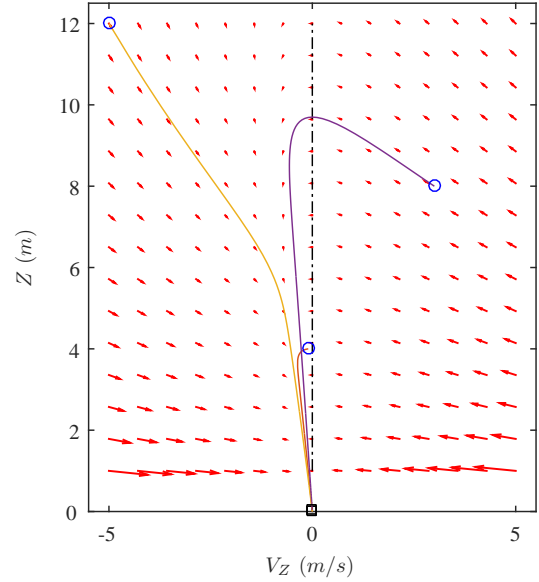}
\end{psfrags}%
%

	\caption{Phase portrait of the constant flow divergence landing.}
	\label{fig:phase_portrait_Kp1_Dsp_01}
\end{figure}

To understand the dynamical behavior of this system, we first analyze the phase portrait of the system. Fig.~\ref{fig:phase_portrait_Kp1_Dsp_01} shows the system's trajectories with arrows and three cases with different initial states. All states of these cases converge to zero in the end. Most importantly, we can observe that positive $V_Z$, which could happen due to external disturbances, will eventually become negative (i.e. the MAV descends). This can also be seen from Eq.~(\ref{equation:PIController}) that when $V_Z$ becomes positive, the controller will further reduce the thrust and lead to $V_Z<0$.


Here, we study the stability of the discrete system. We will see that even introducing just a zero-order hold form in which it has a discrete sample time and thus a small delay in the system already suffices to get self-induced instability. By linearizing and discretizing the system model, we obtain:
\begin{equation}
	\Phi = \begin{bmatrix}
		1 & \Delta t \\
		0 & 1
	\end{bmatrix}, \quad
	\Gamma = \begin{bmatrix}
		\frac{\Delta t^2}{2} \\
		\Delta t
	\end{bmatrix}, \quad
	C = \begin{bmatrix}
		-\frac{V_Z}{Z^2} & \frac{1}{Z}
	\end{bmatrix}, \quad
	D = [0].
	\label{equation:LinearziedDiscreteModel}
\end{equation}
\noindent The open-loop transfer function of the discrete system can be determined to be:
\begin{equation}
	\begin{split}
		G(\sigma) & = C\left(\sigma I - \Phi\right)^{-1}\Gamma\\
		& = \frac{\Delta t\left[\left( 2Z-\Delta tV_Z \right)\sigma-\left( 2Z+\Delta tV_Z \right)\right]}{2Z^2\left(\sigma-1\right)^2},
	\end{split}
	\label{equation:DiscreteOLTransferFunction}
\end{equation}
\noindent where $\sigma$ is the discrete frequency domain operator (variable $z$ typically used for this term could be confused with the height $Z$, thus $\sigma$ is used to avoid confusion).

\begin{figure}[thpb]
	\centering
%
%
\begin{psfrags}%
\psfragscanon%
\newcommand{\tsize}{0.7}
%
\psfrag{s01}[t][t][\tsize]{\color[rgb]{0.15,0.15,0.15}\setlength{\tabcolsep}{0pt}\begin{tabular}{c}Real Axis\end{tabular}}%
\psfrag{s02}[b][b][\tsize]{\color[rgb]{0.15,0.15,0.15}\setlength{\tabcolsep}{0pt}\begin{tabular}{c}Imaginary Axis\end{tabular}}%
\psfrag{s03}[l][l][\tsize]{\color[rgb]{0,0,0}\setlength{\tabcolsep}{0pt}\begin{tabular}{l}$K_{Cr} = 6667, Z = 100m$\end{tabular}}%
\psfrag{s04}[l][l][\tsize]{\color[rgb]{0.2,0.2,0.2}\setlength{\tabcolsep}{0pt}\begin{tabular}{l}$K_{Cr} = 667, Z = 10m$\end{tabular}}%
%
\color[rgb]{0.15,0.15,0.15}%
%
\psfrag{x01}[t][t][\tsize]{}%
\psfrag{x02}[t][t][\tsize]{-1}%
\psfrag{x03}[t][t][\tsize]{}%
\psfrag{x04}[t][t][\tsize]{0}%
\psfrag{x05}[t][t][\tsize]{}%
\psfrag{x06}[t][t][\tsize]{1}%
\psfrag{x07}[t][t][\tsize]{}%
%
\psfrag{v01}[r][r][\tsize]{-1}%
\psfrag{v02}[r][r][\tsize]{}%
\psfrag{v03}[r][r][\tsize]{}%
\psfrag{v04}[r][r][\tsize]{}%
\psfrag{v05}[r][r][\tsize]{}%
\psfrag{v06}[r][r][\tsize]{0}%
\psfrag{v07}[r][r][\tsize]{}%
\psfrag{v08}[r][r][\tsize]{}%
\psfrag{v09}[r][r][\tsize]{}%
\psfrag{v10}[r][r][\tsize]{}%
\psfrag{v11}[r][r][\tsize]{1}%
%
\includegraphics[trim = 0mm 0mm 0mm -5mm, clip, width=0.3\textwidth]{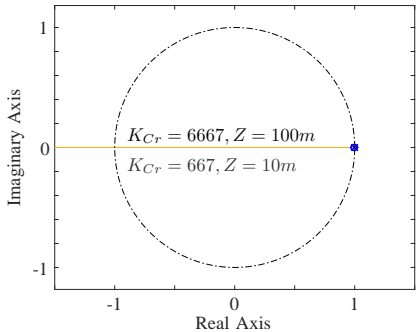}
\end{psfrags}%
%

	\caption{Root locus of the closed-loop discrete system for two different heights.}
	\label{fig:rootlocus_PI}
\end{figure}

\noindent The discrete form of the PI controller can be written as:
\begin{equation}
	F_{PI}(\sigma) = K_p \left[ 1 + \frac{\Delta t}{2 \kappa} \left( \frac{\sigma+1}{\sigma-1} \right) \right].
	\label{equation:DiscretePIController}
\end{equation}
The closed-loop transfer function of the discrete system is:
\begin{equation}
	\begin{split}
		H(\sigma) &= \frac{G(\sigma) \cdot F_{PI}(\sigma)}{1+G(\sigma) \cdot F_{PI}(\sigma)}\\
		& = \frac{\mathcal{N}(\sigma)}{\mathcal{D}(\sigma)},
	\end{split}
	\label{equation:DiscreteCLTransferFunction}
\end{equation}
where
\begin{align*}
\mathcal{N}(\sigma) = & K_p \Delta t \left[ (\Delta t- 2\kappa) + (\Delta t+2\kappa)\sigma \right] \left[(\Delta tV_Z+2Z) \right. \\
& + \left. (\Delta tV_Z-2Z)\sigma\right],           \\
\mathcal{D}(\sigma) = & K_p \Delta t \left[ V_Z \Delta t^2 (\sigma+1)^2 + 2\Delta t(\kappa V_Z - Z)(\sigma^2 - 1) \right. \\
& - \left. 4Z\kappa(\sigma-1)^2 \right]-4\kappa Z^2(\sigma-1)^3.
\end{align*}
The zeros of the system can be obtained to be:
\begin{equation}
	\sigma_{0_1} = \frac{2Z+\Delta tV_Z}{2Z-\Delta tV_Z} \quad \text{or} \quad \sigma_{0_2} = \frac{2\kappa-\Delta t}{2\kappa+\Delta t}.
	\label{equation:ZerosDiscrete}
\end{equation}
We know that due to a relatively small and positive $\Delta t$, $Z>0$ and $V_Z<0$, thus $0<\sigma_{0_1}<1$. In contrast, $\sigma_{0_2}$ depends on $\kappa$. For a stable discrete system, all poles should lie inside a unit circle in $\sigma$-plane. From Eq.~(\ref{equation:DiscreteCLTransferFunction}), we know that all poles are located at $\sigma=1$ when $K_p=0$. As the gain increases, two poles move toward the two finite zeros presented in Eq.~(\ref{equation:ZerosDiscrete}) and the third pole moves toward the negative infinite zero. From this observation, there are two factors which can affect the stability of the system: (1) the gain at $\sigma=-1$, the so-called critical P-gain, $K_{cr}$, and (2) the influence of $\kappa$ on $\sigma_{0_2}$. For the reader to understand the first case, we plot a root locus of the closed-loop discrete system for $Z=100m$ and $Z=10m$ with $\Delta t=0.03s$ in Fig.~\ref{fig:rootlocus_PI}. In this figure, both results of the different heights lead to the same root locus plot, but with different values of the gain (see Eq.~(\ref{equation:CriticalGain})).
Let $\sigma=-1$ in Eq.~(\ref{equation:DiscreteCLTransferFunction}), we obtain:
\begin{equation}
	K_{cr} = \frac{2Z}{\Delta t}.
	\label{equation:CriticalGain}
\end{equation}
This relation is exactly the same as the one found in \cite{de2016monocular} for a pure P-controller. It shows that the critical gain depends on the sample time $\Delta t$, and - importantly - on the height $Z$. From this result, the controller gain should be $0<K_p<K_{cr}$ in order to have a stable system. If $K_0$ is set to be $<2Z/\Delta t$ and $K_p$ scales with the same exponent as $Z$, it will stay below that threshold for the rest of the trajectory.

\begin{figure*}
	\centering
	\begin{subfigure}[thpb]{0.3\textwidth}
%
%
\begin{psfrags}%
\psfragscanon%
\newcommand{\tsize}{0.7}
\newcommand{\tsizeb}{0.7}
%
\psfrag{s01}[t][b][\tsize]{\color[rgb]{0.15,0.15,0.15}\setlength{\tabcolsep}{0pt}\begin{tabular}{c}Real Axis\end{tabular}}%
\psfrag{s02}[b][b][\tsize]{\color[rgb]{0.15,0.15,0.15}\setlength{\tabcolsep}{0pt}\begin{tabular}{c}Imaginary Axis\end{tabular}}%
\psfrag{s03}[l][l][\tsize]{\color[rgb]{0,0,0}\setlength{\tabcolsep}{0pt}\begin{tabular}{l}$\kappa = 3$\end{tabular}}%
%
\color[rgb]{0.15,0.15,0.15}%
%
\psfrag{x01}[t][t][\tsizeb]{0.98}%
\psfrag{x02}[t][t][\tsizeb]{0.99}%
\psfrag{x03}[t][t][\tsizeb]{1}%
\psfrag{x04}[t][t][\tsize]{}%
\psfrag{x05}[t][t][\tsize]{-1}%
\psfrag{x06}[t][t][\tsize]{}%
\psfrag{x07}[t][t][\tsize]{0}%
\psfrag{x08}[t][t][\tsize]{}%
\psfrag{x09}[t][t][\tsize]{1}%
\psfrag{x10}[t][t][\tsize]{}%
\psfrag{x11}[t][t][\tsize]{2}%
\psfrag{x12}[t][t][\tsize]{}%
%
\psfrag{v01}[r][r][\tsize]{-5}%
\psfrag{v02}[r][r][\tsize]{0}%
\psfrag{v03}[r][r][\tsize]{5}%
\psfrag{ypower1}[Bl][Bl][\tsize]{$\times 10^{-3}$}%
\psfrag{v04}[r][r][\tsize]{}%
\psfrag{v05}[r][r][\tsize]{-1}%
\psfrag{v06}[r][r][\tsize]{}%
\psfrag{v07}[r][r][\tsize]{0}%
\psfrag{v08}[r][r][\tsize]{}%
\psfrag{v09}[r][r][\tsize]{1}%
\psfrag{v10}[r][r][\tsize]{}%
%
\includegraphics[width=\textwidth]{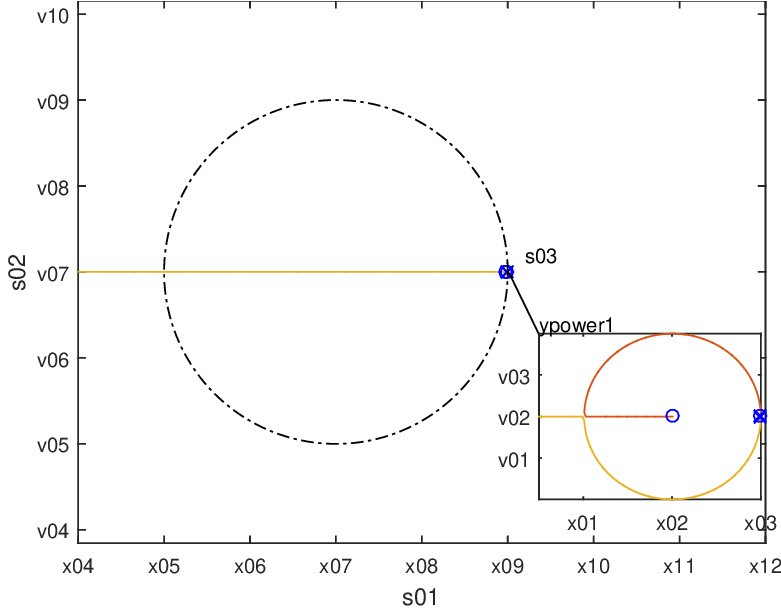}
\end{psfrags}%
%

		\caption{$\kappa>\frac{\Delta t}{2}$.}
		\label{fig:rootlocus_border}
	\end{subfigure}
	\quad 
	\begin{subfigure}[thpb]{0.3\textwidth}
%
%
\begin{psfrags}%
\psfragscanon%
\newcommand{\tsize}{0.7}
%
\psfrag{s01}[t][b][\tsize]{\color[rgb]{0.15,0.15,0.15}\setlength{\tabcolsep}{0pt}\begin{tabular}{c}Real Axis\end{tabular}}%
\psfrag{s02}[b][b][\tsize]{\color[rgb]{0.15,0.15,0.15}\setlength{\tabcolsep}{0pt}\begin{tabular}{c}Imaginary Axis\end{tabular}}%
\psfrag{s03}[l][l][\tsize]{\color[rgb]{0,0,0}\setlength{\tabcolsep}{0pt}\begin{tabular}{l}$\kappa = 0.015$\end{tabular}}%
%
\color[rgb]{0.15,0.15,0.15}%
%
\psfrag{x01}[t][t][\tsize]{}%
\psfrag{x02}[t][t][\tsize]{-1}%
\psfrag{x03}[t][t][\tsize]{}%
\psfrag{x04}[t][t][\tsize]{0}%
\psfrag{x05}[t][t][\tsize]{}%
\psfrag{x06}[t][t][\tsize]{1}%
\psfrag{x07}[t][t][\tsize]{}%
\psfrag{x08}[t][t][\tsize]{2}%
\psfrag{x09}[t][t][\tsize]{}%
%
\psfrag{v01}[r][r][\tsize]{}%
\psfrag{v02}[r][r][\tsize]{-1}%
\psfrag{v03}[r][r][\tsize]{}%
\psfrag{v04}[r][r][\tsize]{0}%
\psfrag{v05}[r][r][\tsize]{}%
\psfrag{v06}[r][r][\tsize]{1}%
\psfrag{v07}[r][r][\tsize]{}%
%
\includegraphics[width=\textwidth]{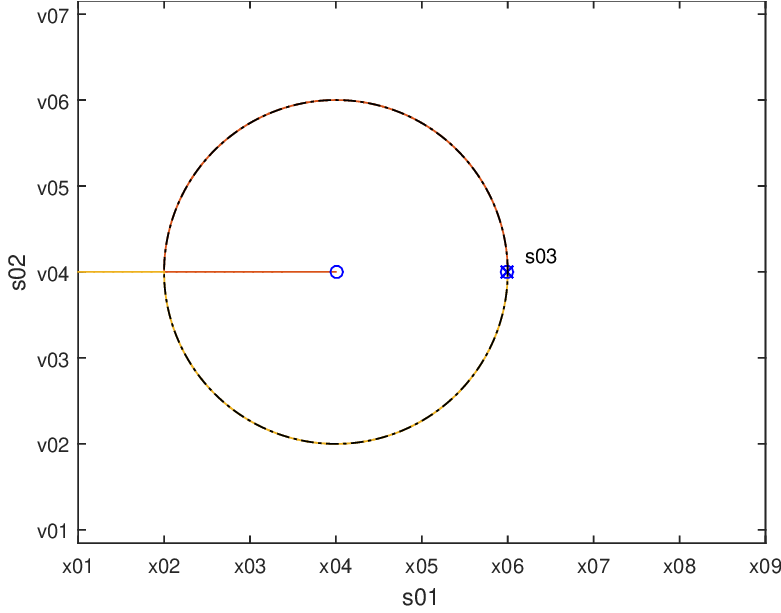}
\end{psfrags}%
%

		\caption{$\kappa=\frac{\Delta t}{2}$.}
		\label{fig:rootlocus_unstable}
	\end{subfigure}
	\quad 
	\begin{subfigure}[thpb]{0.3\textwidth}
%
%
\begin{psfrags}%
\psfragscanon%
\newcommand{\tsize}{0.7}
%
\psfrag{s01}[t][b][\tsize]{\color[rgb]{0.15,0.15,0.15}\setlength{\tabcolsep}{0pt}\begin{tabular}{c}Real Axis\end{tabular}}%
\psfrag{s02}[b][b][\tsize]{\color[rgb]{0.15,0.15,0.15}\setlength{\tabcolsep}{0pt}\begin{tabular}{c}Imaginary Axis\end{tabular}}%
\psfrag{s03}[l][l][\tsize]{\color[rgb]{0,0,0}\setlength{\tabcolsep}{0pt}\begin{tabular}{l}$\kappa = 0.01$\end{tabular}}%
%
\color[rgb]{0.15,0.15,0.15}%
%
\psfrag{x01}[t][t][\tsize]{}%
\psfrag{x02}[t][t][\tsize]{-1}%
\psfrag{x03}[t][t][\tsize]{}%
\psfrag{x04}[t][t][\tsize]{0}%
\psfrag{x05}[t][t][\tsize]{}%
\psfrag{x06}[t][t][\tsize]{1}%
\psfrag{x07}[t][t][\tsize]{}%
\psfrag{x08}[t][t][\tsize]{2}%
\psfrag{x09}[t][t][\tsize]{}%
%
\psfrag{v01}[r][r][\tsize]{}%
\psfrag{v02}[r][r][\tsize]{-1}%
\psfrag{v03}[r][r][\tsize]{}%
\psfrag{v04}[r][r][\tsize]{0}%
\psfrag{v05}[r][r][\tsize]{}%
\psfrag{v06}[r][r][\tsize]{1}%
\psfrag{v07}[r][r][\tsize]{}%
%
\includegraphics[width=\textwidth]{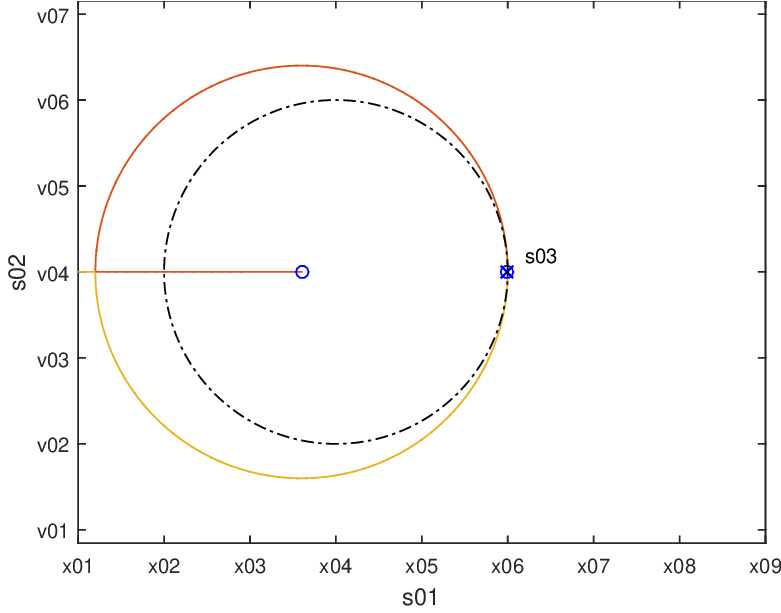}
\end{psfrags}%
%

		\caption{$\kappa<\frac{\Delta t}{2}$.}
		\label{fig:rootlocus_typical}
	\end{subfigure}
	\caption{Root locus of the closed-loop discrete system for three different integral time constants.}
	\label{fig:rootlocusIntegralTime}
\end{figure*}

For the second case, we investigate the influence of $\kappa$ on $\sigma_{0_2}$, and thus the stability of the system. Fig.~\ref{fig:rootlocusIntegralTime} shows three root loci of the closed-loop discrete system for three different values of $\kappa$. 
As mentioned above, one pole moves to $\sigma_{0_2}$, and another pole moves to negative infinite zero. We will prove that a part of this root locus is a circle with center located at $\sigma_{0_2}$. Since we know that these two poles are $1$ and the finite zero is $\sigma_{0_2}$, we can write the system characteristic equation for this specific case to be approximately \cite{ganesh2010control}:
\begin{equation}
	1+\frac{K_p\left(\sigma-\sigma_{0_2}\right)}{\left(\sigma-1\right)^2} = 0
	\label{equation:CharPolyZero2}
\end{equation}
By substituting $\sigma=\varrho+j\chi$ where $\varrho$ and $\chi$ are the real and imaginary part of $\sigma$ into Eq.~(\ref{equation:CharPolyZero2}), we obtain:
\begin{equation}
	\frac{K_p\left(\varrho-\sigma_{0_2}+j\chi\right)}{\left(\varrho-1+j\chi\right)^2} = -1
	\label{equation:CharPolyZero2ReIm}
\end{equation}
We know that for every point on root locus, the angle condition must be satisfied:
\begin{equation}
	\textrm{tan}^{-1}\left(\frac{\chi}{\varrho-\sigma_{0_2}}\right)-\textrm{tan}^{-1}\left(\frac{\chi}{\varrho-1}\right)-\textrm{tan}^{-1}\left(\frac{\chi}{\varrho-1}\right) = 180^o
	\label{equation:AngleCondition}
\end{equation}
By solving Eq.~(\ref{equation:AngleCondition}), we get the following equation:
\begin{equation}
	\left(\varrho-\sigma_{0_2}\right)^2+\chi^2 = \left(1-\sigma_{0_2}\right)^2
	\label{equation:Circle}
\end{equation}
Eq.~(\ref{equation:Circle}) shows that this part of the root locus is a circle centered at $(\sigma_{0_2},0)$ with radius $1-\sigma_{0_2}$ in the $\sigma$-plane. Since this circle is tangent to the unit circle at $1$, it will coincide with the unit circle if $\sigma_{0,2} = 0$. From this result, the system is stable if $\sigma_{0,2} \geq 0$ and hence $\kappa \geq \frac{\Delta t}{2}$ obtained from Eq.~(\ref{equation:ZerosDiscrete}). In fact, in practice, the value of $\kappa$ is usually set to be much greater than $1$, meaning that $K_p>>K_i$. Still, since $K_p$ has to decrease when going down, it means that $K_i$ needs to decrease exponentially as well in order to prevent $K_i>K_p$ and keep $\kappa$ constant.

To summarize, for the discrete system which possesses a small delay, instability of the closed-loop system will happen if $K_p>K_{cr}$ or $\kappa<\Delta t/2$. With the proposed adaptive controller shown in Eq.~(\ref{equation:adaptivegain}), $K_p$ and $K_i$ are kept small enough to prevent self-induced instability, while always being as high as possible to maximize control performance.

\subsection{Real-Time Oscillations Detection}
\label{subsec:RealTimeOscillationDetection}
In the first phase of our landing strategy, the MAV has to increase its control gains until it starts to self-induced oscillations. In addition, in phase $II$ it can happen that the MAV is not able to keep the flow divergence constant, and that it descends too fast. In that case, the exponentially decreasing gains $K_p$ and $K_i$ may cause self-induced oscillations that have to be detected and dealt with (by resetting the gains to appropriate values). For both these cases, we need a method to detect self-induced oscillations experienced by the vehicle in real-time. 

There are a few methods in the literature to detect self-induced oscillations, typically relying on a fast Fourier transform (FFT) \cite{chowdhary2011frequency, rzucidlo2007detection}. The reason an FFT is used, is that self-induced oscillations are a ``resonance'' property of the control system and hence have a typical frequency. However, FFT methods are computationally expensive. Therefore, we detect self-induced oscillations by examining the covariance function of a windowed flow divergence $\widehat{D}$ and a time shifted windowed flow divergence $\widehat{D}'$:
\begin{equation}
	cov(\widehat{D},\widehat{D}')=E[(\widehat{D}-E(\widehat{D}))(\widehat{D}'-E(\widehat{D}')^T)],
	\label{equation:covariance}
\end{equation}
where $E(\widehat{*})$ is the expected value of windowed flow divergence. The covariance is chosen, since it is much faster to compute than an FFT, while capturing both the relative phase and the magnitude of deviations in the signal. The ``price paid'' is that it will only react to a single frequency, but this is exactly what we want for the detection of self-induced oscillations.

\begin{figure}[thpb]
	\centering
%
%
\begin{psfrags}%
\psfragscanon%
\newcommand{\tsize}{0.7}
%
\psfrag{a}[t][t][\tsize]{\color[rgb]{0,0,0}\setlength{\tabcolsep}{0pt}\begin{tabular}{c}$\widehat{D}$~samples\end{tabular}}%
\psfrag{b}[t][t][\tsize]{\color[rgb]{0,0,0}\setlength{\tabcolsep}{0pt}\begin{tabular}{c}$\widehat{D}'$~samples\end{tabular}}%
\psfrag{c}[b][b][\tsize]{\color[rgb]{0,0,0}\setlength{\tabcolsep}{0pt}\begin{tabular}{c}$\frac{1}{2}\mathcal{P}$\end{tabular}}%
%
\includegraphics[width=0.3\textwidth]{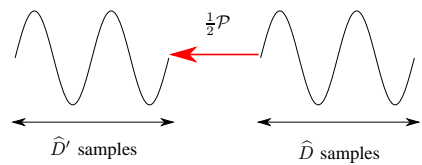}
\end{psfrags}%
%

	\caption{Oscillations detection method using the covariance function, $cov(\widehat{D},\widehat{D}')$.}
	\label{fig:cov_description}
\end{figure}

\noindent Fig.~\ref{fig:cov_description} illustrates the oscillations detection method using the covariance function. 
The $\widehat{D}'$ samples set is the previous samples set of $\widehat{D}$, which is $\frac{1}{2}\mathcal{P}$ lagging behind the current samples set of $\widehat{D}$, where $\mathcal{P}$ is the period of one full oscillation. 

\begin{figure}
	\centering
	\begin{subfigure}[thpb]{0.45\textwidth}
%
%
\begin{psfrags}%
\psfragscanon%
\newcommand{\tsize}{0.7}
%
\psfrag{s11}[b][b][\tsize]{\color[rgb]{0,0,0}\setlength{\tabcolsep}{0pt}\begin{tabular}{c}Magnitude\end{tabular}}%
\psfrag{s12}[t][t][\tsize]{\color[rgb]{0,0,0}\setlength{\tabcolsep}{0pt}\begin{tabular}{c}Time~(s)\end{tabular}}%
%
\psfrag{x01}[t][t][\tsize]{0}%
\psfrag{x02}[t][t][\tsize]{1}%
\psfrag{x03}[t][t][\tsize]{2}%
\psfrag{x04}[t][t][\tsize]{3}%
\psfrag{x05}[t][t][\tsize]{4}%
\psfrag{x06}[t][t][\tsize]{5}%
\psfrag{x07}[t][t][\tsize]{6}%
\psfrag{x08}[t][t][\tsize]{7}%
\psfrag{x09}[t][t][\tsize]{8}%
\psfrag{x10}[t][t][\tsize]{9}%
\psfrag{x11}[t][t][\tsize]{10}%
\psfrag{x12}[t][t][\tsize]{0}%
\psfrag{x13}[t][t][\tsize]{1}%
\psfrag{x14}[t][t][\tsize]{2}%
\psfrag{x15}[t][t][\tsize]{3}%
\psfrag{x16}[t][t][\tsize]{4}%
\psfrag{x17}[t][t][\tsize]{5}%
\psfrag{x18}[t][t][\tsize]{6}%
\psfrag{x19}[t][t][\tsize]{7}%
\psfrag{x20}[t][t][\tsize]{8}%
\psfrag{x21}[t][t][\tsize]{9}%
\psfrag{x22}[t][t][\tsize]{10}%
\psfrag{x23}[t][t][\tsize]{0}%
\psfrag{x24}[t][t][\tsize]{1}%
\psfrag{x25}[t][t][\tsize]{2}%
\psfrag{x26}[t][t][\tsize]{3}%
\psfrag{x27}[t][t][\tsize]{4}%
\psfrag{x28}[t][t][\tsize]{5}%
\psfrag{x29}[t][t][\tsize]{6}%
\psfrag{x30}[t][t][\tsize]{7}%
\psfrag{x31}[t][t][\tsize]{8}%
\psfrag{x32}[t][t][\tsize]{9}%
\psfrag{x33}[t][t][\tsize]{10}%
%
\psfrag{v01}[r][r][\tsize]{-0.4}%
\psfrag{v02}[r][r][\tsize]{}%
\psfrag{v03}[r][r][\tsize]{0}%
\psfrag{v04}[r][r][\tsize]{}%
\psfrag{v05}[r][r][\tsize]{0.4}%
\psfrag{v06}[r][r][\tsize]{-0.4}%
\psfrag{v07}[r][r][\tsize]{}%
\psfrag{v08}[r][r][\tsize]{0}%
\psfrag{v09}[r][r][\tsize]{}%
\psfrag{v10}[r][r][\tsize]{0.4}%
\psfrag{v11}[r][r][\tsize]{-0.2}%
\psfrag{v12}[r][r][\tsize]{}%
\psfrag{v13}[r][r][\tsize]{0}%
\psfrag{v14}[r][r][\tsize]{}%
\psfrag{v15}[r][r][\tsize]{0.2}%
%
\includegraphics[width=\textwidth]{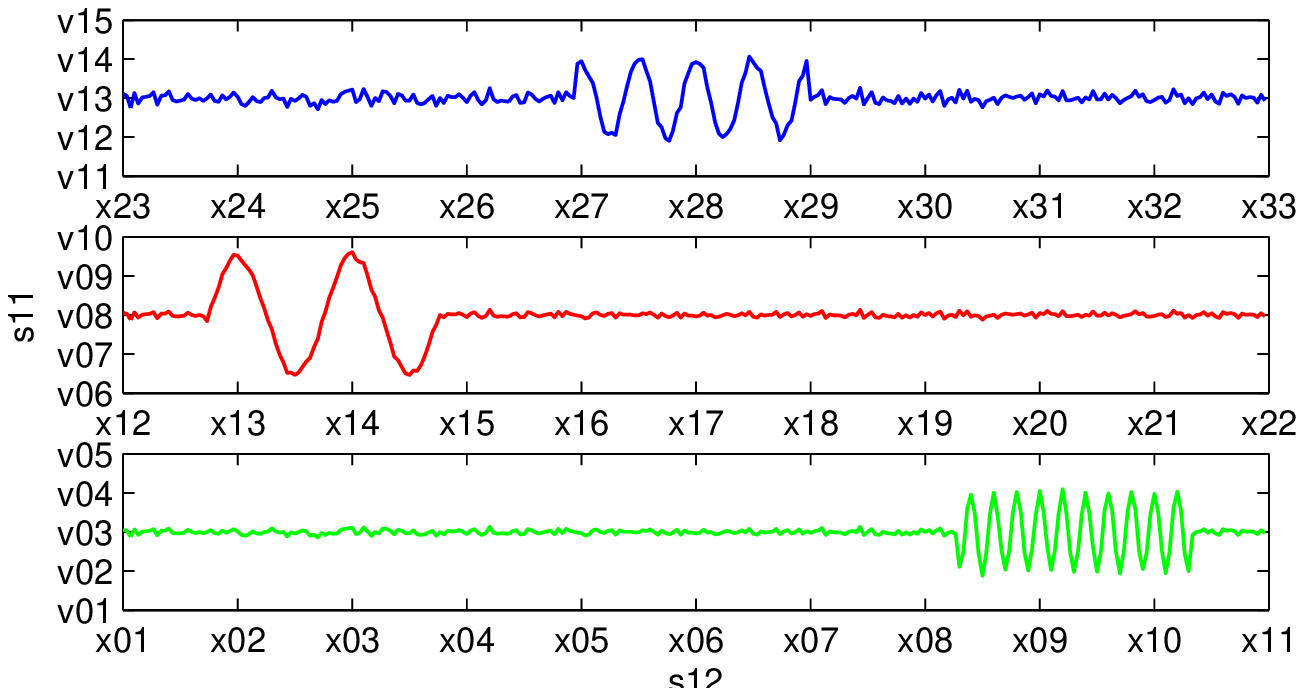}
\end{psfrags}%
%

		\caption{Generated signals of different frequencies and amplitudes, with additional noise.}
		\label{fig:generated_multi_signal}
	\end{subfigure}
	\\ 
	\begin{subfigure}[thpb]{0.45\textwidth}
%
%
\begin{psfrags}%
\psfragscanon%
\newcommand{\tsize}{0.7}
%
\psfrag{s11}[b][b][\tsize]{\color[rgb]{0,0,0}\setlength{\tabcolsep}{0pt}\begin{tabular}{c}Magnitude\end{tabular}}%
\psfrag{s12}[t][t][\tsize]{\color[rgb]{0,0,0}\setlength{\tabcolsep}{0pt}\begin{tabular}{c}Frequency (Hz)\end{tabular}}%
%
\psfrag{x01}[t][t][\tsize]{-15}%
\psfrag{x02}[t][t][\tsize]{-10}%
\psfrag{x03}[t][t][\tsize]{-5}%
\psfrag{x04}[t][t][\tsize]{0}%
\psfrag{x05}[t][t][\tsize]{5}%
\psfrag{x06}[t][t][\tsize]{10}%
\psfrag{x07}[t][t][\tsize]{15}%
\psfrag{x08}[t][t][\tsize]{-15}%
\psfrag{x09}[t][t][\tsize]{-10}%
\psfrag{x10}[t][t][\tsize]{-5}%
\psfrag{x11}[t][t][\tsize]{0}%
\psfrag{x12}[t][t][\tsize]{5}%
\psfrag{x13}[t][t][\tsize]{10}%
\psfrag{x14}[t][t][\tsize]{15}%
\psfrag{x15}[t][t][\tsize]{-15}%
\psfrag{x16}[t][t][\tsize]{-10}%
\psfrag{x17}[t][t][\tsize]{-5}%
\psfrag{x18}[t][t][\tsize]{0}%
\psfrag{x19}[t][t][\tsize]{5}%
\psfrag{x20}[t][t][\tsize]{10}%
\psfrag{x21}[t][t][\tsize]{15}%
%
\psfrag{v01}[r][r][\tsize]{0}%
\psfrag{v02}[r][r][\tsize]{}%
\psfrag{v03}[r][r][\tsize]{4}%
\psfrag{v04}[r][r][\tsize]{}%
\psfrag{v05}[r][r][\tsize]{8}%
\psfrag{v06}[r][r][\tsize]{0}%
\psfrag{v07}[r][r][\tsize]{}%
\psfrag{v08}[r][r][\tsize]{4}%
\psfrag{v09}[r][r][\tsize]{}%
\psfrag{v10}[r][r][\tsize]{8}%
\psfrag{v11}[r][r][\tsize]{}%
\psfrag{v12}[r][r][\tsize]{0}%
\psfrag{v13}[r][r][\tsize]{}%
\psfrag{v14}[r][r][\tsize]{2}%
\psfrag{v15}[r][r][\tsize]{}%
\psfrag{v16}[r][r][\tsize]{4}%
%
\includegraphics[width=\textwidth]{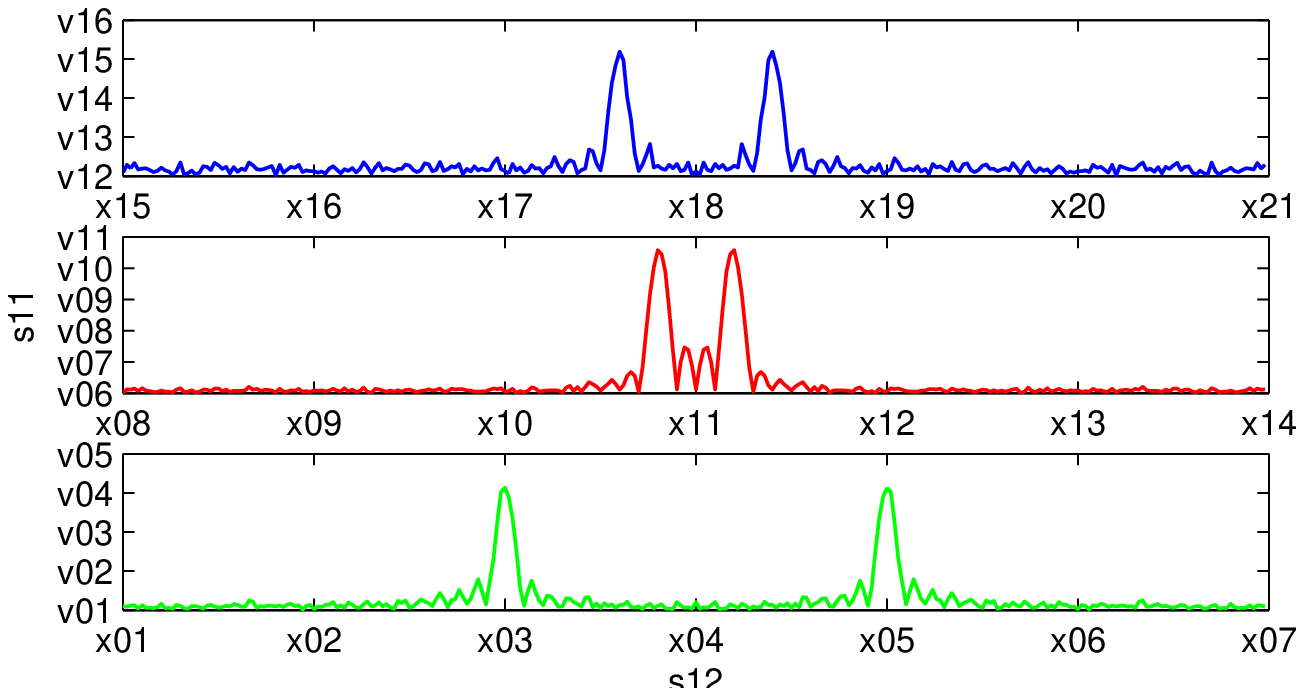}
\end{psfrags}%
%

		\caption{Discrete Fourier Transform of the generated signals.}
		\label{fig:fft_generated_multi_signal}
	\end{subfigure}
	\\ 
	\begin{subfigure}[thpb]{0.45\textwidth}
%
%
\begin{psfrags}%
\psfragscanon%
\newcommand{\tsize}{0.7}
%
\psfrag{s11}[b][b][\tsize]{\color[rgb]{0,0,0}\setlength{\tabcolsep}{0pt}\begin{tabular}{c}$cov(\widehat{D},\widehat{D}')$\end{tabular}}%
\psfrag{s12}[t][t][\tsize]{\color[rgb]{0,0,0}\setlength{\tabcolsep}{0pt}\begin{tabular}{c}Time~(s)\end{tabular}}%
%
\psfrag{x01}[t][t][\tsize]{0}%
\psfrag{x02}[t][t][\tsize]{1}%
\psfrag{x03}[t][t][\tsize]{2}%
\psfrag{x04}[t][t][\tsize]{3}%
\psfrag{x05}[t][t][\tsize]{4}%
\psfrag{x06}[t][t][\tsize]{5}%
\psfrag{x07}[t][t][\tsize]{6}%
\psfrag{x08}[t][t][\tsize]{7}%
\psfrag{x09}[t][t][\tsize]{8}%
\psfrag{x10}[t][t][\tsize]{9}%
\psfrag{x11}[t][t][\tsize]{10}%
\psfrag{x12}[t][t][\tsize]{0}%
\psfrag{x13}[t][t][\tsize]{1}%
\psfrag{x14}[t][t][\tsize]{2}%
\psfrag{x15}[t][t][\tsize]{3}%
\psfrag{x16}[t][t][\tsize]{4}%
\psfrag{x17}[t][t][\tsize]{5}%
\psfrag{x18}[t][t][\tsize]{6}%
\psfrag{x19}[t][t][\tsize]{7}%
\psfrag{x20}[t][t][\tsize]{8}%
\psfrag{x21}[t][t][\tsize]{9}%
\psfrag{x22}[t][t][\tsize]{10}%
\psfrag{x23}[t][t][\tsize]{0}%
\psfrag{x24}[t][t][\tsize]{1}%
\psfrag{x25}[t][t][\tsize]{2}%
\psfrag{x26}[t][t][\tsize]{3}%
\psfrag{x27}[t][t][\tsize]{4}%
\psfrag{x28}[t][t][\tsize]{5}%
\psfrag{x29}[t][t][\tsize]{6}%
\psfrag{x30}[t][t][\tsize]{7}%
\psfrag{x31}[t][t][\tsize]{8}%
\psfrag{x32}[t][t][\tsize]{9}%
\psfrag{x33}[t][t][\tsize]{10}%
%
\psfrag{v01}[r][r][\tsize]{}%
\psfrag{v02}[r][r][\tsize]{-0.1}%
\psfrag{v03}[r][r][\tsize]{}%
\psfrag{v04}[r][r][\tsize]{0}%
\psfrag{v05}[r][r][\tsize]{}%
\psfrag{v06}[r][r][\tsize]{}%
\psfrag{v07}[r][r][\tsize]{-1}%
\psfrag{v08}[r][r][\tsize]{}%
\psfrag{v09}[r][r][\tsize]{0}%
\psfrag{v10}[r][r][\tsize]{}%
\psfrag{v11}[r][r][\tsize]{-0.1}%
\psfrag{v12}[r][r][\tsize]{}%
\psfrag{v13}[r][r][\tsize]{0}%
\psfrag{v14}[r][r][\tsize]{}%
\psfrag{v15}[r][r][\tsize]{0.1}%
%
\includegraphics[width=\textwidth]{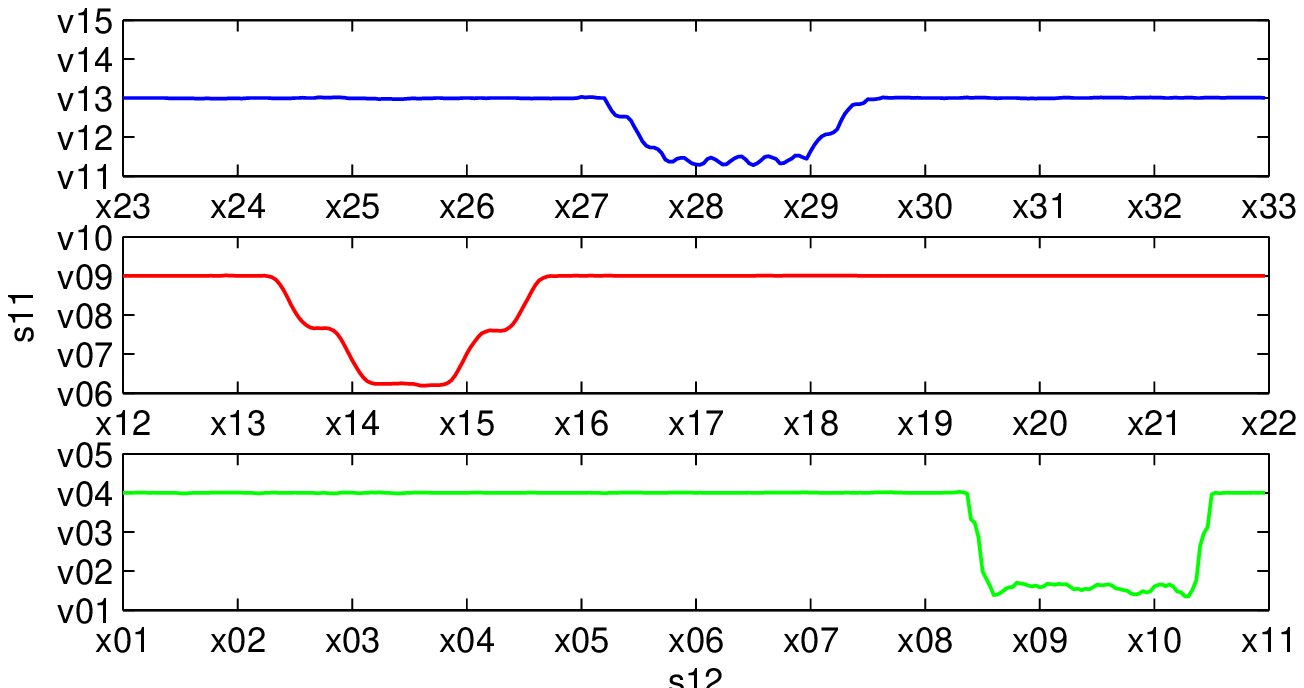}
\end{psfrags}%
%

		\caption{Covariances of the generated signals with delays $\frac{1}{2}\mathcal{P}$ matching the half periods of the frequencies.}
		\label{fig:cov_generated_multi_signal}
	\end{subfigure}
	\caption{Generated signals with different magnitudes (0.1, 0.3, 0.2) and frequencies (2, 1, 5 Hz) of oscillations and noise, and the corresponding Discrete Fourier Transform and covariances.}
	\label{fig:cov_example}
\end{figure}

To show the feasibility of this method, consider the signals which are generated with oscillations of different magnitudes (0.1, 0.3, 0.2) and frequencies (2, 1, 5 Hz), and noise is added to these signals as shown in Fig.~\ref{fig:generated_multi_signal}. By computing their discrete Fourier Transforms (DFT), the frequency of the oscillation in each signal can be found as presented in Fig.~\ref{fig:fft_generated_multi_signal}. Then, the proposed covariance function of the generated signals can be computed, based on the period of the oscillation known from the DFT. Fig.~\ref{fig:cov_generated_multi_signal} clearly shows that the covariances of the generated signals are large at the instance when the oscillations occur. 

\begin{figure}[thpb]
	\centering
%
%
\begin{psfrags}%
\psfragscanon%
\newcommand{\tsize}{0.7}
%
\psfrag{s03}[t][t][\tsize]{\color[rgb]{0,0,0}\setlength{\tabcolsep}{0pt}\begin{tabular}{c}Time~(s)\end{tabular}}%
\psfrag{s04}[b][b][\tsize]{\color[rgb]{0,0,0}\setlength{\tabcolsep}{0pt}\begin{tabular}{c}$cov(\widehat{D},\widehat{D}')~[\times10^{-3}]$\end{tabular}}%
%
\psfrag{x01}[t][t][\tsize]{0}%
\psfrag{x02}[t][t][\tsize]{2}%
\psfrag{x03}[t][t][\tsize]{4}%
\psfrag{x04}[t][t][\tsize]{6}%
\psfrag{x05}[t][t][\tsize]{8}%
\psfrag{x06}[t][t][\tsize]{10}%
\psfrag{x07}[t][t][\tsize]{12}%
\psfrag{x08}[t][t][\tsize]{14}%
\psfrag{x09}[t][t][\tsize]{16}%
\psfrag{x10}[t][t][\tsize]{18}%
\psfrag{x11}[t][t][\tsize]{20}%
%
\psfrag{v01}[r][r][\tsize]{-60}%
\psfrag{v02}[r][r][\tsize]{-50}%
\psfrag{v03}[r][r][\tsize]{-40}%
\psfrag{v04}[r][r][\tsize]{-30}%
\psfrag{v05}[r][r][\tsize]{-20}%
\psfrag{v06}[r][r][\tsize]{-10}%
\psfrag{v07}[r][r][\tsize]{0}%
\psfrag{v08}[r][r][\tsize]{10}%
%
\includegraphics[width=0.35\textwidth]{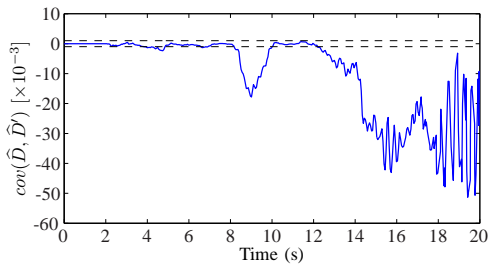}
\end{psfrags}%
%

	\caption{Covariance of the flow divergence obtained from the experiment presented in Fig.~\ref{fig:fixed_gain_size_divergence_PI}.}
	\label{fig:cov_fixed_gain_size_divergence_PI}
\end{figure}

This method is implemented on the MAV and tested in the same experiment presented in Fig.~\ref{fig:fixed_gain_size_divergence_PI}. Fig.~\ref{fig:cov_fixed_gain_size_divergence_PI} shows the covariance of the flow divergence, indicating that we can successfully detect the oscillations in a real flight. For instance, in Fig.~\ref{fig:fixed_gain_size_divergence_PI}, we observe that strong oscillations happen from $8s$ to $10s$ and from $12s$ to $20s$, and this leads to highly negative covariance values at these time instances as shown in Fig.~\ref{fig:cov_fixed_gain_size_divergence_PI}. Thus, this method provides a new way to detect oscillations in real-time.

\section{Flight Tests}
\label{sec:FlightTests}
\begin{figure}[thpb]
	\centering
	\includegraphics[width=0.3\textwidth]{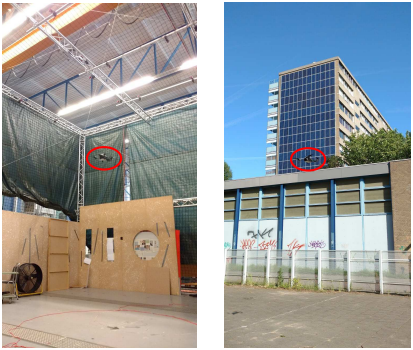}
	\caption{Indoor and outdoor environments for the flight tests.}
	\label{fig:indoor_outdoor_environments}
\end{figure}

In this section, we present the flight tests results for vertical landing control using the adaptive controller. The same MAV platform as described in Section~\ref{subsec:TestingPlatform} is used for the experiments, with all vision and control algorithms running on-board. To show the robustness of the proposed method in the face of external disturbances, flight tests are performed not only in indoor but also in outdoor environments as shown in Fig.~\ref{fig:indoor_outdoor_environments}. \footnote{Explanatory video with experiments: \url{https://goo.gl/ZDPP3m}}

\subsection{Indoor Flight Tests}
\label{subsec:IndoorExperiments}
\begin{figure}
	\centering
	\begin{subfigure}[b]{0.5\textwidth}
%
%
\begin{psfrags}%
\psfragscanon%
\newcommand{\tsize}{0.7}
\newcommand{\tsizeb}{0.7}
%
\psfrag{s15}[b][b][\tsize]{\color[rgb]{0,0,0}\setlength{\tabcolsep}{0pt}\begin{tabular}{c}Time~(s)\end{tabular}}%
\psfrag{s16}[t][b][\tsize]{\color[rgb]{0,0,0}\setlength{\tabcolsep}{0pt}\begin{tabular}{c}$Z~(m)$\end{tabular}}%
\psfrag{s17}[b][b][\tsize]{\color[rgb]{0,0,0}\setlength{\tabcolsep}{0pt}\begin{tabular}{c}Time~(s)\end{tabular}}%
\psfrag{s18}[t][b][\tsize]{\color[rgb]{0,0,0}\setlength{\tabcolsep}{0pt}\begin{tabular}{c}$V_Z~(\times 10^{-1}m/s)$\end{tabular}}%
\psfrag{s19}[b][b][\tsize]{\color[rgb]{0,0,0}\setlength{\tabcolsep}{0pt}\begin{tabular}{c}Time~(s)\end{tabular}}%
\psfrag{s20}[t][b][\tsize]{\color[rgb]{0,0,0}\setlength{\tabcolsep}{0pt}\begin{tabular}{c}$cov(\widehat{D},\widehat{D}')$\end{tabular}}%
\psfrag{s21}[b][b][\tsize]{\color[rgb]{0,0,0}\setlength{\tabcolsep}{0pt}\begin{tabular}{c}Time~(s)\end{tabular}}%
\psfrag{s22}[t][b][\tsize]{\color[rgb]{0,0,0}\setlength{\tabcolsep}{0pt}\begin{tabular}{c}$\widehat{D}~(\times 10^{-1}/s)$\end{tabular}}%
\psfrag{s23}[b][b][\tsize]{\color[rgb]{0,0,0}\setlength{\tabcolsep}{0pt}\begin{tabular}{c}Time~(s)\end{tabular}}%
\psfrag{s24}[t][b][\tsize]{\color[rgb]{0,0,0}\setlength{\tabcolsep}{0pt}\begin{tabular}{c}$\mu~(\times 10^{-2}m/s^2)$\end{tabular}}%
\psfrag{s25}[b][b][\tsize]{\color[rgb]{0,0,0}\setlength{\tabcolsep}{0pt}\begin{tabular}{c}Time~(s)\end{tabular}}%
\psfrag{s26}[t][b][\tsize]{\color[rgb]{0,0,0}\setlength{\tabcolsep}{0pt}\begin{tabular}{c}$K~(\times 10^{-1})$\end{tabular}}%
\psfrag{s30}[][]{\color[rgb]{0,0,0}\setlength{\tabcolsep}{0pt}\begin{tabular}{c} \end{tabular}}%
\psfrag{s31}[][]{\color[rgb]{0,0,0}\setlength{\tabcolsep}{0pt}\begin{tabular}{c} \end{tabular}}%
\psfrag{s32}[l][l][\tsize]{\color[rgb]{0,0,0}$K_i$}%
\psfrag{s33}[l][l][\tsize]{\color[rgb]{0,0,0}$K_p$}%
\psfrag{s34}[l][l][\tsize]{\color[rgb]{0,0,0}$K_i$}%
%
\psfrag{x01}[t][t][\tsizeb]{0}%
\psfrag{x02}[t][t][\tsizeb]{5}%
\psfrag{x03}[t][t][\tsizeb]{10}%
\psfrag{x04}[t][t][\tsizeb]{15}%
\psfrag{x05}[t][t][\tsizeb]{20}%
\psfrag{x06}[t][t][\tsizeb]{0}%
\psfrag{x07}[t][t][\tsizeb]{5}%
\psfrag{x08}[t][t][\tsizeb]{10}%
\psfrag{x09}[t][t][\tsizeb]{15}%
\psfrag{x10}[t][t][\tsizeb]{20}%
\psfrag{x11}[t][t][\tsizeb]{0}%
\psfrag{x12}[t][t][\tsizeb]{5}%
\psfrag{x13}[t][t][\tsizeb]{10}%
\psfrag{x14}[t][t][\tsizeb]{15}%
\psfrag{x15}[t][t][\tsizeb]{20}%
\psfrag{x16}[t][t][\tsizeb]{0}%
\psfrag{x17}[t][t][\tsizeb]{5}%
\psfrag{x18}[t][t][\tsizeb]{10}%
\psfrag{x19}[t][t][\tsizeb]{15}%
\psfrag{x20}[t][t][\tsizeb]{20}%
\psfrag{x21}[t][t][\tsizeb]{0}%
\psfrag{x22}[t][t][\tsizeb]{5}%
\psfrag{x23}[t][t][\tsizeb]{10}%
\psfrag{x24}[t][t][\tsizeb]{15}%
\psfrag{x25}[t][t][\tsizeb]{20}%
\psfrag{x26}[t][t][\tsizeb]{0}%
\psfrag{x27}[t][t][\tsizeb]{5}%
\psfrag{x28}[t][t][\tsizeb]{10}%
\psfrag{x29}[t][t][\tsizeb]{15}%
\psfrag{x30}[t][t][\tsizeb]{20}%
%
\psfrag{v01}[r][r][\tsizeb]{0}%
\psfrag{v02}[r][r][\tsizeb]{}%
\psfrag{v03}[r][r][\tsizeb]{1}%
\psfrag{v04}[r][r][\tsizeb]{}%
\psfrag{v05}[r][r][\tsizeb]{2}%
\psfrag{v06}[r][r][\tsizeb]{}%
\psfrag{v07}[r][r][\tsizeb]{3}%
\psfrag{v08}[r][r][\tsizeb]{}%
\psfrag{v09}[r][r][\tsizeb]{4}%
\psfrag{v10}[r][r][\tsizeb]{}%
\psfrag{v11}[r][r][\tsizeb]{-8}%
\psfrag{v12}[r][r][\tsizeb]{}%
\psfrag{v13}[r][r][\tsizeb]{-4}%
\psfrag{v14}[r][r][\tsizeb]{}%
\psfrag{v15}[r][r][\tsizeb]{0}%
\psfrag{v16}[r][r][\tsizeb]{}%
\psfrag{v17}[r][r][\tsizeb]{4}%
\psfrag{v18}[r][r][\tsizeb]{}%
\psfrag{v19}[r][r][\tsizeb]{-3}%
\psfrag{v20}[r][r][\tsizeb]{}%
\psfrag{v21}[r][r][\tsizeb]{-2}%
\psfrag{v22}[r][r][\tsizeb]{}%
\psfrag{v23}[r][r][\tsizeb]{-1}%
\psfrag{v24}[r][r][\tsizeb]{}%
\psfrag{v25}[r][r][\tsizeb]{0}%
\psfrag{v26}[r][r][\tsizeb]{}%
\psfrag{v27}[r][r][\tsizeb]{1}%
\psfrag{v28}[r][r][\tsizeb]{-2}%
\psfrag{v29}[r][r][\tsizeb]{}%
\psfrag{v30}[r][r][\tsizeb]{-1}%
\psfrag{v31}[r][r][\tsizeb]{}%
\psfrag{v32}[r][r][\tsizeb]{0}%
\psfrag{v33}[r][r][\tsizeb]{}%
\psfrag{v34}[r][r][\tsizeb]{1}%
\psfrag{v35}[r][r][\tsizeb]{}%
\psfrag{ypower5}[Bl][Bl][\tsizeb]{$\times 10^{-3}$}%
\psfrag{v36}[r][r][\tsizeb]{-4}%
\psfrag{v37}[r][r][\tsizeb]{}%
\psfrag{v38}[r][r][\tsizeb]{-2}%
\psfrag{v39}[r][r][\tsizeb]{}%
\psfrag{v40}[r][r][\tsizeb]{0}%
\psfrag{v41}[r][r][\tsizeb]{}%
\psfrag{v42}[r][r][\tsizeb]{2}%
\psfrag{v43}[r][r][\tsizeb]{}%
\psfrag{v44}[r][r][\tsizeb]{0}%
\psfrag{v45}[r][r][\tsizeb]{}%
\psfrag{v46}[r][r][\tsizeb]{1}%
\psfrag{v47}[r][r][\tsizeb]{}%
\psfrag{v48}[r][r][\tsizeb]{2}%
\psfrag{v49}[r][r][\tsizeb]{}%
\psfrag{v50}[r][r][\tsizeb]{3}%
\psfrag{v51}[r][r][\tsizeb]{}%
\psfrag{v52}[r][r][\tsizeb]{4}%
%
\includegraphics[width=\textwidth]{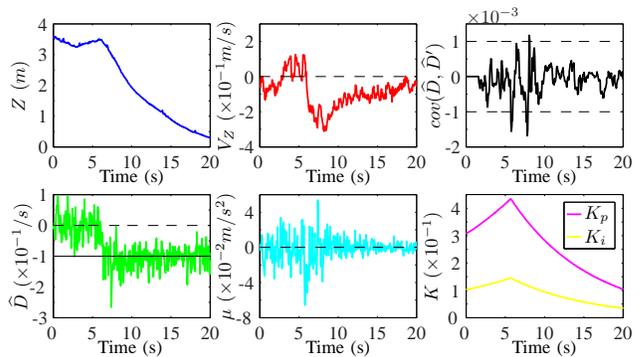}
\end{psfrags}%
%

		\caption{$D^*=-0.1$.}
		\label{fig:indoor_adaptive_control_sp_0_1}
	\end{subfigure}
	\\ 
	\begin{subfigure}[b]{0.5\textwidth}
%
%
\begin{psfrags}%
\psfragscanon%
\newcommand{\tsize}{0.7}
%
\psfrag{s15}[b][b][\tsize]{\color[rgb]{0,0,0}\setlength{\tabcolsep}{0pt}\begin{tabular}{c}Time~(s)\end{tabular}}%
\psfrag{s16}[t][b][\tsize]{\color[rgb]{0,0,0}\setlength{\tabcolsep}{0pt}\begin{tabular}{c}$Z~(m)$\end{tabular}}%
\psfrag{s17}[b][b][\tsize]{\color[rgb]{0,0,0}\setlength{\tabcolsep}{0pt}\begin{tabular}{c}Time~(s)\end{tabular}}%
\psfrag{s18}[t][b][\tsize]{\color[rgb]{0,0,0}\setlength{\tabcolsep}{0pt}\begin{tabular}{c}$V_Z~(\times 10^{-1}m/s)$\end{tabular}}%
\psfrag{s19}[b][b][\tsize]{\color[rgb]{0,0,0}\setlength{\tabcolsep}{0pt}\begin{tabular}{c}Time~(s)\end{tabular}}%
\psfrag{s20}[t][b][\tsize]{\color[rgb]{0,0,0}\setlength{\tabcolsep}{0pt}\begin{tabular}{c}$cov(\widehat{D},\widehat{D}')$\end{tabular}}%
\psfrag{s21}[b][b][\tsize]{\color[rgb]{0,0,0}\setlength{\tabcolsep}{0pt}\begin{tabular}{c}Time~(s)\end{tabular}}%
\psfrag{s22}[t][b][\tsize]{\color[rgb]{0,0,0}\setlength{\tabcolsep}{0pt}\begin{tabular}{c}$\widehat{D}~(\times 10^{-1}/s)$\end{tabular}}%
\psfrag{s23}[b][b][\tsize]{\color[rgb]{0,0,0}\setlength{\tabcolsep}{0pt}\begin{tabular}{c}Time~(s)\end{tabular}}%
\psfrag{s24}[t][b][\tsize]{\color[rgb]{0,0,0}\setlength{\tabcolsep}{0pt}\begin{tabular}{c}$\mu~(\times 10^{-2}m/s^2)$\end{tabular}}%
\psfrag{s25}[b][b][\tsize]{\color[rgb]{0,0,0}\setlength{\tabcolsep}{0pt}\begin{tabular}{c}Time~(s)\end{tabular}}%
\psfrag{s26}[t][b][\tsize]{\color[rgb]{0,0,0}\setlength{\tabcolsep}{0pt}\begin{tabular}{c}$K~(\times 10^{-1})$\end{tabular}}%
\psfrag{s30}[][]{\color[rgb]{0,0,0}\setlength{\tabcolsep}{0pt}\begin{tabular}{c} \end{tabular}}%
\psfrag{s31}[][]{\color[rgb]{0,0,0}\setlength{\tabcolsep}{0pt}\begin{tabular}{c} \end{tabular}}%
\psfrag{s32}[l][l][\tsize]{\color[rgb]{0,0,0}$K_i$}%
\psfrag{s33}[l][l][\tsize]{\color[rgb]{0,0,0}$K_p$}%
\psfrag{s34}[l][l][\tsize]{\color[rgb]{0,0,0}$K_i$}%
%
\psfrag{x01}[t][t][\tsize]{0}%
\psfrag{x02}[t][t][\tsize]{2}%
\psfrag{x03}[t][t][\tsize]{4}%
\psfrag{x04}[t][t][\tsize]{6}%
\psfrag{x05}[t][t][\tsize]{8}%
\psfrag{x06}[t][t][\tsize]{10}%
\psfrag{x07}[t][t][\tsize]{0}%
\psfrag{x08}[t][t][\tsize]{2}%
\psfrag{x09}[t][t][\tsize]{4}%
\psfrag{x10}[t][t][\tsize]{6}%
\psfrag{x11}[t][t][\tsize]{8}%
\psfrag{x12}[t][t][\tsize]{10}%
\psfrag{x13}[t][t][\tsize]{0}%
\psfrag{x14}[t][t][\tsize]{2}%
\psfrag{x15}[t][t][\tsize]{4}%
\psfrag{x16}[t][t][\tsize]{6}%
\psfrag{x17}[t][t][\tsize]{8}%
\psfrag{x18}[t][t][\tsize]{10}%
\psfrag{x19}[t][t][\tsize]{0}%
\psfrag{x20}[t][t][\tsize]{2}%
\psfrag{x21}[t][t][\tsize]{4}%
\psfrag{x22}[t][t][\tsize]{6}%
\psfrag{x23}[t][t][\tsize]{8}%
\psfrag{x24}[t][t][\tsize]{10}%
\psfrag{x25}[t][t][\tsize]{0}%
\psfrag{x26}[t][t][\tsize]{2}%
\psfrag{x27}[t][t][\tsize]{4}%
\psfrag{x28}[t][t][\tsize]{6}%
\psfrag{x29}[t][t][\tsize]{8}%
\psfrag{x30}[t][t][\tsize]{10}%
\psfrag{x31}[t][t][\tsize]{0}%
\psfrag{x32}[t][t][\tsize]{2}%
\psfrag{x33}[t][t][\tsize]{4}%
\psfrag{x34}[t][t][\tsize]{6}%
\psfrag{x35}[t][t][\tsize]{8}%
\psfrag{x36}[t][t][\tsize]{10}%
%
\psfrag{v01}[r][r][\tsize]{0}%
\psfrag{v02}[r][r][\tsize]{}%
\psfrag{v03}[r][r][\tsize]{1}%
\psfrag{v04}[r][r][\tsize]{}%
\psfrag{v05}[r][r][\tsize]{2}%
\psfrag{v06}[r][r][\tsize]{}%
\psfrag{v07}[r][r][\tsize]{3}%
\psfrag{v08}[r][r][\tsize]{}%
\psfrag{v09}[r][r][\tsize]{4}%
\psfrag{v10}[r][r][\tsize]{}%
\psfrag{v11}[r][r][\tsize]{-8}%
\psfrag{v12}[r][r][\tsize]{}%
\psfrag{v13}[r][r][\tsize]{-4}%
\psfrag{v14}[r][r][\tsize]{}%
\psfrag{v15}[r][r][\tsize]{0}%
\psfrag{v16}[r][r][\tsize]{}%
\psfrag{v17}[r][r][\tsize]{4}%
\psfrag{v18}[r][r][\tsize]{}%
\psfrag{v19}[r][r][\tsize]{}%
\psfrag{v20}[r][r][\tsize]{-3}%
\psfrag{v21}[r][r][\tsize]{}%
\psfrag{v22}[r][r][\tsize]{-2}%
\psfrag{v23}[r][r][\tsize]{}%
\psfrag{v24}[r][r][\tsize]{-1}%
\psfrag{v25}[r][r][\tsize]{}%
\psfrag{v26}[r][r][\tsize]{0}%
\psfrag{v27}[r][r][\tsize]{}%
\psfrag{v28}[r][r][\tsize]{1}%
\psfrag{v29}[r][r][\tsize]{}%
\psfrag{v30}[r][r][\tsize]{-2}%
\psfrag{v31}[r][r][\tsize]{}%
\psfrag{v32}[r][r][\tsize]{0}%
\psfrag{v33}[r][r][\tsize]{}%
\psfrag{v34}[r][r][\tsize]{2}%
\psfrag{v35}[r][r][\tsize]{}%
\psfrag{ypower5}[Bl][Bl][\tsize]{$\times 10^{-3}$}%
\psfrag{v36}[r][r][\tsize]{}%
\psfrag{v37}[r][r][\tsize]{-4}%
\psfrag{v38}[r][r][\tsize]{}%
\psfrag{v39}[r][r][\tsize]{-2}%
\psfrag{v40}[r][r][\tsize]{}%
\psfrag{v41}[r][r][\tsize]{0}%
\psfrag{v42}[r][r][\tsize]{}%
\psfrag{v43}[r][r][\tsize]{2}%
\psfrag{v44}[r][r][\tsize]{0}%
\psfrag{v45}[r][r][\tsize]{}%
\psfrag{v46}[r][r][\tsize]{1}%
\psfrag{v47}[r][r][\tsize]{}%
\psfrag{v48}[r][r][\tsize]{2}%
\psfrag{v49}[r][r][\tsize]{}%
\psfrag{v50}[r][r][\tsize]{3}%
\psfrag{v51}[r][r][\tsize]{}%
\psfrag{v52}[r][r][\tsize]{4}%
%
\includegraphics[width=\textwidth]{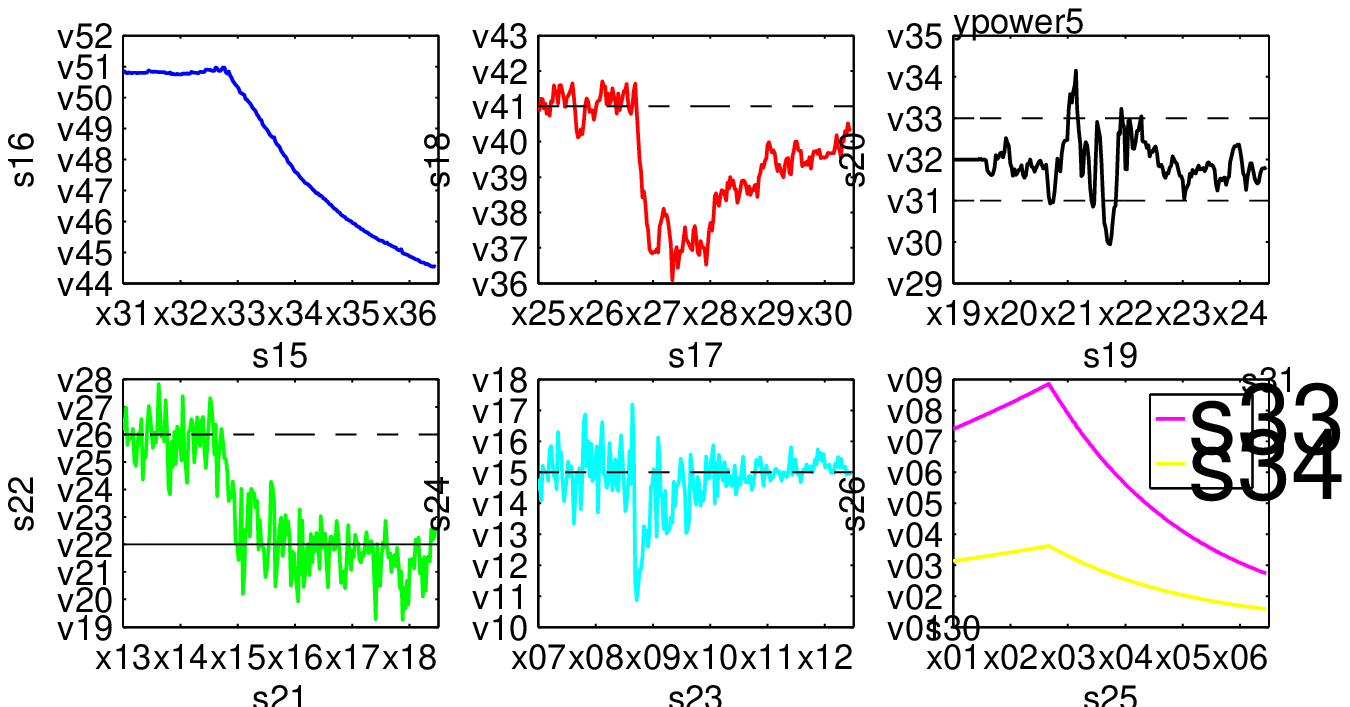}
\end{psfrags}%
%

		\caption{$D^*=-0.2$.}
		\label{fig:indoor_adaptive_control_sp_0_2}
	\end{subfigure}
	\\ 
	\begin{subfigure}[b]{0.5\textwidth}
%
%
\begin{psfrags}%
\psfragscanon%
\newcommand{\tsize}{0.7}
%
\psfrag{s15}[b][b][\tsize]{\color[rgb]{0,0,0}\setlength{\tabcolsep}{0pt}\begin{tabular}{c}Time~(s)\end{tabular}}%
\psfrag{s16}[t][b][\tsize]{\color[rgb]{0,0,0}\setlength{\tabcolsep}{0pt}\begin{tabular}{c}$Z~(m)$\end{tabular}}%
\psfrag{s17}[b][b][\tsize]{\color[rgb]{0,0,0}\setlength{\tabcolsep}{0pt}\begin{tabular}{c}Time~(s)\end{tabular}}%
\psfrag{s18}[t][b][\tsize]{\color[rgb]{0,0,0}\setlength{\tabcolsep}{0pt}\begin{tabular}{c}$V_Z~(\times 10^{-1}m/s)$\end{tabular}}%
\psfrag{s19}[b][b][\tsize]{\color[rgb]{0,0,0}\setlength{\tabcolsep}{0pt}\begin{tabular}{c}Time~(s)\end{tabular}}%
\psfrag{s20}[t][b][\tsize]{\color[rgb]{0,0,0}\setlength{\tabcolsep}{0pt}\begin{tabular}{c}$cov(\widehat{D},\widehat{D}')$\end{tabular}}%
\psfrag{s21}[b][b][\tsize]{\color[rgb]{0,0,0}\setlength{\tabcolsep}{0pt}\begin{tabular}{c}Time~(s)\end{tabular}}%
\psfrag{s22}[t][b][\tsize]{\color[rgb]{0,0,0}\setlength{\tabcolsep}{0pt}\begin{tabular}{c}$\widehat{D}~(\times 10^{-1}/s)$\end{tabular}}%
\psfrag{s23}[b][b][\tsize]{\color[rgb]{0,0,0}\setlength{\tabcolsep}{0pt}\begin{tabular}{c}Time~(s)\end{tabular}}%
\psfrag{s24}[t][b][\tsize]{\color[rgb]{0,0,0}\setlength{\tabcolsep}{0pt}\begin{tabular}{c}$\mu~(\times 10^{-2}m/s^2)$\end{tabular}}%
\psfrag{s25}[b][b][\tsize]{\color[rgb]{0,0,0}\setlength{\tabcolsep}{0pt}\begin{tabular}{c}Time~(s)\end{tabular}}%
\psfrag{s26}[t][b][\tsize]{\color[rgb]{0,0,0}\setlength{\tabcolsep}{0pt}\begin{tabular}{c}$K~(\times 10^{-1})$\end{tabular}}%
\psfrag{s30}[][]{\color[rgb]{0,0,0}\setlength{\tabcolsep}{0pt}\begin{tabular}{c} \end{tabular}}%
\psfrag{s31}[][]{\color[rgb]{0,0,0}\setlength{\tabcolsep}{0pt}\begin{tabular}{c} \end{tabular}}%
\psfrag{s32}[l][l][\tsize]{\color[rgb]{0,0,0}$K_i$}%
\psfrag{s33}[l][l][\tsize]{\color[rgb]{0,0,0}$K_p$}%
\psfrag{s34}[l][l][\tsize]{\color[rgb]{0,0,0}$K_i$}%
%
\psfrag{x01}[t][t][\tsize]{0}%
\psfrag{x02}[t][t][\tsize]{2}%
\psfrag{x03}[t][t][\tsize]{4}%
\psfrag{x04}[t][t][\tsize]{6}%
\psfrag{x05}[t][t][\tsize]{8}%
\psfrag{x06}[t][t][\tsize]{10}%
\psfrag{x07}[t][t][\tsize]{0}%
\psfrag{x08}[t][t][\tsize]{2}%
\psfrag{x09}[t][t][\tsize]{4}%
\psfrag{x10}[t][t][\tsize]{6}%
\psfrag{x11}[t][t][\tsize]{8}%
\psfrag{x12}[t][t][\tsize]{10}%
\psfrag{x13}[t][t][\tsize]{0}%
\psfrag{x14}[t][t][\tsize]{2}%
\psfrag{x15}[t][t][\tsize]{4}%
\psfrag{x16}[t][t][\tsize]{6}%
\psfrag{x17}[t][t][\tsize]{8}%
\psfrag{x18}[t][t][\tsize]{10}%
\psfrag{x19}[t][t][\tsize]{0}%
\psfrag{x20}[t][t][\tsize]{2}%
\psfrag{x21}[t][t][\tsize]{4}%
\psfrag{x22}[t][t][\tsize]{6}%
\psfrag{x23}[t][t][\tsize]{8}%
\psfrag{x24}[t][t][\tsize]{10}%
\psfrag{x25}[t][t][\tsize]{0}%
\psfrag{x26}[t][t][\tsize]{2}%
\psfrag{x27}[t][t][\tsize]{4}%
\psfrag{x28}[t][t][\tsize]{6}%
\psfrag{x29}[t][t][\tsize]{8}%
\psfrag{x30}[t][t][\tsize]{10}%
\psfrag{x31}[t][t][\tsize]{0}%
\psfrag{x32}[t][t][\tsize]{2}%
\psfrag{x33}[t][t][\tsize]{4}%
\psfrag{x34}[t][t][\tsize]{6}%
\psfrag{x35}[t][t][\tsize]{8}%
\psfrag{x36}[t][t][\tsize]{10}%
%
\psfrag{v01}[r][r][\tsize]{0}%
\psfrag{v02}[r][r][\tsize]{}%
\psfrag{v03}[r][r][\tsize]{1}%
\psfrag{v04}[r][r][\tsize]{}%
\psfrag{v05}[r][r][\tsize]{2}%
\psfrag{v06}[r][r][\tsize]{}%
\psfrag{v07}[r][r][\tsize]{3}%
\psfrag{v08}[r][r][\tsize]{}%
\psfrag{v09}[r][r][\tsize]{4}%
\psfrag{v10}[r][r][\tsize]{-12}%
\psfrag{v11}[r][r][\tsize]{}%
\psfrag{v12}[r][r][\tsize]{-8}%
\psfrag{v13}[r][r][\tsize]{}%
\psfrag{v14}[r][r][\tsize]{-4}%
\psfrag{v15}[r][r][\tsize]{}%
\psfrag{v16}[r][r][\tsize]{0}%
\psfrag{v17}[r][r][\tsize]{}%
\psfrag{v18}[r][r][\tsize]{4}%
\psfrag{v19}[r][r][\tsize]{-4}%
\psfrag{v20}[r][r][\tsize]{-3}%
\psfrag{v21}[r][r][\tsize]{-2}%
\psfrag{v22}[r][r][\tsize]{-1}%
\psfrag{v23}[r][r][\tsize]{0}%
\psfrag{v24}[r][r][\tsize]{1}%
\psfrag{v25}[r][r][\tsize]{-2}%
\psfrag{v26}[r][r][\tsize]{}%
\psfrag{v27}[r][r][\tsize]{0}%
\psfrag{v28}[r][r][\tsize]{}%
\psfrag{v29}[r][r][\tsize]{2}%
\psfrag{v30}[r][r][\tsize]{}%
\psfrag{v31}[r][r][\tsize]{4}%
\psfrag{v32}[r][r][\tsize]{}%
\psfrag{v33}[r][r][\tsize]{6}%
\psfrag{ypower5}[Bl][Bl][\tsize]{$\times 10^{-3}$}%
\psfrag{v34}[r][r][\tsize]{-8}%
\psfrag{v35}[r][r][\tsize]{}%
\psfrag{v36}[r][r][\tsize]{-6}%
\psfrag{v37}[r][r][\tsize]{}%
\psfrag{v38}[r][r][\tsize]{-4}%
\psfrag{v39}[r][r][\tsize]{}%
\psfrag{v40}[r][r][\tsize]{-2}%
\psfrag{v41}[r][r][\tsize]{}%
\psfrag{v42}[r][r][\tsize]{0}%
\psfrag{v43}[r][r][\tsize]{}%
\psfrag{v44}[r][r][\tsize]{0}%
\psfrag{v45}[r][r][\tsize]{}%
\psfrag{v46}[r][r][\tsize]{1}%
\psfrag{v47}[r][r][\tsize]{}%
\psfrag{v48}[r][r][\tsize]{2}%
\psfrag{v49}[r][r][\tsize]{}%
\psfrag{v50}[r][r][\tsize]{3}%
\psfrag{v51}[r][r][\tsize]{}%
\psfrag{v52}[r][r][\tsize]{4}%
%
\includegraphics[width=\textwidth]{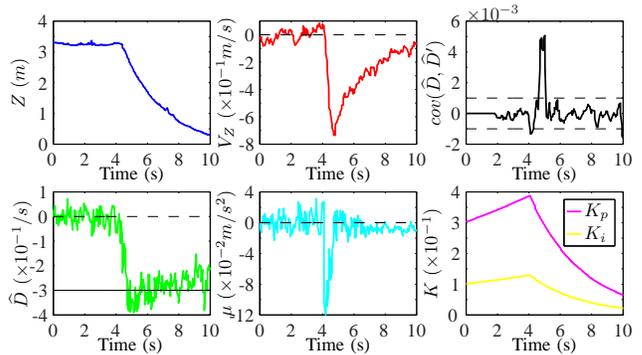}
\end{psfrags}%
%

		\caption{$D^*=-0.3$.}
		\label{fig:indoor_adaptive_control_sp_0_3}
	\end{subfigure}
	\caption{Constant flow divergence landing using the adaptive controller (indoor environment).}
	\label{fig:indoor_adaptive_control}
\end{figure}

For indoor landing tests, the vertical control is performed using the flow divergence from an on-board camera while the horizontal control is executed using the position and velocity provided by the OptiTrack system. Fig.~\ref{fig:indoor_adaptive_control} shows the experiment results of the landings at different desired flow divergence values ($D^*=-0.1,-0.2,-0.3$).

From the results shown in this figure, the controller gains are gradually increased until the first oscillation is detected while hovering, i.e., by tracking $D^*=0$. After obtaining the initial gains, the landing starts by tracking $D^*=-0.1$, $-0.2$, or $-0.3$. There are two important observations from these results: 1) $\widehat{D} \approx D^*$, and 2) $cov(\widehat{D},\widehat{D}')$ is small and bounded ($\approx 10^{-3}$). This means that the tracking performance is good, and the self-induced oscillations are prevented (see Figs.~\ref{fig:fixed_gain_size_divergence_PI} and \ref{fig:cov_fixed_gain_size_divergence_PI} for comparison). Additionally, the adaptive control scheme performs well for the three desired flow divergences.

\subsection{Outdoor Flight Tests}
\label{subsec:OutdoorExperiments}
\begin{figure}
	\centering
	\begin{subfigure}[b]{0.5\textwidth}
%
%
\begin{psfrags}%
\psfragscanon%
\newcommand{\tsize}{0.7}
%
\psfrag{s15}[b][b][\tsize]{\color[rgb]{0,0,0}\setlength{\tabcolsep}{0pt}\begin{tabular}{c}Time~(s)\end{tabular}}%
\psfrag{s16}[t][b][\tsize]{\color[rgb]{0,0,0}\setlength{\tabcolsep}{0pt}\begin{tabular}{c}$Z~(m)$\end{tabular}}%
\psfrag{s17}[b][b][\tsize]{\color[rgb]{0,0,0}\setlength{\tabcolsep}{0pt}\begin{tabular}{c}Time~(s)\end{tabular}}%
\psfrag{s18}[t][b][\tsize]{\color[rgb]{0,0,0}\setlength{\tabcolsep}{0pt}\begin{tabular}{c}$V_Z~(\times 10^{-1}m/s)$\end{tabular}}%
\psfrag{s19}[b][b][\tsize]{\color[rgb]{0,0,0}\setlength{\tabcolsep}{0pt}\begin{tabular}{c}Time~(s)\end{tabular}}%
\psfrag{s20}[t][b][\tsize]{\color[rgb]{0,0,0}\setlength{\tabcolsep}{0pt}\begin{tabular}{c}$cov(\widehat{D},\widehat{D}')$\end{tabular}}%
\psfrag{s21}[b][b][\tsize]{\color[rgb]{0,0,0}\setlength{\tabcolsep}{0pt}\begin{tabular}{c}Time~(s)\end{tabular}}%
\psfrag{s22}[t][b][\tsize]{\color[rgb]{0,0,0}\setlength{\tabcolsep}{0pt}\begin{tabular}{c}$\widehat{D}~(\times 10^{-1}/s)$\end{tabular}}%
\psfrag{s23}[b][b][\tsize]{\color[rgb]{0,0,0}\setlength{\tabcolsep}{0pt}\begin{tabular}{c}Time~(s)\end{tabular}}%
\psfrag{s24}[t][b][\tsize]{\color[rgb]{0,0,0}\setlength{\tabcolsep}{0pt}\begin{tabular}{c}$\mu~(\times 10^{-2}m/s^2)$\end{tabular}}%
\psfrag{s25}[b][b][\tsize]{\color[rgb]{0,0,0}\setlength{\tabcolsep}{0pt}\begin{tabular}{c}Time~(s)\end{tabular}}%
\psfrag{s26}[t][b][\tsize]{\color[rgb]{0,0,0}\setlength{\tabcolsep}{0pt}\begin{tabular}{c}$K~(\times 10^{-1})$\end{tabular}}%
\psfrag{s30}[][]{\color[rgb]{0,0,0}\setlength{\tabcolsep}{0pt}\begin{tabular}{c} \end{tabular}}%
\psfrag{s31}[][]{\color[rgb]{0,0,0}\setlength{\tabcolsep}{0pt}\begin{tabular}{c} \end{tabular}}%
\psfrag{s32}[l][l][\tsize]{\color[rgb]{0,0,0}$K_i$}%
\psfrag{s33}[l][l][\tsize]{\color[rgb]{0,0,0}$K_p$}%
\psfrag{s34}[l][l][\tsize]{\color[rgb]{0,0,0}$K_i$}%
%
\psfrag{x01}[t][t][\tsize]{0}%
\psfrag{x02}[t][t][\tsize]{5}%
\psfrag{x03}[t][t][\tsize]{10}%
\psfrag{x04}[t][t][\tsize]{15}%
\psfrag{x05}[t][t][\tsize]{20}%
\psfrag{x06}[t][t][\tsize]{0}%
\psfrag{x07}[t][t][\tsize]{5}%
\psfrag{x08}[t][t][\tsize]{10}%
\psfrag{x09}[t][t][\tsize]{15}%
\psfrag{x10}[t][t][\tsize]{20}%
\psfrag{x11}[t][t][\tsize]{0}%
\psfrag{x12}[t][t][\tsize]{5}%
\psfrag{x13}[t][t][\tsize]{10}%
\psfrag{x14}[t][t][\tsize]{15}%
\psfrag{x15}[t][t][\tsize]{20}%
\psfrag{x16}[t][t][\tsize]{0}%
\psfrag{x17}[t][t][\tsize]{5}%
\psfrag{x18}[t][t][\tsize]{10}%
\psfrag{x19}[t][t][\tsize]{15}%
\psfrag{x20}[t][t][\tsize]{20}%
\psfrag{x21}[t][t][\tsize]{0}%
\psfrag{x22}[t][t][\tsize]{5}%
\psfrag{x23}[t][t][\tsize]{10}%
\psfrag{x24}[t][t][\tsize]{15}%
\psfrag{x25}[t][t][\tsize]{20}%
\psfrag{x26}[t][t][\tsize]{0}%
\psfrag{x27}[t][t][\tsize]{5}%
\psfrag{x28}[t][t][\tsize]{10}%
\psfrag{x29}[t][t][\tsize]{15}%
\psfrag{x30}[t][t][\tsize]{20}%
%
\psfrag{v01}[r][r][\tsize]{0}%
\psfrag{v02}[r][r][\tsize]{}%
\psfrag{v03}[r][r][\tsize]{1}%
\psfrag{v04}[r][r][\tsize]{}%
\psfrag{v05}[r][r][\tsize]{2}%
\psfrag{v06}[r][r][\tsize]{}%
\psfrag{v07}[r][r][\tsize]{3}%
\psfrag{v08}[r][r][\tsize]{}%
\psfrag{v09}[r][r][\tsize]{4}%
\psfrag{v10}[r][r][\tsize]{-8}%
\psfrag{v11}[r][r][\tsize]{}%
\psfrag{v12}[r][r][\tsize]{-4}%
\psfrag{v13}[r][r][\tsize]{}%
\psfrag{v14}[r][r][\tsize]{0}%
\psfrag{v15}[r][r][\tsize]{}%
\psfrag{v16}[r][r][\tsize]{4}%
\psfrag{v17}[r][r][\tsize]{}%
\psfrag{v18}[r][r][\tsize]{8}%
\psfrag{v19}[r][r][\tsize]{}%
\psfrag{v20}[r][r][\tsize]{-4}%
\psfrag{v21}[r][r][\tsize]{}%
\psfrag{v22}[r][r][\tsize]{-2}%
\psfrag{v23}[r][r][\tsize]{}%
\psfrag{v24}[r][r][\tsize]{0}%
\psfrag{v25}[r][r][\tsize]{}%
\psfrag{v26}[r][r][\tsize]{2}%
\psfrag{v27}[r][r][\tsize]{}%
\psfrag{v28}[r][r][\tsize]{}%
\psfrag{v29}[r][r][\tsize]{-8}%
\psfrag{v30}[r][r][\tsize]{}%
\psfrag{v31}[r][r][\tsize]{-4}%
\psfrag{v32}[r][r][\tsize]{}%
\psfrag{v33}[r][r][\tsize]{0}%
\psfrag{v34}[r][r][\tsize]{}%
\psfrag{v35}[r][r][\tsize]{4}%
\psfrag{ypower5}[Bl][Bl][\tsize]{$\times 10^{-3}$}%
\psfrag{v36}[r][r][\tsize]{-8}%
\psfrag{v37}[r][r][\tsize]{}%
\psfrag{v38}[r][r][\tsize]{-4}%
\psfrag{v39}[r][r][\tsize]{}%
\psfrag{v40}[r][r][\tsize]{0}%
\psfrag{v41}[r][r][\tsize]{}%
\psfrag{v42}[r][r][\tsize]{4}%
\psfrag{v43}[r][r][\tsize]{}%
\psfrag{v44}[r][r][\tsize]{8}%
\psfrag{v45}[r][r][\tsize]{0}%
\psfrag{v46}[r][r][\tsize]{}%
\psfrag{v47}[r][r][\tsize]{2}%
\psfrag{v48}[r][r][\tsize]{}%
\psfrag{v49}[r][r][\tsize]{4}%
\psfrag{v50}[r][r][\tsize]{}%
\psfrag{v51}[r][r][\tsize]{6}%
%
\includegraphics[width=\textwidth]{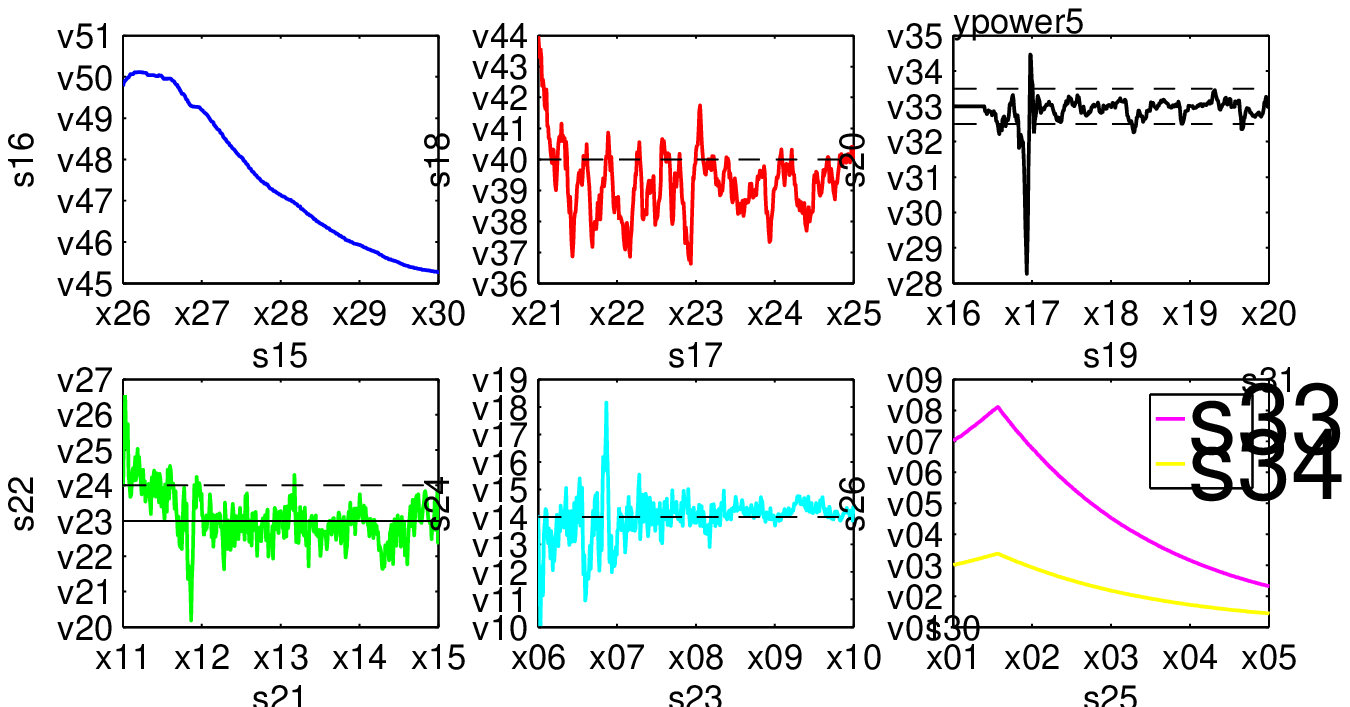}
\end{psfrags}%
%

		\caption{$D^*=-0.1$.}
		\label{fig:outdoor_adaptive_control_sp_0_1}
	\end{subfigure}
	\\ 
	\begin{subfigure}[b]{0.5\textwidth}
%
%
\begin{psfrags}%
\psfragscanon%
\newcommand{\tsize}{0.7}
%
\psfrag{s15}[b][b][\tsize]{\color[rgb]{0,0,0}\setlength{\tabcolsep}{0pt}\begin{tabular}{c}Time~(s)\end{tabular}}%
\psfrag{s16}[t][b][\tsize]{\color[rgb]{0,0,0}\setlength{\tabcolsep}{0pt}\begin{tabular}{c}$Z~(m)$\end{tabular}}%
\psfrag{s17}[b][b][\tsize]{\color[rgb]{0,0,0}\setlength{\tabcolsep}{0pt}\begin{tabular}{c}Time~(s)\end{tabular}}%
\psfrag{s18}[t][b][\tsize]{\color[rgb]{0,0,0}\setlength{\tabcolsep}{0pt}\begin{tabular}{c}$V_Z~(\times 10^{-1}m/s)$\end{tabular}}%
\psfrag{s19}[b][b][\tsize]{\color[rgb]{0,0,0}\setlength{\tabcolsep}{0pt}\begin{tabular}{c}Time~(s)\end{tabular}}%
\psfrag{s20}[t][b][\tsize]{\color[rgb]{0,0,0}\setlength{\tabcolsep}{0pt}\begin{tabular}{c}$cov(\widehat{D},\widehat{D}')$\end{tabular}}%
\psfrag{s21}[b][b][\tsize]{\color[rgb]{0,0,0}\setlength{\tabcolsep}{0pt}\begin{tabular}{c}Time~(s)\end{tabular}}%
\psfrag{s22}[t][b][\tsize]{\color[rgb]{0,0,0}\setlength{\tabcolsep}{0pt}\begin{tabular}{c}$\widehat{D}~(\times 10^{-1}/s)$\end{tabular}}%
\psfrag{s23}[b][b][\tsize]{\color[rgb]{0,0,0}\setlength{\tabcolsep}{0pt}\begin{tabular}{c}Time~(s)\end{tabular}}%
\psfrag{s24}[t][b][\tsize]{\color[rgb]{0,0,0}\setlength{\tabcolsep}{0pt}\begin{tabular}{c}$\mu~(\times 10^{-2}m/s^2)$\end{tabular}}%
\psfrag{s25}[b][b][\tsize]{\color[rgb]{0,0,0}\setlength{\tabcolsep}{0pt}\begin{tabular}{c}Time~(s)\end{tabular}}%
\psfrag{s26}[t][b][\tsize]{\color[rgb]{0,0,0}\setlength{\tabcolsep}{0pt}\begin{tabular}{c}$K~(\times 10^{-1})$\end{tabular}}%
\psfrag{s30}[][]{\color[rgb]{0,0,0}\setlength{\tabcolsep}{0pt}\begin{tabular}{c} \end{tabular}}%
\psfrag{s31}[][]{\color[rgb]{0,0,0}\setlength{\tabcolsep}{0pt}\begin{tabular}{c} \end{tabular}}%
\psfrag{s32}[l][l][\tsize]{\color[rgb]{0,0,0}$K_i$}%
\psfrag{s33}[l][l][\tsize]{\color[rgb]{0,0,0}$K_p$}%
\psfrag{s34}[l][l][\tsize]{\color[rgb]{0,0,0}$K_i$}%
%
\psfrag{x01}[t][t][\tsize]{0}%
\psfrag{x02}[t][t][\tsize]{2}%
\psfrag{x03}[t][t][\tsize]{4}%
\psfrag{x04}[t][t][\tsize]{6}%
\psfrag{x05}[t][t][\tsize]{8}%
\psfrag{x06}[t][t][\tsize]{10}%
\psfrag{x07}[t][t][\tsize]{0}%
\psfrag{x08}[t][t][\tsize]{2}%
\psfrag{x09}[t][t][\tsize]{4}%
\psfrag{x10}[t][t][\tsize]{6}%
\psfrag{x11}[t][t][\tsize]{8}%
\psfrag{x12}[t][t][\tsize]{10}%
\psfrag{x13}[t][t][\tsize]{0}%
\psfrag{x14}[t][t][\tsize]{2}%
\psfrag{x15}[t][t][\tsize]{4}%
\psfrag{x16}[t][t][\tsize]{6}%
\psfrag{x17}[t][t][\tsize]{8}%
\psfrag{x18}[t][t][\tsize]{10}%
\psfrag{x19}[t][t][\tsize]{0}%
\psfrag{x20}[t][t][\tsize]{2}%
\psfrag{x21}[t][t][\tsize]{4}%
\psfrag{x22}[t][t][\tsize]{6}%
\psfrag{x23}[t][t][\tsize]{8}%
\psfrag{x24}[t][t][\tsize]{10}%
\psfrag{x25}[t][t][\tsize]{0}%
\psfrag{x26}[t][t][\tsize]{2}%
\psfrag{x27}[t][t][\tsize]{4}%
\psfrag{x28}[t][t][\tsize]{6}%
\psfrag{x29}[t][t][\tsize]{8}%
\psfrag{x30}[t][t][\tsize]{10}%
\psfrag{x31}[t][t][\tsize]{0}%
\psfrag{x32}[t][t][\tsize]{2}%
\psfrag{x33}[t][t][\tsize]{4}%
\psfrag{x34}[t][t][\tsize]{6}%
\psfrag{x35}[t][t][\tsize]{8}%
\psfrag{x36}[t][t][\tsize]{10}%
%
\psfrag{v01}[r][r][\tsize]{0}%
\psfrag{v02}[r][r][\tsize]{}%
\psfrag{v03}[r][r][\tsize]{1}%
\psfrag{v04}[r][r][\tsize]{}%
\psfrag{v05}[r][r][\tsize]{2}%
\psfrag{v06}[r][r][\tsize]{}%
\psfrag{v07}[r][r][\tsize]{3}%
\psfrag{v08}[r][r][\tsize]{}%
\psfrag{v09}[r][r][\tsize]{-12}%
\psfrag{v10}[r][r][\tsize]{}%
\psfrag{v11}[r][r][\tsize]{-8}%
\psfrag{v12}[r][r][\tsize]{}%
\psfrag{v13}[r][r][\tsize]{-4}%
\psfrag{v14}[r][r][\tsize]{}%
\psfrag{v15}[r][r][\tsize]{0}%
\psfrag{v16}[r][r][\tsize]{}%
\psfrag{v17}[r][r][\tsize]{4}%
\psfrag{v18}[r][r][\tsize]{}%
\psfrag{v19}[r][r][\tsize]{}%
\psfrag{v20}[r][r][\tsize]{-4}%
\psfrag{v21}[r][r][\tsize]{}%
\psfrag{v22}[r][r][\tsize]{-2}%
\psfrag{v23}[r][r][\tsize]{}%
\psfrag{v24}[r][r][\tsize]{0}%
\psfrag{v25}[r][r][\tsize]{}%
\psfrag{v26}[r][r][\tsize]{2}%
\psfrag{v27}[r][r][\tsize]{-6}%
\psfrag{v28}[r][r][\tsize]{-4}%
\psfrag{v29}[r][r][\tsize]{-2}%
\psfrag{v30}[r][r][\tsize]{0}%
\psfrag{v31}[r][r][\tsize]{2}%
\psfrag{v32}[r][r][\tsize]{4}%
\psfrag{v33}[r][r][\tsize]{6}%
\psfrag{ypower5}[Bl][Bl][\tsize]{$\times 10^{-3}$}%
\psfrag{v34}[r][r][\tsize]{-8}%
\psfrag{v35}[r][r][\tsize]{}%
\psfrag{v36}[r][r][\tsize]{-4}%
\psfrag{v37}[r][r][\tsize]{}%
\psfrag{v38}[r][r][\tsize]{0}%
\psfrag{v39}[r][r][\tsize]{}%
\psfrag{v40}[r][r][\tsize]{4}%
\psfrag{v41}[r][r][\tsize]{}%
\psfrag{v42}[r][r][\tsize]{8}%
\psfrag{v43}[r][r][\tsize]{0}%
\psfrag{v44}[r][r][\tsize]{}%
\psfrag{v45}[r][r][\tsize]{2}%
\psfrag{v46}[r][r][\tsize]{}%
\psfrag{v47}[r][r][\tsize]{4}%
\psfrag{v48}[r][r][\tsize]{}%
\psfrag{v49}[r][r][\tsize]{6}%
%
\includegraphics[width=\textwidth]{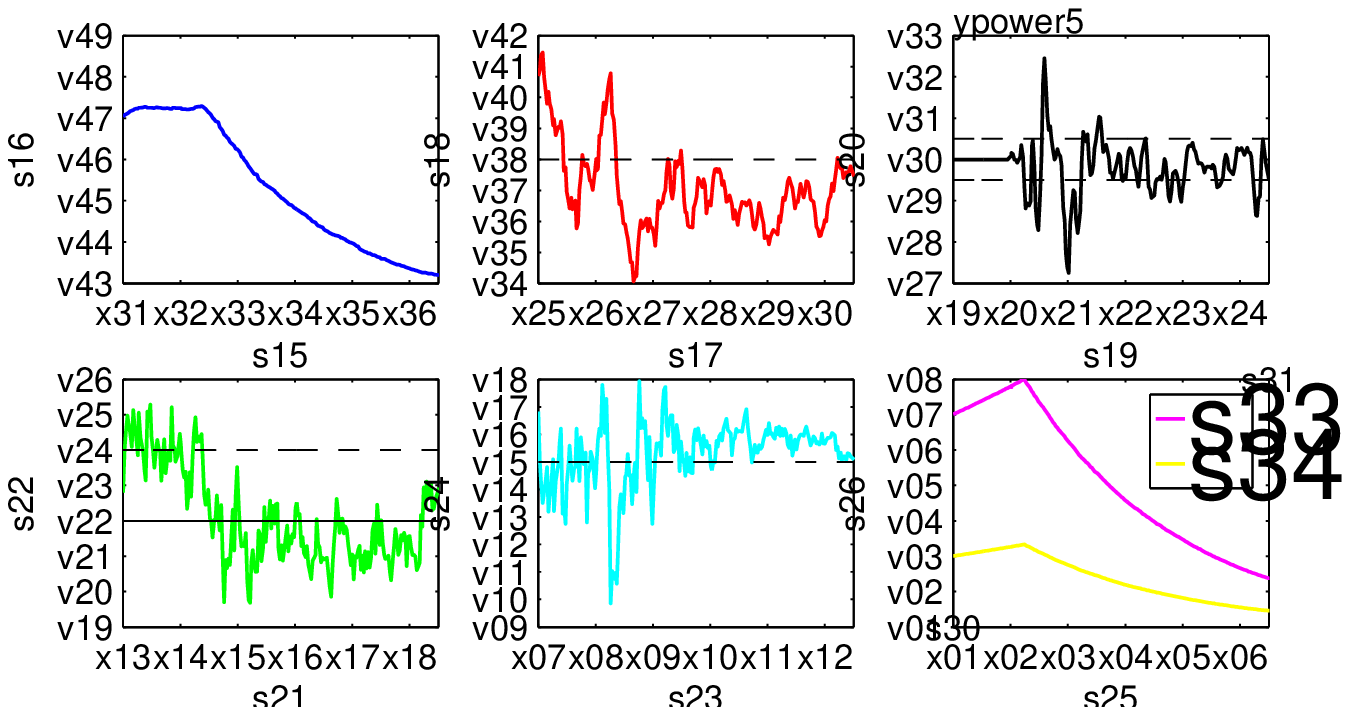}
\end{psfrags}%
%

		\caption{$D^*=-0.2$.}
		\label{fig:outdoor_adaptive_control_sp_0_2}
	\end{subfigure}
	\\ 
	\begin{subfigure}[b]{0.5\textwidth}
%
%
\begin{psfrags}%
\psfragscanon%
\newcommand{\tsize}{0.7}
%
\psfrag{s15}[b][b][\tsize]{\color[rgb]{0,0,0}\setlength{\tabcolsep}{0pt}\begin{tabular}{c}Time~(s)\end{tabular}}%
\psfrag{s16}[t][b][\tsize]{\color[rgb]{0,0,0}\setlength{\tabcolsep}{0pt}\begin{tabular}{c}$Z~(m)$\end{tabular}}%
\psfrag{s17}[b][b][\tsize]{\color[rgb]{0,0,0}\setlength{\tabcolsep}{0pt}\begin{tabular}{c}Time~(s)\end{tabular}}%
\psfrag{s18}[t][b][\tsize]{\color[rgb]{0,0,0}\setlength{\tabcolsep}{0pt}\begin{tabular}{c}$V_Z~(\times 10^{-1}m/s)$\end{tabular}}%
\psfrag{s19}[b][b][\tsize]{\color[rgb]{0,0,0}\setlength{\tabcolsep}{0pt}\begin{tabular}{c}Time~(s)\end{tabular}}%
\psfrag{s20}[t][b][\tsize]{\color[rgb]{0,0,0}\setlength{\tabcolsep}{0pt}\begin{tabular}{c}$cov(\widehat{D},\widehat{D}')$\end{tabular}}%
\psfrag{s21}[b][b][\tsize]{\color[rgb]{0,0,0}\setlength{\tabcolsep}{0pt}\begin{tabular}{c}Time~(s)\end{tabular}}%
\psfrag{s22}[t][b][\tsize]{\color[rgb]{0,0,0}\setlength{\tabcolsep}{0pt}\begin{tabular}{c}$\widehat{D}~(\times 10^{-1}/s)$\end{tabular}}%
\psfrag{s23}[b][b][\tsize]{\color[rgb]{0,0,0}\setlength{\tabcolsep}{0pt}\begin{tabular}{c}Time~(s)\end{tabular}}%
\psfrag{s24}[t][b][\tsize]{\color[rgb]{0,0,0}\setlength{\tabcolsep}{0pt}\begin{tabular}{c}$\mu~(\times 10^{-1}m/s^2)$\end{tabular}}%
\psfrag{s25}[b][b][\tsize]{\color[rgb]{0,0,0}\setlength{\tabcolsep}{0pt}\begin{tabular}{c}Time~(s)\end{tabular}}%
\psfrag{s26}[t][b][\tsize]{\color[rgb]{0,0,0}\setlength{\tabcolsep}{0pt}\begin{tabular}{c}$K~(\times 10^{-1})$\end{tabular}}%
\psfrag{s30}[][]{\color[rgb]{0,0,0}\setlength{\tabcolsep}{0pt}\begin{tabular}{c} \end{tabular}}%
\psfrag{s31}[][]{\color[rgb]{0,0,0}\setlength{\tabcolsep}{0pt}\begin{tabular}{c} \end{tabular}}%
\psfrag{s32}[l][l][\tsize]{\color[rgb]{0,0,0}$K_i$}%
\psfrag{s33}[l][l][\tsize]{\color[rgb]{0,0,0}$K_p$}%
\psfrag{s34}[l][l][\tsize]{\color[rgb]{0,0,0}$K_i$}%
%
\psfrag{x01}[t][t][\tsize]{0}%
\psfrag{x02}[t][t][\tsize]{2}%
\psfrag{x03}[t][t][\tsize]{4}%
\psfrag{x04}[t][t][\tsize]{6}%
\psfrag{x05}[t][t][\tsize]{8}%
\psfrag{x06}[t][t][\tsize]{10}%
\psfrag{x07}[t][t][\tsize]{0}%
\psfrag{x08}[t][t][\tsize]{2}%
\psfrag{x09}[t][t][\tsize]{4}%
\psfrag{x10}[t][t][\tsize]{6}%
\psfrag{x11}[t][t][\tsize]{8}%
\psfrag{x12}[t][t][\tsize]{10}%
\psfrag{x13}[t][t][\tsize]{0}%
\psfrag{x14}[t][t][\tsize]{2}%
\psfrag{x15}[t][t][\tsize]{4}%
\psfrag{x16}[t][t][\tsize]{6}%
\psfrag{x17}[t][t][\tsize]{8}%
\psfrag{x18}[t][t][\tsize]{10}%
\psfrag{x19}[t][t][\tsize]{0}%
\psfrag{x20}[t][t][\tsize]{2}%
\psfrag{x21}[t][t][\tsize]{4}%
\psfrag{x22}[t][t][\tsize]{6}%
\psfrag{x23}[t][t][\tsize]{8}%
\psfrag{x24}[t][t][\tsize]{10}%
\psfrag{x25}[t][t][\tsize]{0}%
\psfrag{x26}[t][t][\tsize]{2}%
\psfrag{x27}[t][t][\tsize]{4}%
\psfrag{x28}[t][t][\tsize]{6}%
\psfrag{x29}[t][t][\tsize]{8}%
\psfrag{x30}[t][t][\tsize]{10}%
\psfrag{x31}[t][t][\tsize]{0}%
\psfrag{x32}[t][t][\tsize]{2}%
\psfrag{x33}[t][t][\tsize]{4}%
\psfrag{x34}[t][t][\tsize]{6}%
\psfrag{x35}[t][t][\tsize]{8}%
\psfrag{x36}[t][t][\tsize]{10}%
%
\psfrag{v01}[r][r][\tsize]{0}%
\psfrag{v02}[r][r][\tsize]{}%
\psfrag{v03}[r][r][\tsize]{1}%
\psfrag{v04}[r][r][\tsize]{}%
\psfrag{v05}[r][r][\tsize]{2}%
\psfrag{v06}[r][r][\tsize]{}%
\psfrag{v07}[r][r][\tsize]{3}%
\psfrag{v08}[r][r][\tsize]{}%
\psfrag{v09}[r][r][\tsize]{4}%
\psfrag{v10}[r][r][\tsize]{}%
\psfrag{v11}[r][r][\tsize]{-1}%
\psfrag{v12}[r][r][\tsize]{}%
\psfrag{v13}[r][r][\tsize]{0}%
\psfrag{v14}[r][r][\tsize]{}%
\psfrag{v15}[r][r][\tsize]{1}%
\psfrag{v16}[r][r][\tsize]{-6}%
\psfrag{v17}[r][r][\tsize]{}%
\psfrag{v18}[r][r][\tsize]{-4}%
\psfrag{v19}[r][r][\tsize]{}%
\psfrag{v20}[r][r][\tsize]{-2}%
\psfrag{v21}[r][r][\tsize]{}%
\psfrag{v22}[r][r][\tsize]{0}%
\psfrag{v23}[r][r][\tsize]{}%
\psfrag{v24}[r][r][\tsize]{2}%
\psfrag{v25}[r][r][\tsize]{}%
\psfrag{v26}[r][r][\tsize]{-2}%
\psfrag{v27}[r][r][\tsize]{}%
\psfrag{v28}[r][r][\tsize]{0}%
\psfrag{v29}[r][r][\tsize]{}%
\psfrag{v30}[r][r][\tsize]{2}%
\psfrag{v31}[r][r][\tsize]{}%
\psfrag{v32}[r][r][\tsize]{4}%
\psfrag{v33}[r][r][\tsize]{}%
\psfrag{v34}[r][r][\tsize]{6}%
\psfrag{ypower5}[Bl][Bl][\tsize]{$\times 10^{-3}$}%
\psfrag{v35}[r][r][\tsize]{-8}%
\psfrag{v36}[r][r][\tsize]{}%
\psfrag{v37}[r][r][\tsize]{-4}%
\psfrag{v38}[r][r][\tsize]{}%
\psfrag{v39}[r][r][\tsize]{0}%
\psfrag{v40}[r][r][\tsize]{}%
\psfrag{v41}[r][r][\tsize]{4}%
\psfrag{v42}[r][r][\tsize]{}%
\psfrag{v43}[r][r][\tsize]{0}%
\psfrag{v44}[r][r][\tsize]{}%
\psfrag{v45}[r][r][\tsize]{2}%
\psfrag{v46}[r][r][\tsize]{}%
\psfrag{v47}[r][r][\tsize]{4}%
\psfrag{v48}[r][r][\tsize]{}%
\psfrag{v49}[r][r][\tsize]{6}%
%
\includegraphics[width=\textwidth]{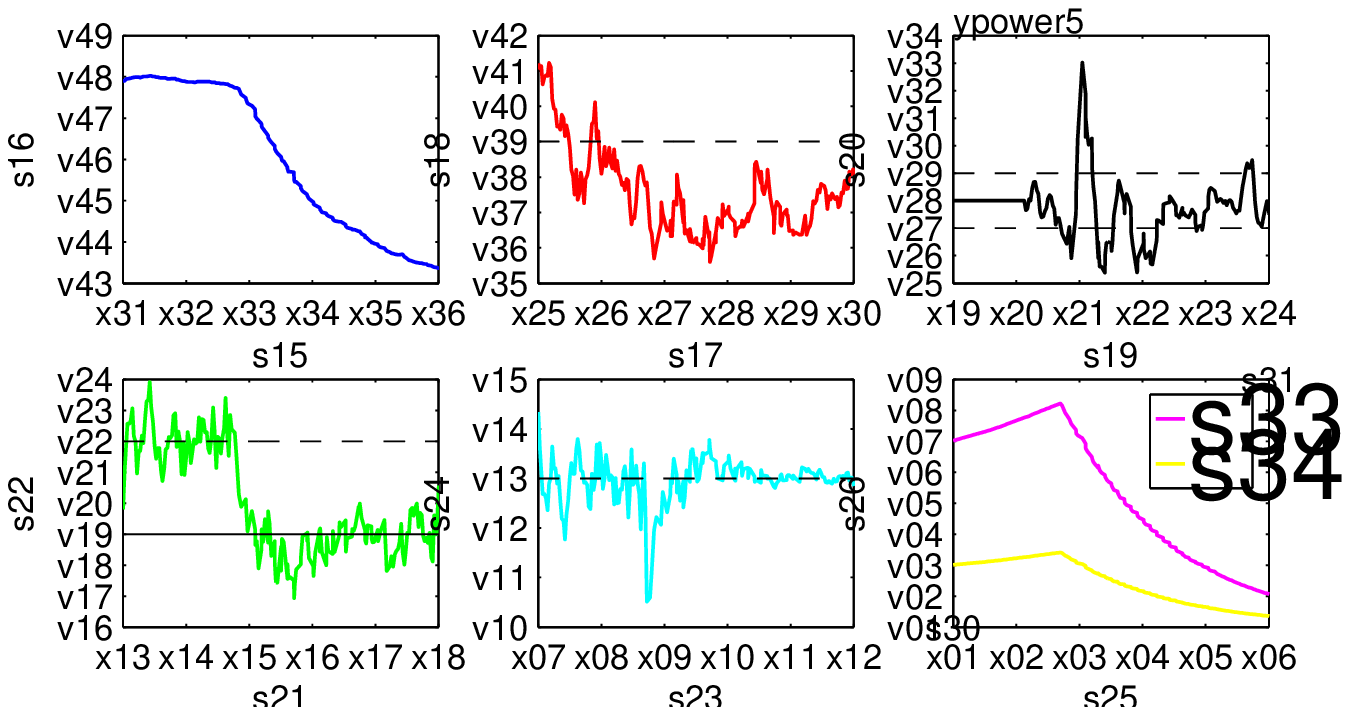}
\end{psfrags}%
%

		\caption{$D^*=-0.3$.}
		\label{fig:outdoor_adaptive_control_sp_0_3}
	\end{subfigure}
	\caption{Constant flow divergence landing using the adaptive controller (outdoor environment).}
	\label{fig:outdoor_adaptive_control}
\end{figure}

Outdoor flight is more challenging, due to the presence of wind. In outdoor experiments, the vertical dynamics are also controlled using flow divergence, while the horizontal dynamics are stabilized using translational optical flow estimates. The wind speed during the flights was reported to be around $8~knots$, and the maximum gust was approximately $10~knots$\footnote{Wind speed reference: https://www.windfinder.com/}. Fig.~\ref{fig:outdoor_adaptive_control} shows the results of the outdoor landing tests at different desired flow divergences ($D^*=-0.1,-0.2,-0.3$). 

From the outdoor flight results, the adaptive controller tracks the desired flow divergences well, and the self-induced oscillations are prevented. Because of the unknown wind disturbances, it can be observed that slight perturbations exist, but the controller is sufficiently robust to deal with wind. Note that in the figure the height is obtained from an ultrasound sensor, while the vertical velocity is provided from a GPS which has an accurate of $<3m$ and has a relatively low update rate, i.e., $5Hz$. Both these sensors have only been used for logging purposes, and not in the control.

\section{Conclusions}
\label{sec:Conclusion}
A control strategy has been developed to solve the fundamental problem of gain selection for constant flow divergence landings. The delay and noise models of the estimates were first obtained, and their effects on closed-loop control performance were investigated. In the presence of the delay and noise, computer simulations as well as real flight tests show that oscillations occur during the landings when a fixed-gain controller is used. We propose an adaptive controller which first initializes a near-optimal gain by means of an oscillating movement and then exponentially reduces this gain during descent. A stability analysis shows that the adaptive gain strategy indeed prevents self-induced oscillations and instability. This strategy was implemented on a Parrot AR.Drone 2.0 with all vision and control algorithms running on-board. Multiple successful landing flight tests, in both indoor and windy outdoor environments, were performed using the adaptive gain strategy. 
	
	%

	%
	%
	%
	%
	%
	
	\ifCLASSOPTIONcaptionsoff
	\newpage
	\fi

	
	
	\bibliographystyle{IEEEtran}
	\bibliography{IEEEabrv,Divergence_Landing_ArXiv}
\end{document}